\documentclass{article}

\PassOptionsToPackage{numbers}{natbib}
\usepackage[main, final]{neurips_2026}

\usepackage[utf8]{inputenc} 
\usepackage[T1]{fontenc}    
\usepackage{hyperref}       
\usepackage{url}            
\usepackage{booktabs}       
\usepackage{amsfonts}       
\usepackage{nicefrac}       
\usepackage{microtype}      
\usepackage{xcolor}         
\usepackage{amsmath}
\usepackage{longtable}
\usepackage{amsfonts}
\usepackage{amsthm}
\usepackage{hyperref}
\usepackage{nameref}
\usepackage{url}
\usepackage{amssymb}
\usepackage{graphicx}
\usepackage{subfigure}
\usepackage{threeparttable}
\usepackage{booktabs}
\usepackage{thm-restate}
\usepackage{algorithm}
\usepackage{algorithmic}
\usepackage{tikz}
\usepackage[percent]{overpic}
\usepackage{caption}
\usepackage{xurl}
\usetikzlibrary{arrows.meta, positioning}
\newtheorem{theorem}{Theorem}[section] 
\newtheorem{lemma}[theorem]{Lemma}

\title{Aligning Inductive Bias for Data-Efficient Generalization in State Space Models}

\author{%
  Qiyu Chen$^{1,2}$ \qquad Guozhang Chen$^{1,3}\thanks{Corresponding author: \texttt{guozhang.chen@pku.edu.cn}}$ \\
  $^{1}$State Key Laboratory of Multimedia Information Processing, School of Computer Science \\
  $^{2}$School of Physics \\
  $^{3}$National Engineering Research Center of Visual Technology, School of Computer Science\\
  Peking University, Beijing 100871, China \\
  \texttt{cqy@stu.pku.edu.cn} \qquad
  \texttt{guozhang.chen@pku.edu.cn}
}

\begin{document}

\maketitle

\begin{abstract}
The remarkable success of modern AI has been closely tied to scaling laws, yet the finite supply of high-quality data makes data efficiency—learning more from less—an increasingly important frontier. A model’s inductive bias is a critical lever for data efficiency, but foundational sequence models such as State Space Models (SSMs) often rely on fixed, task-agnostic biases. When this fixed prior is misaligned with the underlying structure of a task, the model may require additional samples to overcome its own bias before learning the relevant signal. In this work, we introduce a principled framework for understanding and aligning the inductive bias of linear time-invariant SSMs. We first formalize this bias through an SSM-induced kernel and show theoretically and empirically that its spectrum is governed by the model’s frequency response. This characterization motivates Task-Dependent Initialization (TDI), a fast power-spectrum matching method that aligns the initial SSM bias with the task’s spectral characteristics before downstream training. Across controlled synthetic experiments, trainable one-layer SSMs, and deep SSMs on diverse real-world benchmarks, TDI can improve learning under fixed training budgets primarily when labeled data are limited, label-relevant spectral structure is estimable, and the default SSM initialization is spectrally mismatched. These benefits include better early learning, but do not necessarily imply higher accuracy after extended training. Our results provide both a theoretical lens and a practical tool for task-adaptive inductive bias, suggesting a path toward more data-efficient sequence modeling. Code is available at
\url{https://github.com/Criticality-Cognitive-Computation-Lab/Aligning-Inductive-Bias-for-State-Space-Models}.
\end{abstract}
\section{Introduction}
The paradigm of scaling laws, wherein model performance predictably improves with more data, parameters, and compute, has driven unprecedented progress in AI~\citep{kaplan2020scaling, hestness2017deep, bahri2024explaining}. However, high-quality training data is finite and often difficult to obtain in scientific, medical, industrial, and long-tail domains~\citep{shorten2019survey, sun2017revisiting, citovsky2021batch}. This makes data efficiency---learning more from fewer samples---an increasingly important goal~\citep{shorten2019survey, adadi2021survey}. A central lever for data efficiency is \emph{inductive bias}: the built-in assumptions that make some functions easier to learn than others~\citep{hastie2009elements, battaglia2018relational, goyal2022inductive}. When this bias is aligned with the task, it can reduce sample complexity; when it is misaligned, the model may require additional data to compensate for an unfavorable prior.

\begin{figure}[ht]
\centering
\includegraphics[width=\textwidth]{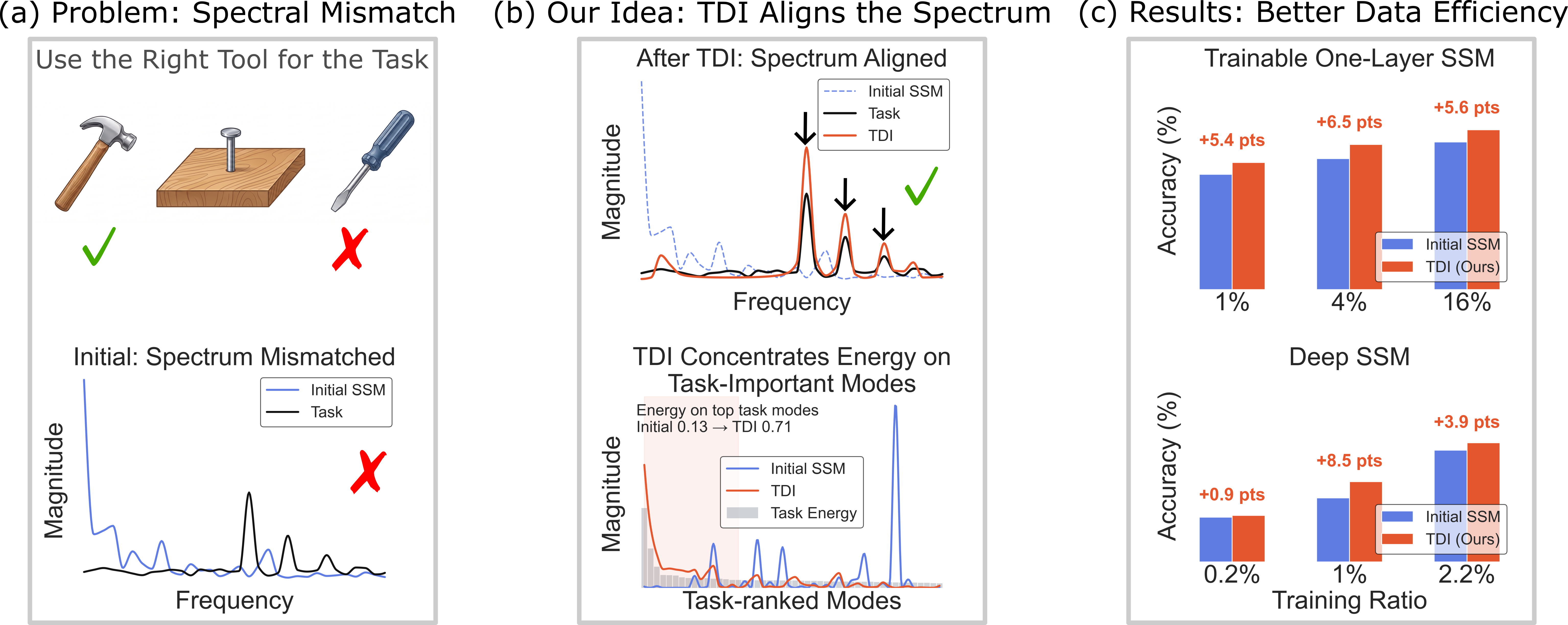}
\caption{\textbf{Graphic overview.} (a) A fixed, task-agnostic SSM initialization can place spectral energy in frequency regions that do not match the task-relevant spectrum, analogous to using an unsuitable tool for a job.
(b) TDI initializes the SSM so that its frequency response aligns with task-relevant spectral components and assigns more energy to task-important modes.
(c) Across trainable one-layer SSMs and deep SSMs, TDI achieves higher mean final-epoch test accuracy under fixed training budgets. A training ratio \(r\) denotes training on an \(r\)-fraction of the original training split, while keeping the validation and test splits fixed.
}
\label{fig:intro}
\end{figure}

State Space Models (SSMs) have emerged as a powerful and efficient architecture for sequence modeling~\citep{gu2022efficientlymodelinglongsequences, gu2022parameterizationinitializationdiagonalstate, smith2023simplified, mamba, mamba2, patro2025mamba}.  Many widely used SSM layers, including S4 and S4D~\citep{gu2022efficientlymodelinglongsequences,gu2022parameterizationinitializationdiagonalstate}, are built from linear time-invariant (LTI) filtering operators, which transform an input sequence through a structured convolution before the readout. Although recent selective models such as Mamba~\citep{mamba, mamba2} introduce input-dependent dynamics, the LTI case remains a fundamental and analytically tractable setting for understanding how SSM filtering induces inductive bias. This filtering operator is not neutral: it emphasizes some temporal and frequency modes more than others.  Existing LTI-based initializations are typically task-agnostic and therefore impose a fixed spectral prior before downstream training~\citep{gu2022parameterizationinitializationdiagonalstate, yu2024tuningfrequencybiasstate}.
As illustrated in Figure~\ref{fig:intro}, this prior can be beneficial when it overlaps with task-relevant frequencies, but can become a sample-efficiency bottleneck when the task-relevant components lie in low-response regions of the model.

In this work, we make this spectral inductive bias explicit and use it to design a task-adaptive initialization method. We view an LTI SSM as a convolutional feature map, define the corresponding SSM-induced kernel, and show that its spectrum is governed by the SSM frequency response. This characterization motivates \emph{Task-Dependent Initialization (TDI)}, which estimates a task-relevant spectral statistic and aligns the initial SSM spectrum with it before downstream training. By placing more model response on task-ranked frequency modes, TDI increases the kernel mass assigned to components predicted to be task-relevant. TDI changes only the initialization, leaving the architecture, supervised objective, and inference cost unchanged. Unlike \citet{yu2024tuningfrequencybiasstate} who tune SSM frequency bias empirically, and \citet{liu2025autocorrelation} who optimize autocorrelation-based initialization, our method aligns the model to task-dependent spectrum, concentrating model's power on task-relevant frequencies (see \hyperref[app:related work]{Appendix~\ref*{app:related work}: \nameref*{app:related work}} for full comparison). TDI targets data-limited settings where label-relevant spectral structure can be estimated from the available training data and the default initialization is spectrally mismatched. Our subsampling experiments simulate limited data availability; they do not recommend discarding available training data.

Our contributions are as follows. 
\begin{itemize}
    \item \textbf{A kernel-theoretic characterization of SSM inductive bias.} We formalize LTI SSMs as convolutional feature maps and define an SSM-induced kernel whose spectrum reveals which modes are favored in the small-sample regime, and show that this spectrum is governed by the SSM frequency response. \item \textbf{Task-Dependent Initialization via spectral alignment.} Based on this characterization, we propose Task-Dependent Initialization (TDI), a lightweight initialization method that aligns the initial SSM spectrum with task-relevant spectral components without changing the downstream architecture, training objective, or inference cost. 
    \item \textbf{Validation from controlled theory to practical models.} We validate the mechanism in kernel regression, trainable one-layer SSMs, and deep SSMs, showing that gains concentrate in finite-sample regimes where task-relevant spectral structure is present and the default SSM bias is mismatched with the task.
\end{itemize}
\section{Preliminaries}
\label{sec: preliminaries}
\subsection{Problem Statement}
In finite-sample regimes, generalization depends not only on model capacity, but also on whether the model's inductive bias makes the task easy to learn. The generalization error, $E_g$, quantifies this ability by averaging the model's performance over all possible training datasets $\mathcal{D}$ of a given size $P$. 

Models rely on an inductive bias—a set of built-in assumptions—to generalize effectively. For SSMs, however, the relationship between their frequency response, induced inductive bias, and data-efficient generalization remains insufficiently characterized. The goal of this work is not merely to analyze the expressive power of SSMs, but to characterize which target components are favored by an SSM in the small-sample regime. To this end, we study the kernel induced by the SSM convolution operator, because kernel spectra provide a direct description of mode-wise learnability.
\subsection{State Space Models as Toeplitz Convolution Operators}
\label{subsec: ssm}
In general, a continuous-time linear time-invariant (LTI) state space model (SSM) is defined as
\begin{align}
        \dot{\mathbf{x}}(t) = \pmb{A} \mathbf{x}(t) + \pmb{B} \mathbf{u}(t), \quad
        \mathbf{y}(t) = \pmb{C} \mathbf{x}(t) + \pmb{D} \mathbf{u}(t),
\end{align}
where $\mathbf{x}(t) \in \mathbb{R} ^n$ denotes the hidden state, $\mathbf{u}(t) \in \mathbb{R}^m$ is the input, and $\mathbf{y}(t) \in \mathbb{R}^p$ is the output. 
We discretize the continuous SSM using the Zero-Order Hold (ZOH) method with step size $\Delta$, yielding the discrete SSM:
\begin{align}
    \mathbf{x}_{k} = \overline{\pmb{A}}\mathbf{x}_{k-1} + \overline{\pmb{B}}u_k,\quad
    y_k = \overline{\pmb{C}}\mathbf{x}_k,\quad k=0,1,\cdots L-1\quad \text{(}L\,\, \text{is input length)},
\end{align}
where $\overline{\pmb{A}}=\exp{(\Delta\pmb{A})}, \overline{\pmb{B}}=(\Delta\pmb{A})^{-1}(\exp{(\Delta \pmb{A})}-\pmb{I})\Delta\pmb{B}, \overline{\pmb{C}}=\pmb{C}$. For simplicity, and in line with modern deep SSMs such as  S4~\citep{gu2022efficientlymodelinglongsequences} and S4D~\citep{gu2022parameterizationinitializationdiagonalstate}, we omit the direct feedthrough term $\pmb{D}$ and treat $u_k$ and $y_k$ as scalars. For an initial state $\mathbf{x}_{-1}=\pmb{0}$, the output sequence $\pmb{y} \equiv [y_0,\cdots,y_{L-1}]^\top$ is the convolution of the input sequence $\pmb{u}\equiv[u_0,\cdots,u_{L-1}]^\top$ with the SSM's impulse response kernel $h$, given by $y_k = \sum_{n=0}^k h_n u_{k-n}$, where the kernel is defined as $h_n = \overline{\pmb{C}}\overline{\pmb{A}}^n\overline{\pmb{B}}.$ The frequency response $H(\omega)$ of an SSM is the discrete-time Fourier transform of its impulse response $h$, $H(\omega)=\operatorname{DTFT}(h)=\sum_{n=0}^\infty h_ne^{-i\omega n}, \,\omega\in(0,2\pi)$.
This convolution process can be represented by a Toeplitz matrix \(\pmb T_h\) that maps the input sequence \(\pmb u\) to the output sequence \(\pmb y\) as \(\pmb y=\pmb T_h\pmb u\). Specifically, \(\pmb T_h\in\mathbb R^{L\times L}\) is the lower-triangular Toeplitz matrix defined entrywise by \((\pmb T_h)_{k\ell}=h_{k-\ell}\) for \(k\ge \ell\) and \((\pmb T_h)_{k\ell}=0\) otherwise, with \(k,\ell=0,\ldots,L-1\).  Since \(\pmb T_h\) is the operator through which an SSM filters the input before the readout, we treat \(\pmb T_h\pmb u\) as the SSM feature representation of \(\pmb u\).
\subsection{A Kernel-Theoretic Framework for Analyzing Inductive Bias}
\label{subsec: kr and rkhs}
To analyze the inductive bias of SSMs, we employ the framework of kernel methods. \emph{Kernel Ridge Regression (KRR)} provides a powerful lens for this by finding a function $f$ in a Reproducing Kernel Hilbert Space (RKHS) $\mathcal{H}_K$ that minimizes the regularized empirical risk:$f^*_\mathcal{D}=\arg\min_{f \in \mathcal{H}_K} \frac{1}{P} \sum_{\mu=1}^P \lVert f(\pmb{u}^\mu)-\pmb{Y}^\mu\rVert_2^2 + \lambda \lVert f\rVert_{\mathcal{H}_K}^2$, 
where $\lambda > 0$ is a regularization parameter, $P$ is the number of training samples, and $\mathcal{D}=\{\pmb{u}^\mu,\pmb{Y}^\mu\}_{\mu=1}^P$ denotes the training set. The inputs $\pmb{u}^\mu$ are sampled i.i.d. from a distribution $p(\pmb{u})$, and labels $\pmb{Y}^\mu\in\mathbb{R}^d$ ($d$ is the dimension of labels) are generated by $\pmb{Y}^\mu=\overline{f}(\pmb{u}^\mu)+\epsilon^\mu$, where $\epsilon^\mu$ are zero-mean noise with covariance $\langle \epsilon^\mu\epsilon^\nu\rangle=\sigma^2\delta_{\mu\nu}$. The RKHS is uniquely defined by a kernel function $K(\pmb{u},\pmb{u}')$ that satisfies the reproducing property~\citep{aronszajn1950theory,scholkopf2002learning}.

The spectral properties of the kernel, revealed by Mercer's Theorem, are crucial for understanding generalization.
Mercer's Theorem~\citep{williams2006gaussian} states that, a continuous, symmetric, and positive semi-definite kernel $K(\pmb{u}, \pmb{u}')$ can be decomposed as a series expansion in terms of its eigenfunctions $\{\phi_\rho\}_{\rho=1}^\infty$ and non-negative eigenvalues $\{\eta_\rho\}_{\rho=1}^\infty$: $K(\pmb{u},\pmb{u}')=\sum_{\rho=1}^{\infty} \eta_{\rho} \phi_{\rho}(\pmb{u}) \phi_{\rho}(\pmb{u}'), $ where the eigenfunctions are normalized with respect to a given distribution $p(\pmb{u})$.

The inductive bias of the kernel is encoded in its spectrum. As shown by \citet{canatar2021spectral,bordelon2021population}, the generalization error can be decomposed into a sum of mode-wise errors, revealing a \emph{spectral bias}: modes $\rho$ corresponding to larger eigenvalues $\eta_\rho$ are learned more rapidly as the number of data samples increases. Therefore, if the target function has most of its power on large-eigenvalue modes, it is well \emph{aligned} with the model bias and can be learned with fewer samples. \emph{Cumulative power} $C(\rho)$ is introduced to quantify the alignment between task and model. After sorting the modes $\rho$ by decreasing eigenvalues $\eta_1\ge\eta_2\ge\cdots$ and expanding the target function $\overline{f}(\pmb{u})=\sum_\rho \overline{w}_\rho\phi_\rho(\pmb{u})$ in the kernel's eigenbasis, $C(\rho)=\frac{\sum_{l=1}^\rho\overline{w}_l^2}{\sum_{l=1}^\infty \overline{w}_l^2}$. Target functions whose power is concentrated in the top kernel modes (i.e., those with a fast-rising $C(\rho)$) can be learned with fewer samples. For a $d$-dimensional target function, we set $\overline{w}^2_\rho=\sum_{j=1}^d\overline{w}_{\rho,j}^2$, where each $\overline{w}_{\rho,j}$ is defined by the expansion $\overline{f}_j(\pmb{u})=\sum_\rho \overline{w}_{\rho,j}\phi_\rho(\pmb{u})$.

\section{Theory and Method}
\subsection{SSM-Induced Kernel}
\label{subsec: ssm-induced kernel}
In Section~\ref{subsec: ssm}, we showed that a discrete-time LTI SSM maps an input sequence 
$\pmb u \in \mathbb R^L$ to the filtered sequence \(\pmb T_h \pmb u\), where 
\(\pmb T_h\) is the Toeplitz convolution matrix generated by the SSM impulse response \(h\). We treat 
the SSM convolution as a feature map,
$\Phi_{\mathrm{SSM}}(\pmb u)=\frac{1}{\sqrt L}\pmb T_h \pmb u$ . The inner product between these SSM features defines a kernel over input sequences.
\begin{restatable}[SSM-Induced Kernel]{definition}{SSMInducedKernel}
\label{def: ssm induced kernel}
Let $\pmb{u}, \pmb{u}' \in \mathbb{R}^L$ be two input sequences of length $L$. The SSM-induced kernel is defined as the inner product of their corresponding Toeplitz matrix features: $
K_{\mathrm{SSM}}(\pmb{u}, \pmb{u}') = \frac{1}{L} (\pmb{T}_h \pmb{u})^\top (\pmb{T}_h \pmb{u}')$,
where $\pmb{T}_h$ denotes the Toeplitz matrix defined by the SSM's impulse response kernel.
\end{restatable}
Since \(K_{\rm SSM}\) is an inner product in the feature space \(\Phi_{\rm SSM}\), it is positive semi-definite (see \hyperref[Appendix: proof 1]{Appendix~\ref*{Appendix: proof 1}: \nameref*{Appendix: proof 1}}).
The structure of the kernel is determined by the SSM parameters $\pmb{A}, \pmb{B}, \pmb{C},$ and $\Delta$, which govern the temporal dynamics of the system.  Unlike the standard linear kernel on raw inputs, this kernel first filters the sequence through the SSM convolution operator. 
Thus, its geometry reflects the temporal filtering bias encoded by the Toeplitz operator.
\subsection{Characterizing the SSM-Induced Kernel}
\label{subsec: characterize}
We characterize the SSM-induced kernel directly through the singular value decomposition of its finite-dimensional feature operator. Write $\pmb T_h=\sqrt{L}\,\pmb U\pmb S\pmb V^\top$, where $\pmb S=\operatorname{diag}(s_1,\ldots,s_L)$ contains the scaled singular values and $\pmb v_\rho$ denotes the $\rho$-th column of $\pmb V$. Then $ \frac{1}{L}\pmb T_h^\top\pmb T_h=\pmb V\pmb S^2\pmb V^\top$. Thus, the kernel decomposition follows from the ordinary spectral decomposition of this positive semi-definite matrix. To identify these eigenvectors with orthonormal kernel eigenfunctions, we first assume whitened inputs, isolating the contribution of the SSM operator. For centered inputs with covariance $\pmb\Sigma\succ0$, writing $\pmb u=\pmb\Sigma^{1/2}\pmb x$ gives whitened coordinates $\pmb x$ and the effective feature operator $\pmb T_h\pmb\Sigma^{1/2}$. Its spectrum therefore reflects both the SSM and the input covariance, and this effective operator need not retain the Toeplitz structure (see \hyperref[Appendix: relation]{Appendix~\ref*{Appendix: relation}: \nameref*{Appendix: relation}}).
\begin{restatable}[Eigen-Decomposition of SSM-Induced Kernel]{theorem}{EigenDecomposition}
\label{Theorem: eigen}
Assume $\mathbb E_{\pmb u}[\pmb u\pmb u^\top]=\pmb I_L$. Then the SSM-induced kernel admits the finite expansion $K_{\mathrm{SSM}}(\pmb u,\pmb u') = \pmb u^\top\pmb V\pmb S^2\pmb V^\top\pmb u' = \sum_{\rho=1}^{L} \eta_\rho\phi_\rho(\pmb u)\phi_\rho(\pmb u')$,
where $\eta_\rho=s_\rho^2$ and $\phi_\rho(\pmb u)=\pmb v_\rho^\top\pmb u$. These eigenfunctions satisfy $\mathbb E_{\pmb u}
[\phi_\rho(\pmb u)\phi_\sigma(\pmb u)] =\delta_{\rho\sigma}$. Consequently, the kernel has rank at most $L$.
\end{restatable}
This result shows that the SSM-induced kernel favors the right singular directions of the SSM convolution operator. As discussed in Section~\ref{subsec: kr and rkhs}, modes associated with larger \(s_\rho\) become high-eigenvalue kernel modes and are therefore learned more rapidly in the small-sample regime. Thus, the inductive bias of an LTI SSM is encoded in the singular spectrum of its Toeplitz matrix. The next subsection connects this singular spectrum to the SSM frequency response.
\subsection{Frequency Response Reveals the Spectral Bias of SSMs}
Theorem~\ref{Theorem: eigen} shows that the eigenvalues of the SSM-induced kernel are determined by the singular values of the Toeplitz convolution matrix \(\pmb T_h\). We now connect these singular values to the SSM frequency response. This step turns the algebraic characterization above into a frequency-domain interpretation of the model's inductive bias.

Recall that the impulse response $h=\{h_n\}_{n\ge 0}$ of an LTI SSM has frequency response $H(\omega) = \sum_{n=0}^{\infty} h_n e^{-i\omega n},
\omega\in(0,2\pi)$.
Intuitively, $|H(\omega)|$ measures how strongly the SSM amplifies an input component at frequency $\omega$. Therefore, if the singular values of $\pmb T_h$ are governed by $|H(\omega)|$, then the kernel eigenvalues in Theorem~\ref{Theorem: eigen} encode the frequency preference of the SSM.

For a finite sequence length $L$, let $\omega_j=2\pi j/L, j=0,\cdots,L-1$, denote the discrete Fourier frequencies. We use $\tilde s_j$ to denote the frequency-indexed singular values associated with $\omega_j$. Unlike the sorted singular values $s_\rho$ in Theorem~\ref{Theorem: eigen}, the quantities $\tilde s_j$ are indexed by frequency and are not sorted.

\begin{restatable}[Frequency response governs the SSM-induced spectrum]{theorem}{FrequencyResponseSpectrum}
\label{thm:frequency-response-spectrum}
For an SSM with absolutely summable impulse response $h$, i.e., $\sum_{n=0}^\infty |h_n|<\infty$, the singular values of the Toeplitz convolution operator $\pmb T_h$ are asymptotically governed by the SSM frequency response. In particular, as $L\to\infty$, $\tilde s_j\sim |H(\omega_j)|/\sqrt{L}, \omega_j=\frac{2\pi j}{L}$.
Equivalently, the eigenvalues of the SSM-induced kernel are governed by
$\tilde \eta_j   \sim|H(\omega_j)|^2 $ up to a scale factor determined by the kernel feature map normalization.
\end{restatable}
The proof is provided in \hyperref[Appendix: relation]{Appendix~\ref*{Appendix: relation}: \nameref*{Appendix: relation}}. The whitened-coordinate analysis above isolates the bias contributed by the SSM operator itself.  We also discuss generalized results for non-whitened inputs with non-trivial covariance.

Combining Theorem~\ref{Theorem: eigen} with Theorem~\ref{thm:frequency-response-spectrum}, we see that the frequency response determines which portions of the kernel spectrum receive large eigenvalues.
Frequency regions with large $|H(\omega)|^2$ are therefore favored by the SSM-induced kernel, while regions with small response require more samples to learn reliably.
This provides a concrete notion of spectral inductive bias for LTI SSMs.
The remaining question is how to estimate which frequency regions are relevant to a task and how to align the model response with them.
\subsection{Task-Dependent Initialization via Spectral Alignment}
\label{subsec: tdi}
The previous sections show that the spectral inductive bias of an LTI SSM is governed by its frequency response. Specifically, frequency regions with larger model spectrum $S_{\mathrm{model}}(\omega)=|H(\omega)|^2$
correspond to larger portions of the SSM-induced kernel spectrum and are therefore favored in the low-sample regime. This suggests a task-adaptive strategy: before downstream training, initialize the SSM so that its high-response frequency regions are aligned with frequency regions that are statistically predictive of the task. We refer to this principle as task-model spectral alignment.

To make this principle concrete and to quantify the task spectrum, we first introduce a task-relevant spectral statistic based on input-output cross-covariance. Given training samples $\{(\pmb u^\mu,\pmb Y^\mu)\}_{\mu=1}^{P}$, where $\pmb u^\mu\in\mathbb{R}^{L}$ and $\pmb Y^\mu\in\mathbb{R}^{d}$, we define
$\mathbf{C}_{\pmb u\pmb Y}=\frac{1}{P} \sum_{\mu=1}^{P}(\pmb u^\mu-\bar{\pmb u})(\pmb Y^\mu-\bar{\pmb Y})^\top \in \mathbb{R}^{L\times d}.$
For classification tasks, $\pmb Y^\mu$ is the one-hot label vector. We then compute the Fourier transform along the sequence dimension and aggregate across output dimensions:
    $S_{\mathrm{cross}}(\omega_j)=\frac{1}{d}\sum_{r=1}^{d} \left|\mathcal{F}\!\left[\mathbf{C}_{\pmb u\pmb Y}^{(:,r)}\right](\omega_j)\right|^2$, where    $\omega_j=\frac{2\pi j}{L}.$
The labels distinguish input variation associated with the prediction target from variation that merely carries large input power. Specifically, $S_{\mathrm{cross}}$ measures the spectral content of input variations that are linearly correlated with the output. It therefore provides an analyzable proxy for task-relevant spectrum, rather than a complete characterization of the task. We then define the spectral matching loss as the $L_2$ distance between the model spectrum and the task spectrum.
\begin{restatable}[Spectral Matching Loss]{definition}{SpectralMatchingLoss}
\label{def:spectral-matching-loss}
For nonzero model and task spectra $S_{\mathrm{model}}$ and $S_{\mathrm{task}}$, we define the spectral matching loss as
\begin{equation}
    \mathcal{L}_{\mathrm{spec}}
    =
    \left\|
    \frac{S_{\mathrm{model}}}
         {\|S_{\mathrm{model}}\|_2}
    -
    \frac{S_{\mathrm{task}}}
         {\|S_{\mathrm{task}}\|_2}
    \right\|_2^2 .
    \label{eq:spec-loss}
\end{equation}
\end{restatable}
The normalization makes the objective focus on the relative allocation of spectral power rather than the overall gain of the filter.
In the theoretical analysis below, we take $S_{\mathrm{task}}=S_{\mathrm{cross}}$ because the cross-spectrum admits a direct connection to model alignment. If $\pmb Y$ is independent of $\pmb u$, the population centered cross-covariance vanishes, and hence the population cross-spectrum is identically zero. In this case, the normalized loss is undefined, and the task statistic provides no preferred spectral alignment.  \hyperref[app:label-permutation]{Appendix~\ref*{app:label-permutation}: \nameref*{app:label-permutation}} examines the role of labels empirically by permuting them only during task-spectrum estimation.

We propose \emph{Task-Dependent Initialization (TDI)}, which aligns the initial SSM spectrum with the estimated task spectrum through peak initialization (see Implementation), optionally followed by gradient refinement of $\mathcal{L}_{\mathrm{spec}}$.
After this initialization step, the model is trained using the standard supervised task loss, with the same architecture, parameter count, and downstream training protocol as the baseline.
Thus, TDI should be interpreted not as an additional model component, but as a theory-motivated way to choose an initial spectral inductive bias. We provide a two-part theoretical justification of this method in \hyperref[Appendix: proof 2]{Appendix~\ref*{Appendix: proof 2}: \nameref*{Appendix: proof 2}}. Then, we obtain our final theorem.
\begin{restatable}[Spectral Matching Promotes Task-Model Spectral Alignment and Generalization]{theorem}{FINAL}
 \label{theorem: spectrum matching}
    Minimizing spectral matching loss $\mathcal{L}_\mathrm{spec}$ is equivalent to maximizing the normalized alignment objective $J/(\|S_{\mathrm{model}}\|_2\|S_{\mathrm{task}}\|_2)$ (Lemma~\ref{lemma: minimize spectrum loss}), which concentrates task-relevant projection power toward higher-eigenvalue kernel modes and thereby yields a faster-rising cumulative-power curve $C(\rho)$ (Lemma~\ref{lemma: maximize J}). Consequently, the resulting spectral alignment improves finite-sample generalization.
\end{restatable}
\paragraph{Implementation.}
TDI first estimates and normalizes a task-relevant spectrum from the available training data. For classification, Fisher log-power provides a practical alternative to the cross-spectrum used in our theory: we compute the log-power features $z(\omega)=\log(|\mathrm{rFFT}(\pmb u)(\omega)|^2+\epsilon)$ and score each frequency by the ratio of class-prior-weighted between-class variance to within-class variance plus a small regularizer. This emphasizes frequencies whose spectral energy distinguishes labels consistently (\hyperref[Appendix:fisher-log-power]{Appendix~\ref*{Appendix:fisher-log-power}: \nameref*{Appendix:fisher-log-power}}).

We then select dominant frequencies and map them to frequency-selective SSM modes (peak initialization). For a complex diagonal SSM, a selected frequency $f_j$, measured in cycles per input step, is encoded by a stable pole $A_j=-\gamma_j-i\,2\pi f_j/\Delta$, where $\gamma_j>0$ controls damping and $\Delta>0$ is the discretization step. Real-valued SSMs use corresponding stable $2\times2$ resonant blocks. Kernel regression and trainable one-layer SSM experiments further refine this peak initialization by minimizing $\mathcal{L}_{\mathrm{spec}}$. Deep SSM experiments instead use the constructed initialization directly, without gradient refinement, and apply TDI only to the first SSM layer. Downstream training otherwise follows the baseline protocol, and TDI introduces no inference-time overhead. Full algorithms and experimental settings are provided in \hyperref[Appendix:fisher-log-power]{Appendix~\ref*{Appendix: tdi-implementation}: \nameref*{Appendix: tdi-implementation}} and \hyperref[app:experimental_details]{Appendix~\ref*{app:experimental_details}: \nameref*{app:experimental_details}}.

We provide a bootstrap diagnostic for assessing the stability of the estimated task spectrum without model training (\hyperref[Appendix:fisher-log-power]{Appendix~\ref*{Appendix:fisher-log-power}: \nameref*{Appendix:fisher-log-power}}). The observed stability transitions inform an empirical screen for considering TDI, but do not guarantee positive downstream gains.
\section{Experiments}
\label{sec: experiments}
\subsection{Empirical Validation of the Kernel's Eigen-Decomposition}
\label{subsec: exp1}
Before evaluating TDI, we first validate the kernel-theoretic characterization underlying our method. Theorem~\ref{Theorem: eigen} predicts that, in whitened input coordinates, the SSM-induced kernel has eigenfunctions given by the right singular directions of the Toeplitz convolution operator, with eigenvalues $\eta_\rho=s_\rho^2$. 

\begin{figure}[htbp]
\centering
\subfigure{%
\begin{overpic}[width=0.25\textwidth]{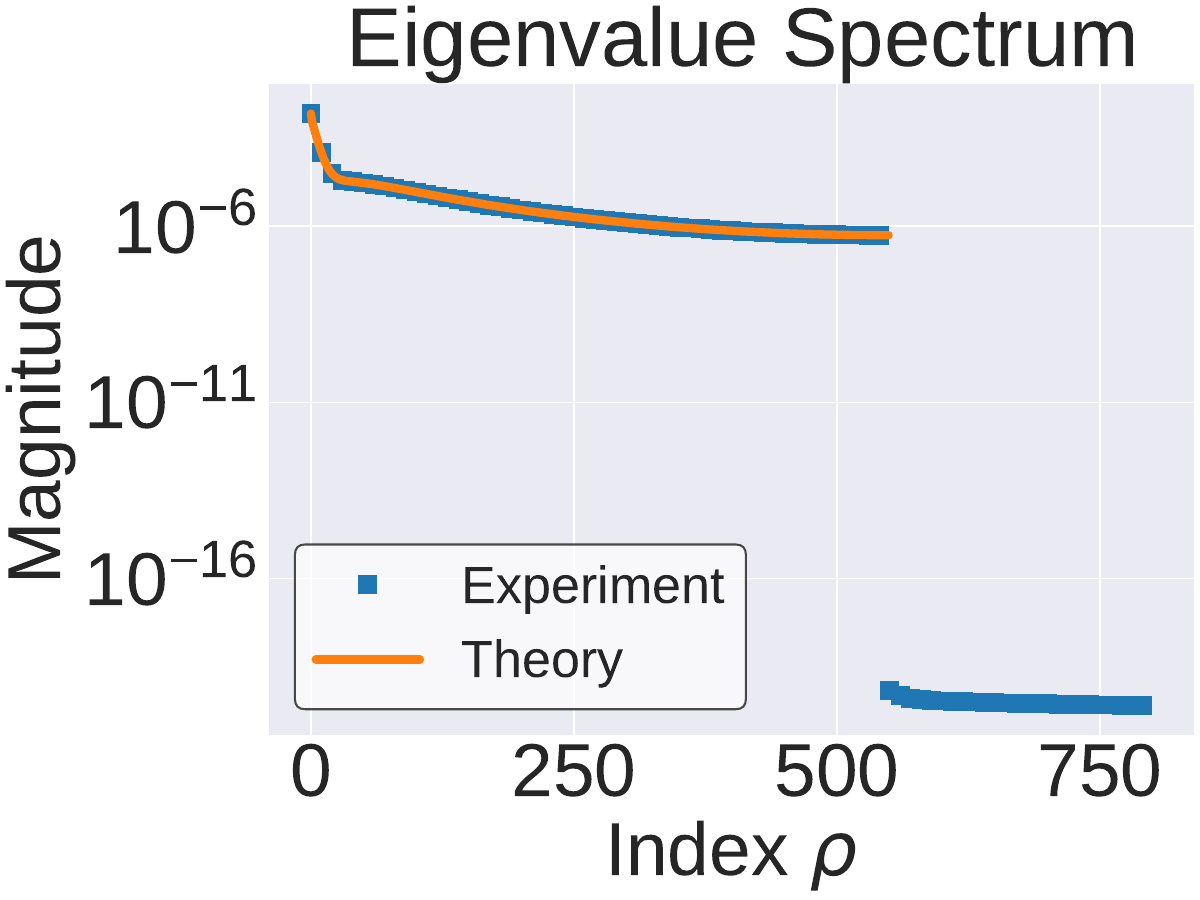}
  \put(1,80){\colorbox{white}{(a)}}
\end{overpic}%
}\hspace{-0.3em}%
\subfigure{%
\begin{overpic}[width=0.25\textwidth]{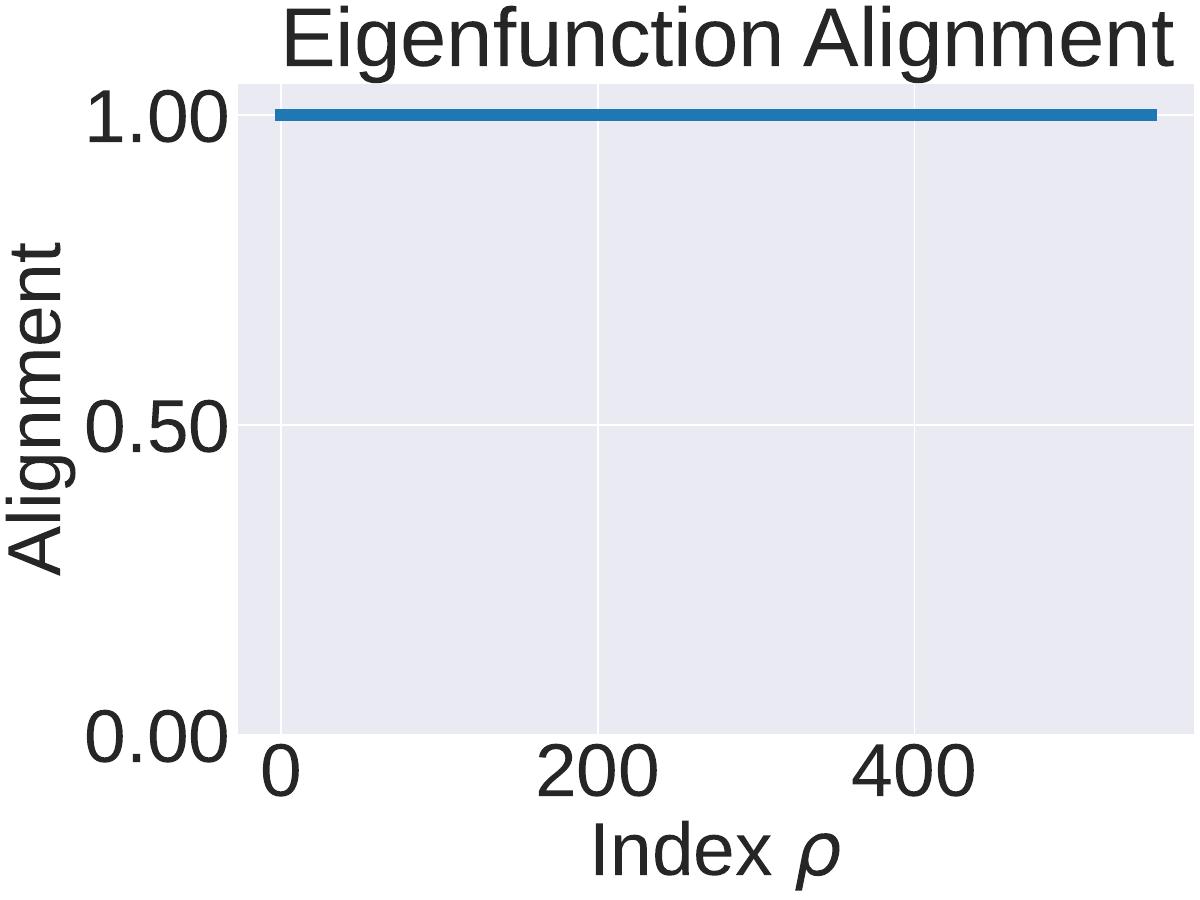}
  \put(1,80){\colorbox{white}{(b)}}
\end{overpic}%
}\hspace{-0.2em}%
\subfigure{%
\begin{overpic}[width=0.25\textwidth]{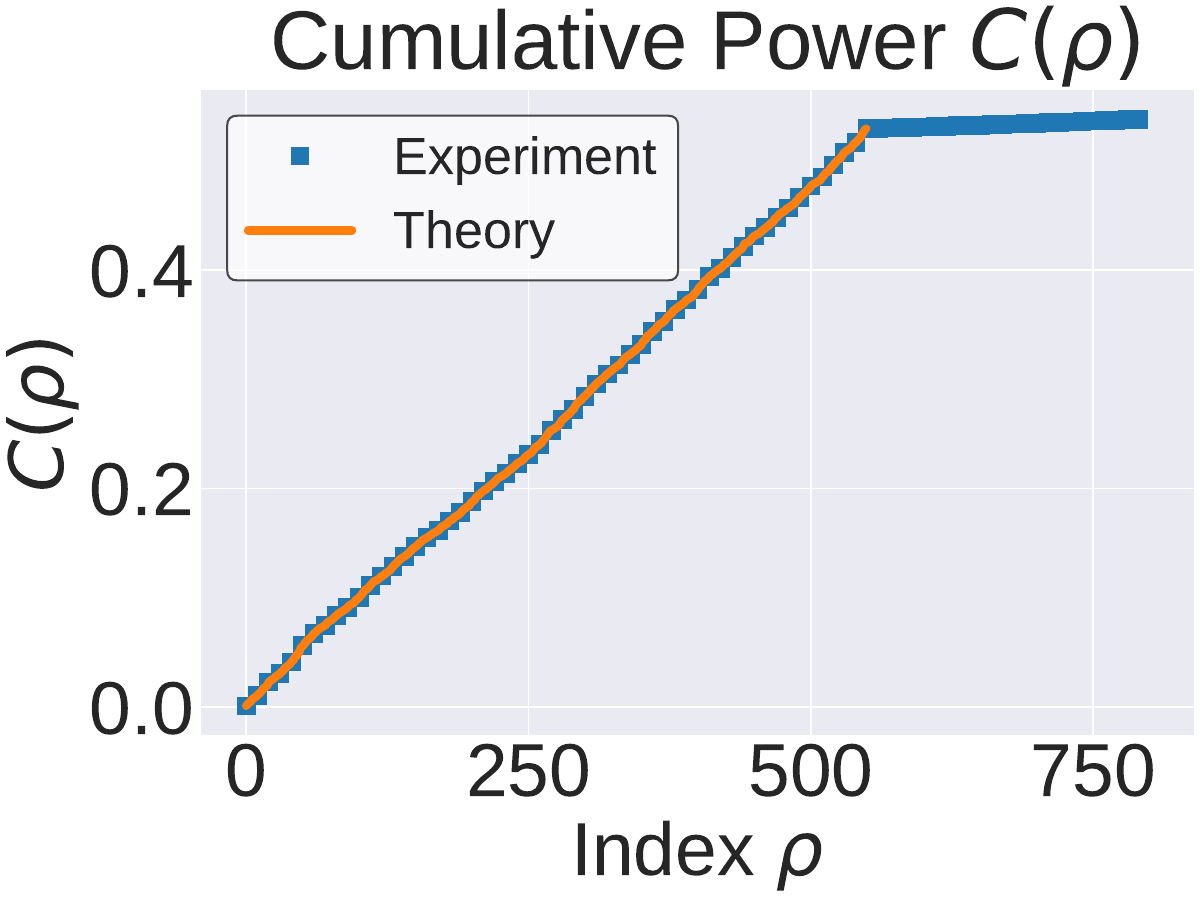}
  \put(1,80){\colorbox{white}{(c)}}
\end{overpic}%
}\hspace{-0.1em}%
\subfigure{%
\begin{overpic}[width=0.25\textwidth]{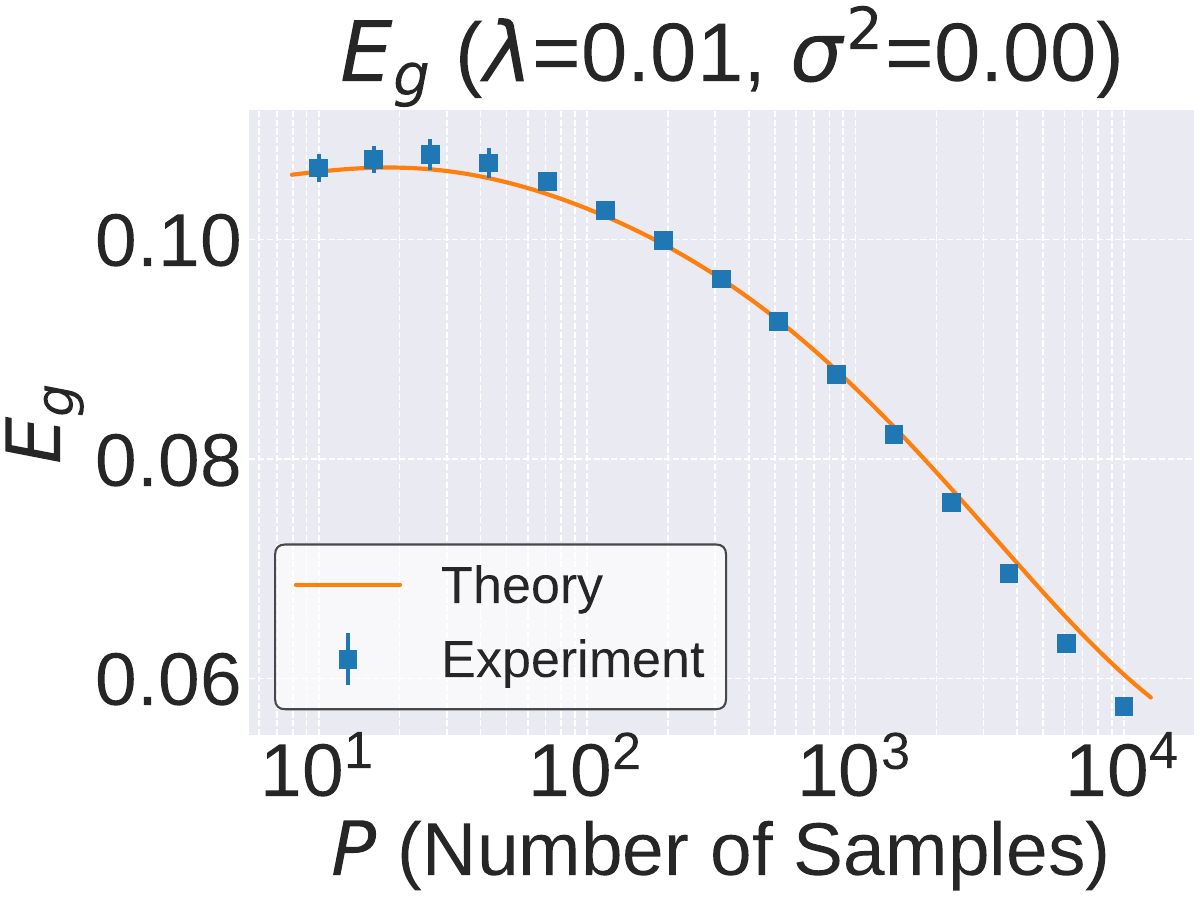}
  \put(1,80){\colorbox{white}{(d)}}
\end{overpic}%
}
\caption{\textbf{Sanity check of the SSM-induced kernel eigendecomposition.}
(a) The empirical eigenvalue spectrum matches the theoretical prediction
$\eta_\rho=s_\rho^2$ obtained from the singular values of the Toeplitz convolution operator.
(b) The empirical and predicted eigenfunctions are well aligned across most modes.
(c) The predicted eigenbasis accurately captures the cumulative task power $C(\rho)$.
(d) Kernel regression generalization predicted from the eigenspectrum matches empirical KRR performance across sample sizes. Error bars indicate standard
deviations over 50 independent runs.
}
\label{fig:sanity}
\end{figure}

On MNIST~\citep{MNIST}, the empirical spectrum and eigenfunctions closely match this prediction (Figure~\ref{fig:sanity}(a,b)). The induced expansion also accurately captures the cumulative power $C(\rho)$ (Figure~\ref{fig:sanity}(c)). Finally, eigenspectrum-based generalization predictions align with empirical kernel ridge regression across sample sizes, up to a small residual from the empirical tail (Figure~\ref{fig:sanity}(d)).
These results provide a sanity check that the SSM-induced kernel analysis captures the relevant spectral structure used in the later TDI experiments. We include additional diagnostics, including unwhitened-input results and the small empirical tail beyond the theoretical rank, in \hyperref[app:exp1_details]{Appendix~\ref*{app:exp1_details}: \nameref*{app:exp1_details}}.

\subsection{Kernel Regression: TDI Corrects Spectral Mismatch}
\label{subsec: exp2}
We next test the mechanism predicted by our theory in a controlled kernel regression setting. This experiment asks whether modifying the SSM spectrum according to the task spectrum changes the generalization behavior of the model.

We fix the initial model to be a low-frequency SSM and compare two synthetic tasks with different spectral structure (see \hyperref[appendix: exp2 details]{Appendix~\ref*{appendix: exp2 details}: \nameref*{appendix: exp2 details}} for details). In the matched case, the task is also low-frequency, so the initial model spectrum already substantially overlaps with the task-relevant frequencies. In the mismatched case, the task is high-frequency, so the task-relevant frequencies are weakly represented by the initial model. This comparison isolates the effect of spectral mismatch while keeping the model class, initialization method, and kernel regression protocol fixed.

\begin{figure}[htbp]
\centering
\subfigure{%
\begin{overpic}[width=0.3\textwidth]{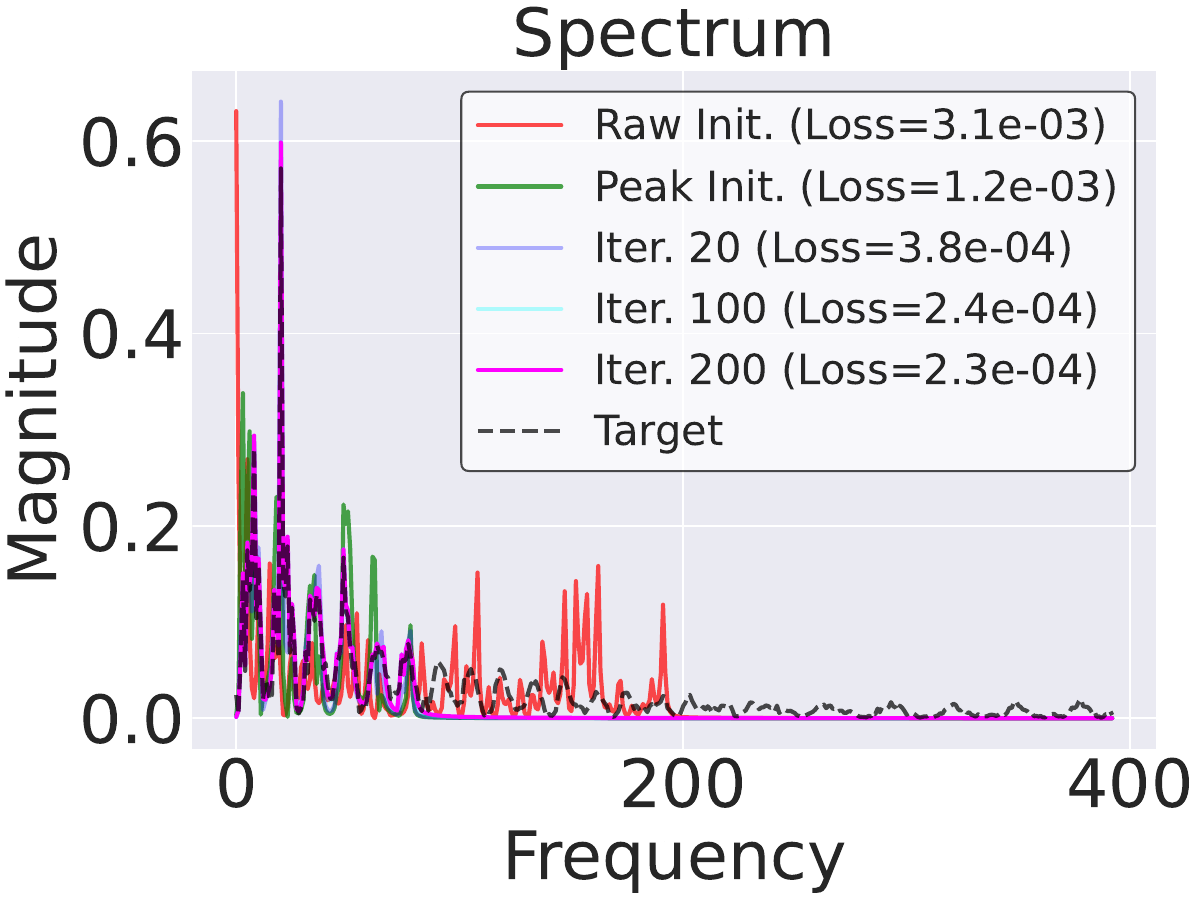}
  \put(1,80){\colorbox{white}{(a)}}
\end{overpic}%
}
\subfigure{%
\begin{overpic}[width=0.3\textwidth]{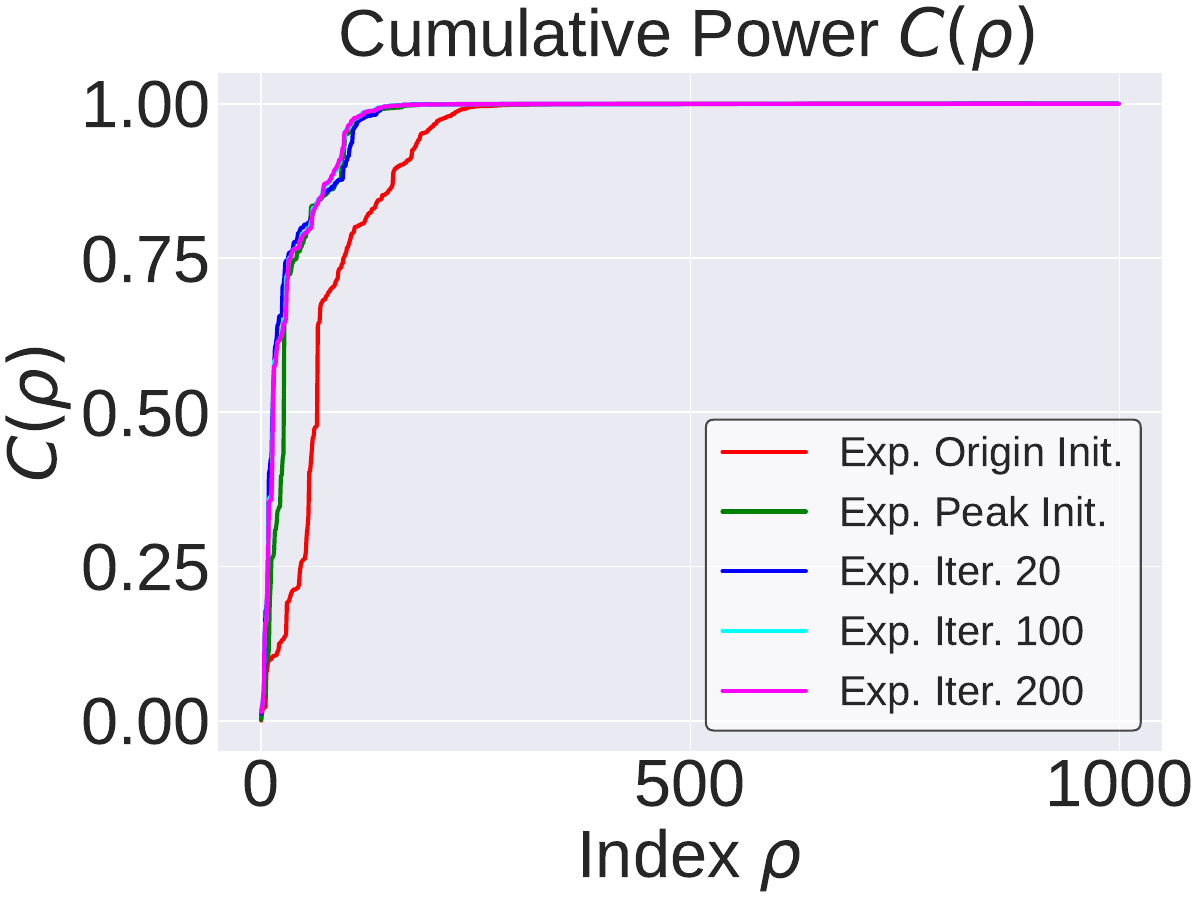}
  \put(1,80){\colorbox{white}{(b)}}
\end{overpic}%
}
\subfigure{%
\begin{overpic}[width=0.3\textwidth]{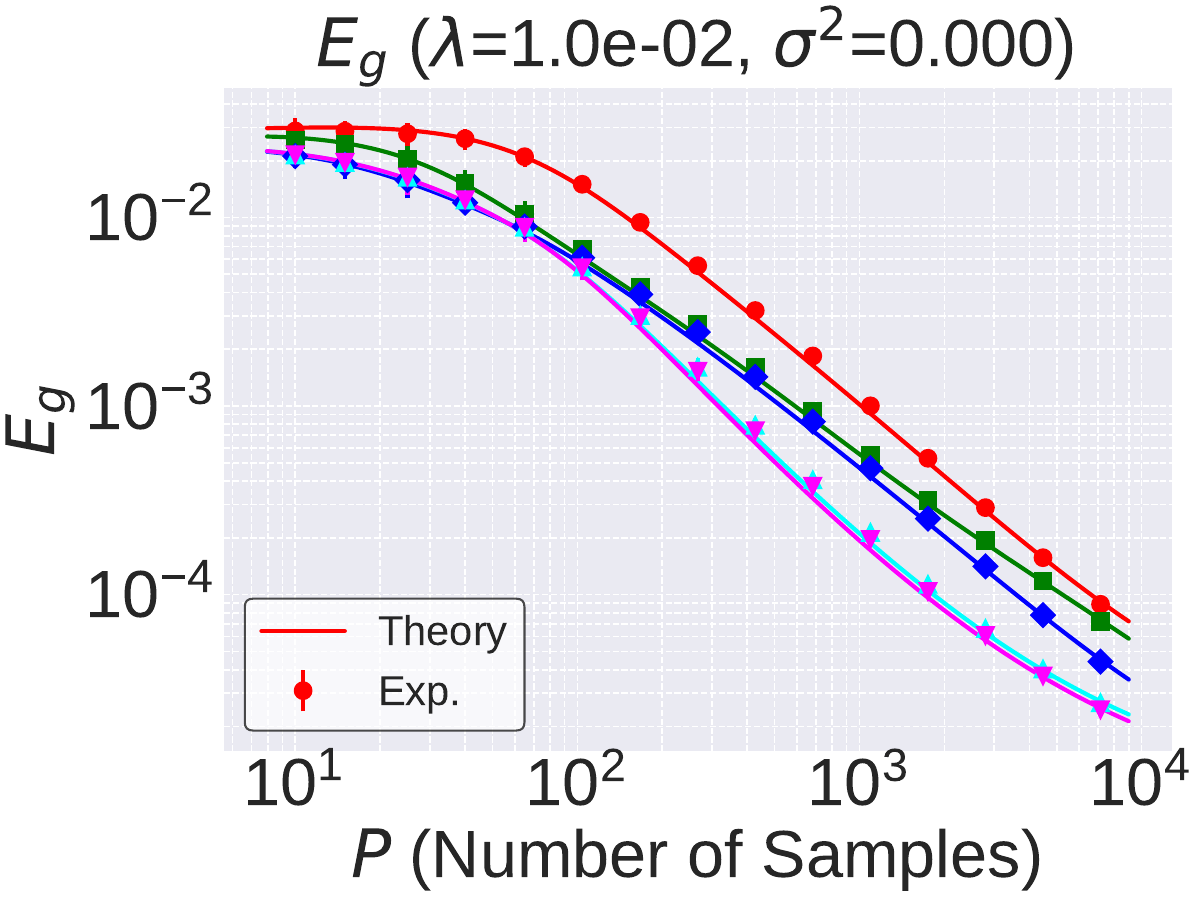}
  \put(1,80){\colorbox{white}{(c)}}
\end{overpic}%
}\\
\subfigure{%
\begin{overpic}[width=0.3\textwidth]{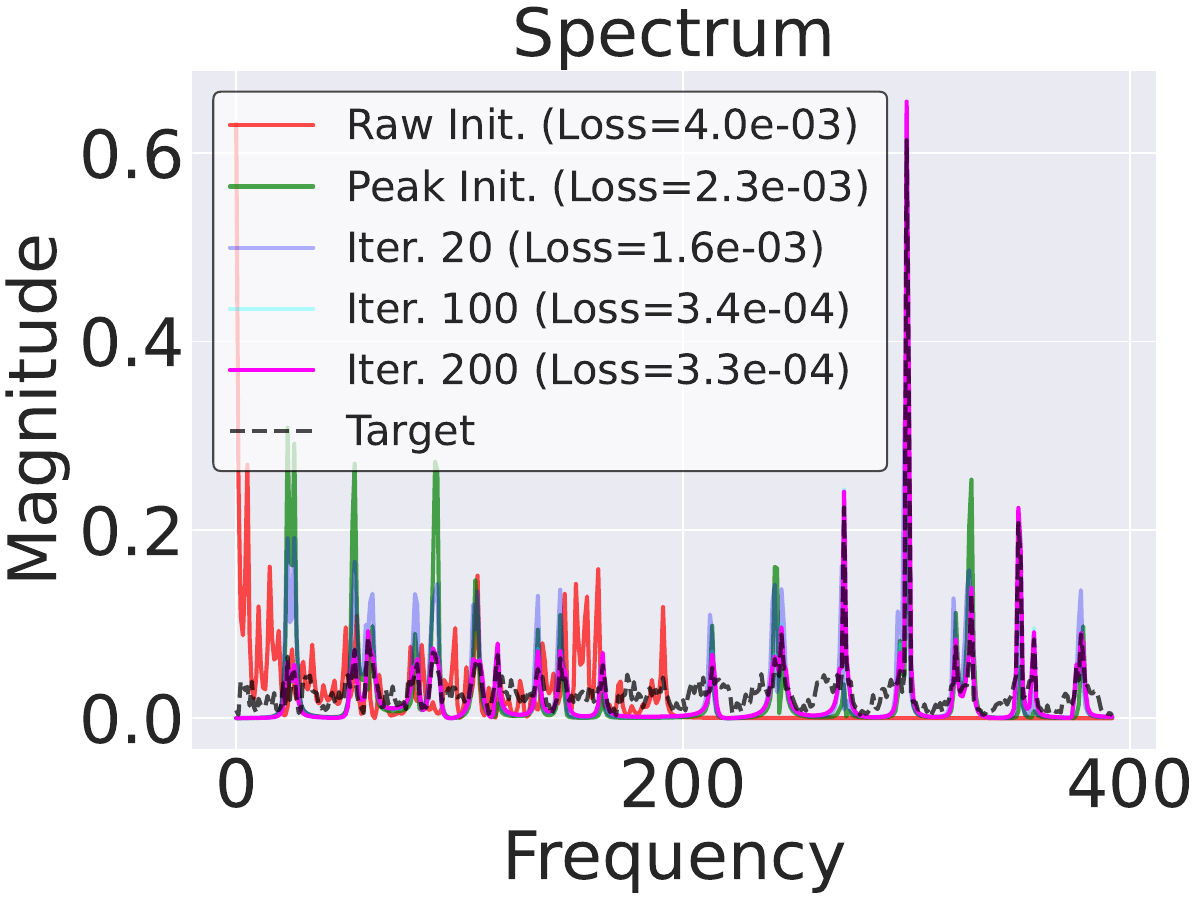}
  \put(1,80){\colorbox{white}{(d)}}
\end{overpic}%
}
\subfigure{%
\begin{overpic}[width=0.3\textwidth]{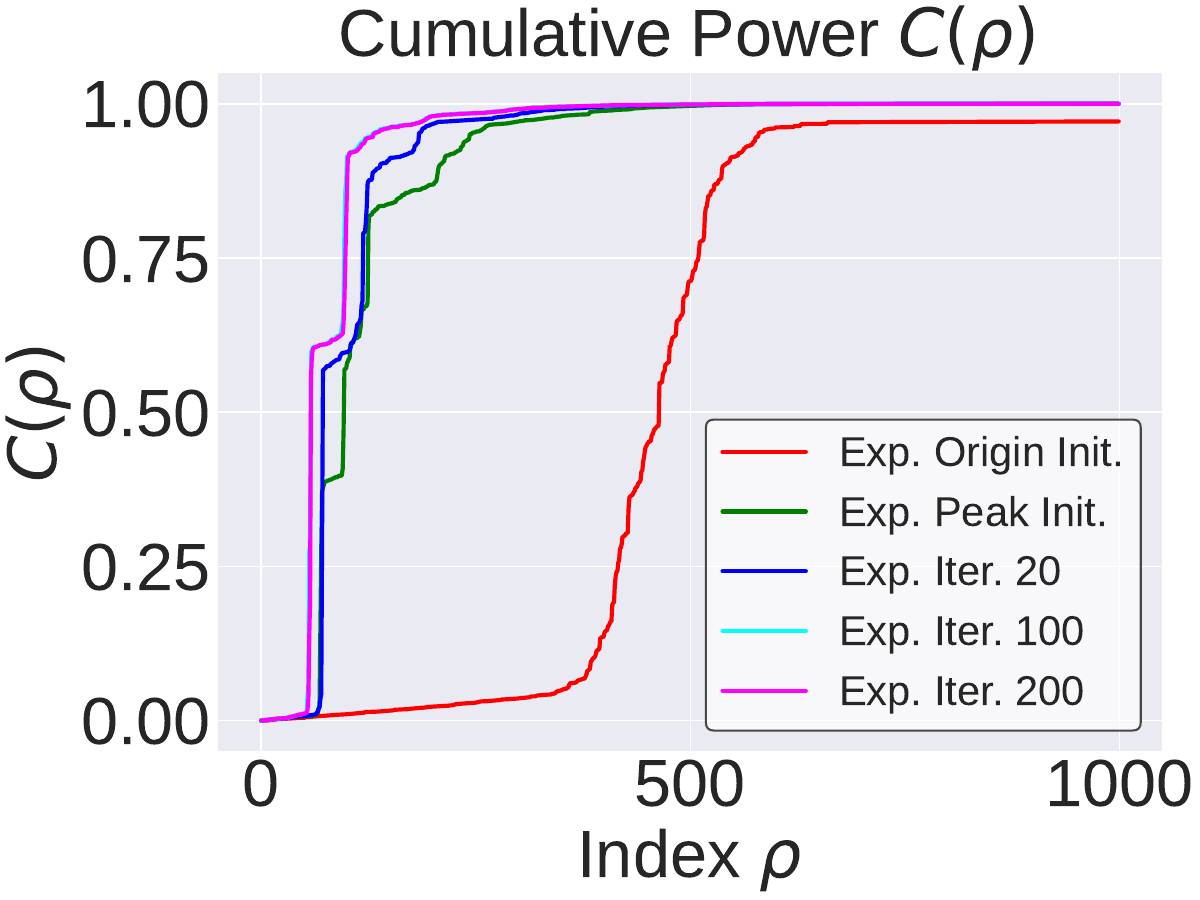}
  \put(1,80){\colorbox{white}{(e)}}
\end{overpic}%
}
\subfigure{%
\begin{overpic}[width=0.3\textwidth]{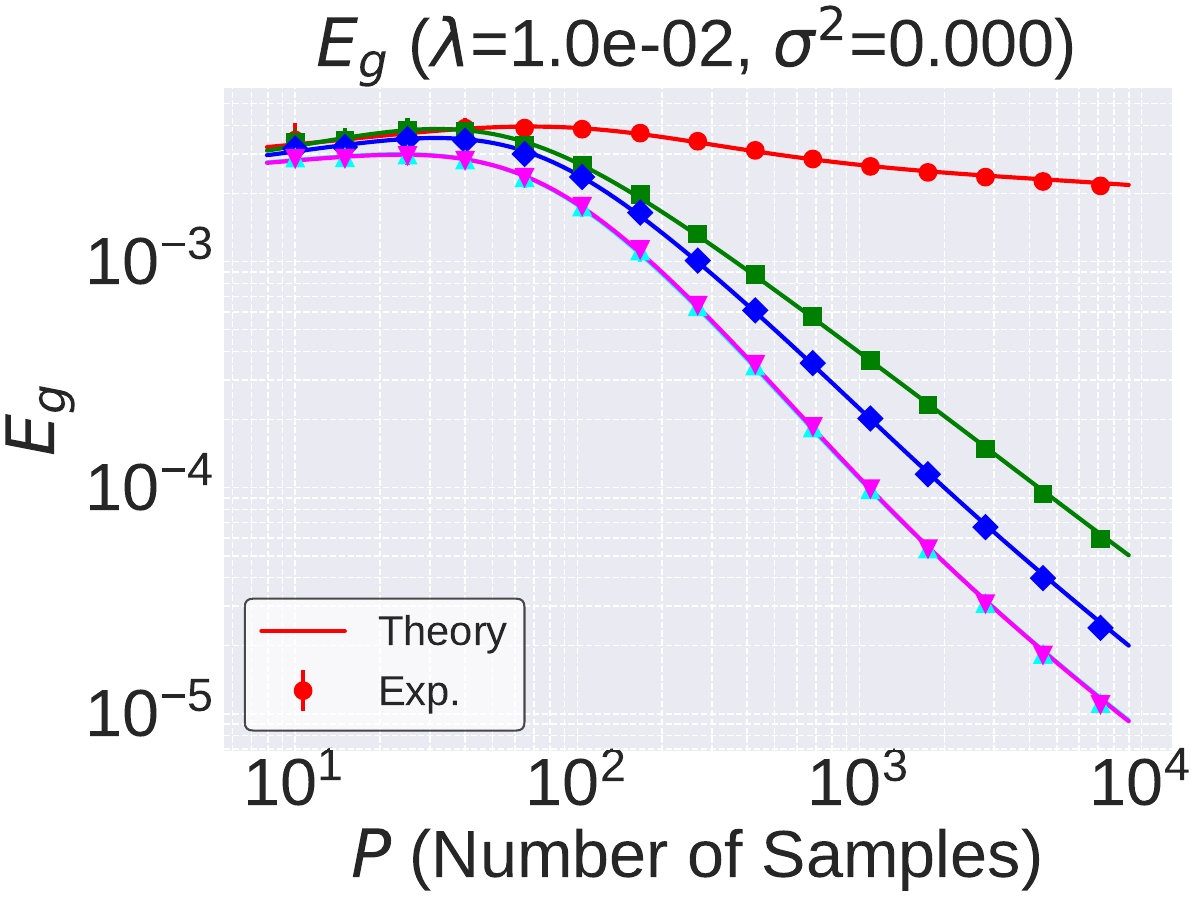}
  \put(1,80){\colorbox{white}{(f)}}
\end{overpic}%
}\hspace{-0.6em}%
\caption{\textbf{Kernel regression mechanism experiment.} We fix the initial model to be a low-frequency SSM and compare a matched low-frequency task with a mismatched high-frequency task. (a)-(c) In the \textit{matched} case, the initial model spectrum already overlaps with the task spectrum, so TDI produces little change in the model spectrum, cumulative task power, or generalization. (d)-(f) In the \textit{mismatched} case, the task-relevant high-frequency components are weakly represented by the initial model. TDI shifts the model spectrum toward these frequencies, moves task power into earlier high-eigenvalue modes, and shifts the kernel regression generalization curve downward across the finite-sample range. This supports the view that TDI improves data efficiency by correcting spectral mismatch rather than by universally improving all initializations.
}\label{fig:kernel_regression}
\end{figure}

For each task, we apply TDI by estimating the task spectrum and aligning the SSM frequency response with it (\hyperref[Appendix: tdi-implementation]{Appendix~\ref*{Appendix: tdi-implementation}: \nameref*{Appendix: tdi-implementation}}). We compare the original and TDI-aligned kernels using the model spectrum, cumulative power $C(\rho)$, and kernel regression generalization curve. Figure~\ref{fig:kernel_regression} shows that, in the matched case, TDI makes only minor changes because the initial spectrum is already close to the task spectrum; the cumulative-power and generalization curves therefore remain similar. In the mismatched case, however, the high-frequency task initially places much of its power in late, small-eigenvalue modes. TDI shifts the model spectrum toward these task-relevant frequencies, moves the target power earlier in the sorted kernel eigenbasis, and lowers the generalization curve across the finite-sample range. Additional parameter settings in \hyperref[appendix: exp2 details]{Appendix~\ref*{appendix: exp2 details}: \nameref*{appendix: exp2 details}} show the same qualitative trend. These results support the mechanism predicted by our analysis: TDI improves data efficiency by correcting spectral mismatch, effectively converting task-relevant directions from late-learned modes into early-learned modes of the SSM-induced kernel.
\subsection{Trainable One-Layer SSMs}
We next test whether the spectral alignment mechanism predicted by our theory continues to govern generalization when the recurrent dynamics are trainable. We use a simple one-layer linear SSM followed by a nonlinear readout. The SSM baseline and TDI model have the same architecture and number of trainable parameters; TDI only changes the initialization using the estimated task spectrum. We evaluate on seven sequence-classification tasks: a synthetic binary frequency-classification task, sequential MNIST~\citep{MNIST}, and
five univariate time-series datasets from the UCR archive~\citep{UCR}:
ECG5000, FordA, FordB, TwoPatterns, and Wafer. For each dataset, we sweep the training ratio over
$\{0.01, 0.02, 0.04, 0.08, 0.16, 0.32, 0.64, 1.0\}$. A training ratio $r$ means that we train on an $r$ fraction of the original training split, while keeping the validation and test splits fixed. Throughout the experiments, ``low-data'' and ``high-data'' refer to lower and higher within-dataset training ratios, respectively, rather than comparable absolute sample counts or task difficulty across datasets. Full model structure, dataset statistics, and hyperparameters are provided in \hyperref[app:exp3_details]{Appendix~\ref*{app:exp3_details}: \nameref*{app:exp3_details}}. Unless otherwise stated, dataset-level summaries average over ratios within each regime after averaging seeds, while paired improvement tables use matched dataset-ratio-seed triples with cluster bootstrap confidence intervals. Full per-dataset, per-ratio comparisons among the SSM baseline, TDI, and the Standard RNN baseline are provided in \hyperref[app:one_layer_full_results]{Appendix~\ref*{app:one_layer_full_results}: \nameref*{app:one_layer_full_results}}.

Table~\ref{tab:dataset-low-final-summary} shows that TDI achieves higher mean final accuracy on all seven datasets in the low-data regime and reduces final loss on most of them. The largest gains appear on ECG5000, sequential MNIST, and Wafer. In the high-data regime, the trend is more mixed: TDI still helps on several datasets, but the gains are smaller and can reverse on some tasks. This supports our interpretation that TDI mainly improves finite-sample learning by correcting spectral mismatch, rather than acting as a universally better initialization.
\begin{table*}[htbp]
\centering
\caption{\textbf{Dataset-level low- and high-data summary for the one-layer SSM baseline and TDI.} Low-data averages ratios $\le 0.16$; high-data averages ratios $\ge0.32$. Each cell reports the mean across ratios within the corresponding regime, where each ratio is first averaged over 3 random seeds; the standard deviation is computed across these ratio-level means. Accuracy and loss are the final-epoch test metrics. The better value between baseline and TDI is bolded.}
\label{tab:dataset-low-final-summary}
\resizebox{\textwidth}{!}{%
\begin{tabular}{lcccc|cccc}
\toprule
& \multicolumn{4}{c}{Low-data regime} & \multicolumn{4}{c}{High-data regime} \\
\cmidrule(lr){2-5} \cmidrule(lr){6-9}
Dataset & Base Acc. & TDI Acc. & Base Loss & TDI Loss & Base Acc. & TDI Acc. & Base Loss & TDI Loss \\
\midrule
Binary Freq. & $86.5 {\scriptstyle \pm 9.0}$\% & \boldmath\textbf{$88.6 {\scriptstyle \pm 6.8}$\%}\unboldmath & $0.292 {\scriptstyle \pm 0.163}$ & \boldmath\textbf{$0.256 {\scriptstyle \pm 0.135}$}\unboldmath & \boldmath\textbf{$97.5 {\scriptstyle \pm 0.5}$\%}\unboldmath & $97.3 {\scriptstyle \pm 0.6}$\% & \boldmath\textbf{$0.070 {\scriptstyle \pm 0.012}$}\unboldmath & $0.076 {\scriptstyle \pm 0.014}$ \\
ECG5000 & $62.9 {\scriptstyle \pm 11.8}$\% & \boldmath\textbf{$72.8 {\scriptstyle \pm 14.0}$\%}\unboldmath & $1.158 {\scriptstyle \pm 0.196}$ & \boldmath\textbf{$0.981 {\scriptstyle \pm 0.615}$}\unboldmath & $90.2 {\scriptstyle \pm 1.6}$\% & \boldmath\textbf{$90.5 {\scriptstyle \pm 1.1}$\%}\unboldmath & $0.375 {\scriptstyle \pm 0.065}$ & \boldmath\textbf{$0.370 {\scriptstyle \pm 0.053}$}\unboldmath \\
FordA & $52.2 {\scriptstyle \pm 3.5}$\% & \boldmath\textbf{$56.8 {\scriptstyle \pm 7.7}$\%}\unboldmath & \boldmath\textbf{$0.689 {\scriptstyle \pm 0.016}$}\unboldmath & $0.837 {\scriptstyle \pm 0.413}$ & $68.9 {\scriptstyle \pm 5.6}$\% & \boldmath\textbf{$70.4 {\scriptstyle \pm 5.1}$\%}\unboldmath & $0.569 {\scriptstyle \pm 0.065}$ & \boldmath\textbf{$0.548 {\scriptstyle \pm 0.052}$}\unboldmath \\
FordB & $53.5 {\scriptstyle \pm 3.5}$\% & \boldmath\textbf{$55.0 {\scriptstyle \pm 3.9}$\%}\unboldmath & \boldmath\textbf{$0.729 {\scriptstyle \pm 0.054}$}\unboldmath & $0.763 {\scriptstyle \pm 0.138}$ & $60.6 {\scriptstyle \pm 2.7}$\% & \boldmath\textbf{$61.7 {\scriptstyle \pm 2.8}$\%}\unboldmath & $0.669 {\scriptstyle \pm 0.039}$ & \boldmath\textbf{$0.640 {\scriptstyle \pm 0.014}$}\unboldmath \\
sMNIST & $61.1 {\scriptstyle \pm 13.3}$\% & \boldmath\textbf{$78.4 {\scriptstyle \pm 11.5}$\%}\unboldmath & $1.144 {\scriptstyle \pm 0.378}$ & \boldmath\textbf{$0.696 {\scriptstyle \pm 0.353}$}\unboldmath & $79.3 {\scriptstyle \pm 26.5}$\% & \boldmath\textbf{$88.2 {\scriptstyle \pm 6.2}$\%}\unboldmath & $0.375 {\scriptstyle \pm 0.144}$ & \boldmath\textbf{$0.364 {\scriptstyle \pm 0.175}$}\unboldmath \\
TwoPatterns & $28.7 {\scriptstyle \pm 4.3}$\% & \boldmath\textbf{$30.8 {\scriptstyle \pm 6.3}$\%}\unboldmath & $1.609 {\scriptstyle \pm 0.778}$ & \boldmath\textbf{$1.479 {\scriptstyle \pm 0.136}$}\unboldmath & \boldmath\textbf{$67.5 {\scriptstyle \pm 17.5}$\%}\unboldmath & $52.6 {\scriptstyle \pm 13.5}$\% & \boldmath\textbf{$0.788 {\scriptstyle \pm 0.328}$}\unboldmath & $1.021 {\scriptstyle \pm 0.258}$ \\
Wafer & $79.8 {\scriptstyle \pm 9.5}$\% & \boldmath\textbf{$87.4 {\scriptstyle \pm 6.2}$\%}\unboldmath & $0.486 {\scriptstyle \pm 0.111}$ & \boldmath\textbf{$0.323 {\scriptstyle \pm 0.115}$}\unboldmath & $94.9 {\scriptstyle \pm 2.9}$\% & \boldmath\textbf{$95.5 {\scriptstyle \pm 3.3}$\%}\unboldmath & $0.129 {\scriptstyle \pm 0.071}$ & \boldmath\textbf{$0.125 {\scriptstyle \pm 0.096}$}\unboldmath \\
\bottomrule
\end{tabular}%
}
\end{table*}
Table~\ref{tab:paired-accuracy-improvements} further confirms this trend with paired comparisons over matched dataset-ratio-seed triples. Final accuracy gains are positive from $1\%$ to $32\%$ training data, but weaken at $64\%$ and $100\%$. Mean dataset-level low-data gains are positive across all datasets, although the confidence intervals include zero for Binary Frequency Classification and FordB; early-epoch improvements are more dataset-dependent. Thus, TDI is not merely a fixed-kernel artifact: task-aligned spectral initialization can also improve data efficiency in trainable one-layer SSMs without adding parameters or changing the architecture.
\begin{table*}[htbp]
\centering
\caption{\textbf{Paired accuracy improvements of TDI over the one-layer SSM baseline.} Each paired point matches the same dataset, ratio, and seed. Final uses the final test accuracy; Early-5 and Early-10 use the mean test accuracy over the first 5 and 10 epochs. All entries are percentage-point improvements with cluster bootstrap 95\% confidence intervals. (a) Ratio-level improvements with datasets as bootstrap clusters. (b) Dataset-level low-data improvements with training ratios as bootstrap clusters. In each cluster, values are first averaged over 3 random seeds.}
\label{tab:paired-accuracy-improvements}
\begin{minipage}[t]{0.47\textwidth}
\centering
\textbf{(a) Ratio-level improvement}\\[0.5ex]
\resizebox{\linewidth}{!}{%
\begin{tabular}{lccc}
\toprule
Ratio & Final $\Delta$Acc. & Early-5 $\Delta$Acc. & Early-10 $\Delta$Acc. \\
\midrule
0.01 & +5.4 {\scriptsize [+1.8, +10.3]} & +3.5 {\scriptsize [-0.7, +9.7]} & +3.2 {\scriptsize [+0.8, +6.1]} \\
0.02 & +7.5 {\scriptsize [+0.9, +15.1]} & -0.2 {\scriptsize [-2.7, +3.1]} & +2.1 {\scriptsize [-0.4, +5.1]} \\
0.04 & +6.5 {\scriptsize [+2.3, +11.2]} & +3.2 {\scriptsize [-1.5, +9.0]} & +3.6 {\scriptsize [+1.2, +6.8]} \\
0.08 & +7.3 {\scriptsize [+4.0, +10.8]} & +2.7 {\scriptsize [+0.4, +5.3]} & +4.7 {\scriptsize [+3.2, +6.1]} \\
0.16 & +5.6 {\scriptsize [+1.5, +10.0]} & +1.7 {\scriptsize [-1.2, +4.6]} & +2.7 {\scriptsize [-0.1, +5.7]} \\
0.32 & +2.3 {\scriptsize [+0.7, +4.0]} & +1.5 {\scriptsize [+0.0, +3.1]} & +1.3 {\scriptsize [-0.1, +2.9]} \\
0.64 & -1.2 {\scriptsize [-6.7, +2.3]} & -1.2 {\scriptsize [-4.3, +1.6]} & -0.6 {\scriptsize [-2.2, +1.0]} \\
1.00 & -2.3 {\scriptsize [-14.1, +7.3]} & -0.6 {\scriptsize [-4.4, +3.0]} & -1.6 {\scriptsize [-7.8, +3.3]} \\
\bottomrule
\end{tabular}%
}
\end{minipage}\hfill
\begin{minipage}[t]{0.50\textwidth}
\centering
\textbf{(b) Dataset-level low-data improvement}\\[0.5ex]
\resizebox{\linewidth}{!}{%
\begin{tabular}{lccc}
\toprule
Dataset & Final $\Delta$Acc. & Early-5 $\Delta$Acc. & Early-10 $\Delta$Acc. \\
\midrule
Binary Freq. & +2.2 {\scriptsize [-0.3, +4.5]} & +3.3 {\scriptsize [+0.3, +6.4]} & +2.4 {\scriptsize [-0.0, +4.8]} \\
ECG5000 & +9.9 {\scriptsize [+4.0, +16.9]} & +6.9 {\scriptsize [-2.1, +16.0]} & +6.1 {\scriptsize [+1.2, +10.7]} \\
FordA & +4.7 {\scriptsize [+0.7, +9.4]} & +1.0 {\scriptsize [-0.8, +3.7]} & +2.5 {\scriptsize [+0.1, +6.1]} \\
FordB & +1.5 {\scriptsize [-1.1, +4.7]} & +0.2 {\scriptsize [-0.8, +1.0]} & +0.4 {\scriptsize [-0.7, +1.6]} \\
sMNIST & +17.3 {\scriptsize [+15.7, +19.4]} & -4.1 {\scriptsize [-5.6, -2.5]} & +2.4 {\scriptsize [+0.3, +4.8]} \\
TwoPatterns & +2.0 {\scriptsize [+0.1, +4.0]} & +3.3 {\scriptsize [+1.8, +5.0]} & +3.0 {\scriptsize [+1.4, +4.8]} \\
Wafer & +7.6 {\scriptsize [+4.8, +10.2]} & +4.7 {\scriptsize [+2.7, +6.7]} & +6.2 {\scriptsize [+5.0, +7.3]} \\
\bottomrule
\end{tabular}%
}
\end{minipage}
\end{table*}
\subsection{Deep SSMs}
\label{subsec: exp4}
We finally evaluate whether the spectral alignment mechanism underlying TDI continues to hold in deep SSMs. We compare the S4D~\citep{gu2022parameterizationinitializationdiagonalstate} baseline with TDI using the same architecture, downstream training protocol, and number of trainable parameters. TDI is applied only to the first S4 layer, while all later layers are initialized and trained as in the baseline (\hyperref[Appendix: tdi-implementation]{Appendix~\ref*{Appendix: tdi-implementation}: \nameref*{Appendix: tdi-implementation}}). We evaluate on eight sequence benchmarks: synthetic Frequency Classification, sequential and permuted MNIST~\citep{MNIST}, PathFinder~\citep{tay2021long}, CIFAR-10~\citep{krizhevsky2009learning}, ListOps~\citep{nangia2018listops,tay2021long}, Speech Commands, and SC10~\citep{warden2018speech}. The training ratios are logarithmically spaced between $10^{-3}$ and $1$. Other experimental details are provided in \hyperref[app:deep_ssm_details]{Appendix~\ref*{app:deep_ssm_details}: \nameref*{app:deep_ssm_details}}. Full per-dataset and per-ratio results, including the TDI-Frozen variant where the TDI-initialized first S4 layer is kept fixed during supervised training, are provided in \hyperref[app:deep-ssm-full-results]{Appendix~\ref*{app:deep-ssm-full-results}: \nameref*{app:deep-ssm-full-results}}.

Table~\ref{tab:paper-dataset-summary} summarizes final-epoch accuracy and loss in low- and high-data regimes. In the low-data regime, TDI achieves higher mean final accuracy on most datasets, although the paired confidence intervals include zero for several datasets (Table~\ref{tab:paper-improvement-efficiency}). The trend is more mixed in the high-data regime, where some datasets still benefit but others match or underperform the baseline. This is consistent with our interpretation that TDI primarily improves data-efficient learning by correcting spectral mismatch, rather than acting as a universally better initialization. Table~\ref{tab:paper-improvement-efficiency} provides paired comparisons over matched dataset-ratio-seed triples. Final accuracy gains are positive across much of the low-data range, but are small or uncertain at the extremely small ratios, where the task spectrum is difficult to estimate reliably. The gains also weaken at larger training ratios. Early-5 and Early-10 improvements are positive on most datasets with clearer spectral structure, suggesting that TDI often improves the initial learning trajectory. However, gains are limited on ListOps and PathFinder, whose labels may be less directly characterized by a one-dimensional input frequency spectrum. This negative evidence is consistent with the limitation of using a one-dimensional spectral statistic for tasks whose label structure is more compositional or spatial. Overall, the deep SSM results support the same conditional claim as the previous experiments: TDI is most useful in finite-sample regimes where task-relevant spectral structure is present and the default SSM initialization is not already well aligned with it.
\begin{table*}[htbp]
\centering
\small
\caption{\textbf{Dataset-level low- and high-data summary for the deep SSM baseline and TDI.} Low-data averages ratios $\leq 0.100$; high-data averages ratios $\geq 0.215$. Each cell reports the mean across ratios within the corresponding regime, where each ratio is first averaged over 5 random seeds; the standard deviation is computed across these ratio-level means. Accuracy and loss are final-epoch test metrics. The better value between baseline and TDI is bolded.}
\label{tab:paper-dataset-summary}
\resizebox{\textwidth}{!}{%
\begin{tabular}{lcccc|cccc}
\toprule
 & \multicolumn{4}{c}{Low-data regime} & \multicolumn{4}{c}{High-data regime} \\
\cmidrule(lr){2-5}\cmidrule(lr){6-9}
Dataset & Base Acc. & TDI Acc. & Base Loss & TDI Loss & Base Acc. & TDI Acc. & Base Loss & TDI Loss \\
\midrule
CIFAR-10 & 33.9{\scriptsize $\pm$15.7}\% & \textbf{35.4{\scriptsize $\pm$14.5}\%} & 2.026{\scriptsize $\pm$0.190} & \textbf{2.012{\scriptsize $\pm$0.184}} & \textbf{73.7{\scriptsize $\pm$7.2}\%} & 69.4{\scriptsize $\pm$8.5}\% & \textbf{0.948{\scriptsize $\pm$0.344}} & 1.094{\scriptsize $\pm$0.406} \\
Freq. Classification & 88.1{\scriptsize $\pm$18.9}\% & \textbf{91.1{\scriptsize $\pm$11.5}\%} & 0.385{\scriptsize $\pm$0.590} & \textbf{0.331{\scriptsize $\pm$0.439}} & \textbf{99.2{\scriptsize $\pm$0.4}\%} & 98.9{\scriptsize $\pm$0.5}\% & \textbf{0.024{\scriptsize $\pm$0.014}} & 0.026{\scriptsize $\pm$0.012} \\
ListOps & \textbf{16.6{\scriptsize $\pm$7.1}\%} & 16.2{\scriptsize $\pm$6.6}\% & \textbf{5.419{\scriptsize $\pm$0.754}} & 5.597{\scriptsize $\pm$0.941} & \textbf{44.7{\scriptsize $\pm$6.9}\%} & 43.2{\scriptsize $\pm$6.2}\% & 1.928{\scriptsize $\pm$0.786} & \textbf{1.816{\scriptsize $\pm$0.584}} \\
sMNIST & 56.3{\scriptsize $\pm$42.2}\% & \textbf{67.4{\scriptsize $\pm$38.1}\%} & 1.145{\scriptsize $\pm$1.084} & \textbf{0.897{\scriptsize $\pm$0.991}} & \textbf{99.1{\scriptsize $\pm$0.2}\%} & 99.0{\scriptsize $\pm$0.2}\% & \textbf{0.034{\scriptsize $\pm$0.011}} & 0.038{\scriptsize $\pm$0.011} \\
Pathfinder & 51.4{\scriptsize $\pm$2.4}\% & \textbf{51.7{\scriptsize $\pm$2.6}\%} & 3.482{\scriptsize $\pm$0.853} & \textbf{3.343{\scriptsize $\pm$0.864}} & 66.6{\scriptsize $\pm$10.4}\% & \textbf{68.4{\scriptsize $\pm$8.1}\%} & 0.935{\scriptsize $\pm$0.567} & \textbf{0.891{\scriptsize $\pm$0.407}} \\
pMNIST & 52.2{\scriptsize $\pm$38.6}\% & \textbf{52.6{\scriptsize $\pm$39.1}\%} & 1.294{\scriptsize $\pm$0.945} & \textbf{1.292{\scriptsize $\pm$0.971}} & \textbf{96.9{\scriptsize $\pm$0.7}\%} & 96.7{\scriptsize $\pm$1.0}\% & \textbf{0.121{\scriptsize $\pm$0.037}} & 0.130{\scriptsize $\pm$0.051} \\
Speech Commands & 51.8{\scriptsize $\pm$31.5}\% & \textbf{55.7{\scriptsize $\pm$29.7}\%} & 3.777{\scriptsize $\pm$3.810} & \textbf{3.716{\scriptsize $\pm$4.483}} & 87.3{\scriptsize $\pm$1.6}\% & \textbf{87.9{\scriptsize $\pm$2.2}\%} & 0.432{\scriptsize $\pm$0.051} & \textbf{0.415{\scriptsize $\pm$0.079}} \\
SC10 & 49.2{\scriptsize $\pm$31.8}\% & \textbf{52.8{\scriptsize $\pm$31.9}\%} & \textbf{6.003{\scriptsize $\pm$7.895}} & 6.492{\scriptsize $\pm$10.229} & 92.4{\scriptsize $\pm$1.1}\% & \textbf{92.7{\scriptsize $\pm$0.7}\%} & 0.246{\scriptsize $\pm$0.041} & \textbf{0.242{\scriptsize $\pm$0.035}} \\
\bottomrule
\end{tabular}%
}
\end{table*}
\begin{table*}[htbp]
\centering
\small
\caption{\textbf{Paired accuracy improvements of TDI over the deep SSM baseline.} Each paired point matches the same dataset, training ratio, and seed. Final uses final-epoch test accuracy; Early-5 and Early-10 use mean test accuracy over the first 5 and 10 epochs. All entries are percentage-point improvements with cluster bootstrap 95\% confidence intervals. (a) Ratio-level improvements with datasets as bootstrap clusters. (b) Dataset-level low-data improvements with training ratios as bootstrap clusters. In each cluster, values are first averaged over 5 random seeds.}
\label{tab:paper-improvement-efficiency}
\begin{minipage}[t]{0.44\textwidth}
\centering
\textbf{(a) Ratio-level improvement}\\[0.35em]
\resizebox{\linewidth}{!}{%
\begin{tabular}{lrrr}
\toprule
Ratio & Final $\Delta$Acc &  Early-5 $\Delta$Acc. & Early-10 $\Delta$Acc. \\
\midrule
0.001 & +0.0 {\scriptsize [-0.3, +0.5]} & +0.2 {\scriptsize [+0.0, +0.5]} & +0.8 {\scriptsize [+0.0, +1.5]} \\
0.002 & +0.9 {\scriptsize [-0.5, +3.2]} & +0.8 {\scriptsize [+0.0, +1.6]} & +1.0 {\scriptsize [+0.0, +2.3]} \\
0.005 & +2.0 {\scriptsize [-0.0, +4.7]} & +0.9 {\scriptsize [-0.0, +1.9]} & +1.5 {\scriptsize [+0.1, +3.2]} \\
0.010 & +8.5 {\scriptsize [+2.6, +15.6]} & +5.2 {\scriptsize [+0.4, +13.2]} & +5.3 {\scriptsize [+0.7, +12.3]} \\
0.022 & +3.9 {\scriptsize [-0.7, +10.9]} & +6.0 {\scriptsize [+1.3, +11.7]} & +4.4 {\scriptsize [+1.2, +8.0]} \\
0.046 & +0.6 {\scriptsize [+0.2, +1.0]} & +3.8 {\scriptsize [+0.5, +7.4]} & +3.6 {\scriptsize [+0.4, +7.5]} \\
0.100 & -0.3 {\scriptsize [-1.2, +0.5]} & +5.5 {\scriptsize [-0.4, +13.4]} & +7.0 {\scriptsize [+0.3, +15.7]} \\
0.215 & -0.3 {\scriptsize [-1.9, +0.9]} & +2.4 {\scriptsize [-1.8, +5.8]} & +1.1 {\scriptsize [-1.8, +3.3]} \\
0.464 & -0.7 {\scriptsize [-2.2, +0.5]} & +0.8 {\scriptsize [-2.3, +4.0]} & -0.4 {\scriptsize [-2.6, +1.8]} \\
1.000 & -0.4 {\scriptsize [-1.2, +0.4]} & -1.8 {\scriptsize [-4.1, -0.1]} & -1.6 {\scriptsize [-3.8, +0.1]} \\
\bottomrule
\end{tabular}%
}
\end{minipage}\hfill
\begin{minipage}[t]{0.54\textwidth}
\centering
\textbf{(b) Dataset-level low-data improvement}\\[0.35em]
\resizebox{\linewidth}{!}{%
\begin{tabular}{lrrr}
\toprule
Dataset & Final $\Delta$Acc &  Early-5 $\Delta$Acc. & Early-10 $\Delta$Acc.  \\
\midrule
CIFAR-10 & +1.5 {\scriptsize [-0.8, +4.1]} & +0.9 {\scriptsize [+0.3, +2.0]} & +1.4 {\scriptsize [+0.5, +2.7]} \\
Freq. Classification & +3.0 {\scriptsize [-2.6, +11.6]} & +14.7 {\scriptsize [+3.4, +26.0]} & +9.6 {\scriptsize [+1.5, +21.9]} \\
ListOps & -0.4 {\scriptsize [-1.1, +0.5]} & -0.6 {\scriptsize [-2.7, +0.9]} & -0.4 {\scriptsize [-1.8, +0.8]} \\
sMNIST & +11.1 {\scriptsize [+0.1, +22.2]} & +6.3 {\scriptsize [-0.0, +18.2]} & +10.0 {\scriptsize [+0.3, +23.3]} \\
Pathfinder & +0.2 {\scriptsize [-0.0, +0.5]} & -0.1 {\scriptsize [-0.1, -0.1]} & -0.1 {\scriptsize [-0.2, -0.0]} \\
pMNIST & +0.3 {\scriptsize [-0.2, +0.7]} & +1.4 {\scriptsize [-0.4, +4.7]} & +3.0 {\scriptsize [-0.9, +9.6]} \\
Speech Commands & +2.6 {\scriptsize [+0.5, +5.2]} & +6.2 {\scriptsize [+2.5, +10.5]} & +5.5 {\scriptsize [+3.1, +8.4]} \\
SC10 & +3.7 {\scriptsize [+0.8, +8.2]} & +4.8 {\scriptsize [+1.9, +7.7]} & +4.8 {\scriptsize [+2.2, +7.6]} \\
\bottomrule
\end{tabular}%
}
\end{minipage}
\end{table*}
\paragraph{Early learning versus extended-training accuracy.}
We extend training from 30 to 100 epochs on sMNIST and from 10 to 50 epochs on Frequency Classification, using five matched seeds. On sMNIST with 2.15\% training data, TDI reduces the mean number of epochs required to reach 80\% test accuracy from 34.2 to 20.0, while the final accuracy difference is only $+0.12 \pm 0.47$ percentage points (mean $\pm$ SD). Frequency Classification similarly shows an early-learning advantage in the low-data setting, with small final accuracy differences after extended training. These results distinguish faster early learning from systematically higher accuracy after longer training (\hyperref[app:extended-training]{Appendix~\ref*{app:extended-training}: \nameref*{app:extended-training}}). As complementary evidence on early learning across the original training budgets, TDI achieves higher Early-10 accuracy in 45/70 deep SSM dataset–ratio conditions, and in 39/50 conditions among the six tasks (excluding PathFinder and ListOps) better matched to the current one-dimensional spectral statistic.

\section{Conclusion and Limitations}
We introduced a kernel-theoretic framework for understanding the spectral inductive bias of LTI State Space Models. By viewing an SSM as a convolutional feature map, we showed that the spectrum of the induced kernel is governed by the SSM frequency response, connecting frequency-domain bias to mode-wise learnability in the small-sample regime. This analysis motivated Task-Dependent Initialization (TDI), a lightweight method that aligns the initial SSM spectrum with task-relevant spectral structure before training. Across controlled kernel regression, trainable one-layer SSMs, and deep SSMs, TDI is most beneficial when data are limited, label-relevant spectral structure is estimable, and the default initialization is spectrally mismatched. Improvements in early learning and accuracy under fixed training budgets should be distinguished from higher accuracy after extended training: on sMNIST and Frequency Classification, the baseline largely closes the gap with longer training.

This work also has several limitations. Our theory focuses on LTI SSMs, leaving extensions to selective architectures such as Mamba as an important direction. Under small, smooth input-dependent modulation around an LTI reference, selective dynamics may admit a local approximation consisting of the LTI operator and a first-order input-dependent correction, suggesting a possible direction for extending the analysis. We primarily analyze one-dimensional scalar sequences and apply TDI to the first SSM layer; an additional ablation shows that applying the same input-derived spectrum to all layers can help but is not uniformly preferable (\hyperref[app:tdi-layer-scope]{Appendix~\ref*{app:tdi-layer-scope}: \nameref*{app:tdi-layer-scope}}). For multivariate inputs, extending the convolutional representation to a block-Toeplitz operator with a matrix-valued frequency response is straightforward, while a full treatment of cross-channel task structure and principled layer-dependent initialization remains future work. Additionally, TDI also depends on reliable estimation of task-relevant spectra, which can be noisy at extremely small training ratios. Finally, TDI is not intended as a universal improvement: its benefits are most pronounced when task-relevant structure is spectral and mismatched with the default SSM bias.

\newpage
\section*{Acknowledgments and Disclosures}
\textbf{Funding.} This work was supported by the National Key R\&D Program of China, Project Number 2025YFA1016700. This work was also supported by the National Natural Science Foundation of China (NSFC) under Grant No. 62576011.
This work was supported in part by the Beijing Major Science and Technology Project under Contract no. Z251100008125055. This work was also supported by Beijing Academy of Artificial Intelligence (BAAI). 

\textbf{Competing interests.}
The authors declare no competing interests.
\bibliographystyle{plainnat}
\bibliography{bib}


\appendix
\setcounter{figure}{0}
\setcounter{table}{0}
\renewcommand{\thefigure}{\Alph{section}.\arabic{figure}}
\renewcommand{\thetable}{\Alph{section}.\arabic{table}}
\section{Related Work}
\label{app:related work}
\paragraph{Inductive bias and kernel generalization.}
Inductive bias shapes generalization by steering models toward solutions aligned with specific assumptions~\citep{hastie2009elements}. In classical kernel regression, the kernel itself encodes the inductive bias~\citep{scholkopf2002learning}, and a large body of work has investigated how different kernels shape generalization~\citep{cucker2002best, gyorfi2002distribution, belkin2018understand, belkin2018overfitting, belkin2019does, spigler2020asymptotic}. In particular, \citet{canatar2021spectral, bordelon2021population} derived analytical characterizations of the generalization error for finite datasets and arbitrary kernels, revealing how spectral bias and code–task alignment govern kernel generalization. Our work builds on this view by deriving the kernel induced by an LTI SSM convolution operator and using its spectrum to characterize the model's frequency-dependent inductive bias.

\paragraph{State space models and spectral bias.} State space models (SSMs) have recently emerged as effective architectures for modeling long sequences~\citep{patro2025mamba, gu2020hippo, gu2022efficientlymodelinglongsequences,gu2022parameterizationinitializationdiagonalstate, gupta2022diagonal, nguyen2022s4nd,smith2023simplified}. 
Recent selective SSMs such as Mamba~\citep{mamba, mamba2} introduce input-dependent dynamics and achieve strong empirical performance, while LTI SSMs such as S4 and S4D remain an important analytically tractable setting for understanding the filtering behavior underlying this model family.  The frequency bias of LTI SSMs has recently attracted attention: \citet{yu2024tuningfrequencybiasstate,solozabal2025uncovering} analyze and tune this bias, and \citet{agarwal2024spectral} propose spectral initializations to better control frequency response.  In contrast, we connect the SSM frequency response to an induced kernel spectrum and explicitly relate this spectrum to task-model alignment and sample efficiency.

\paragraph{Recurrent dynamics and spectral structure.}
Beyond SSMs, spectral and dynamical properties of recurrent networks have been studied as a way to characterize temporal computation. Prior work has related the eigenspectrum of recurrent connectivity to task-dependent dynamics in trained RNNs, and examined how biologically motivated excitatory–inhibitory constraints reshape these spectra and the resulting dynamics~\citep{jarne2023different,jarne2024effect}. Related neuro-inspired work has investigated how recurrent dynamical substrates and criticality support computation~\citep{Pontes-Filho2020A}, while recent task-trained RNN models show that structured attractor-like dynamics can emerge through learning and support robust generalization~\citep{chen2026emergent}. In contrast, our work focuses on LTI SSMs and establishes an explicit link from frequency response to an induced kernel spectrum, mode-wise learnability, and finite-sample generalization.

\paragraph{Initialization and data-dependent adaptation.}
Initialization plays an important role in SSM performance, as the initial state dynamics determine which temporal modes are emphasized before training~\citep{gu2022parameterizationinitializationdiagonalstate,liu2025autocorrelation}.  More broadly, data-dependent initialization and training schemes aim to adapt model biases to the structure of the data~\citep{liu2024from,liu2025autocorrelation}.  Prior work studies SSM frequency bias tuning~\citep{yu2024tuningfrequencybiasstate} and data-dependent initialization~\citep{liu2025autocorrelation}, but does not connect it to kernel spectra or mode-wise learnability. Another line of research uses data-driven self-pretraining as initialization before downstream training~\citep{amos2024never}. Instead, we provide a kernel-theoretic link from frequency response to sample efficiency. Additionally, our method focuses on a lightweight, task-dependent spectral initialization: TDI aligns the initial SSM response to task-specific spectral statistics estimated from input-output pairs, unlike input-agnostic or generic frequency-bias methods.

\section{Proofs and Theoretical Analysis}
\label{Appendix: proof}
This section provides complete proofs of all the theorems and lemmas stated in the main text.
\subsection{Properties of the SSM-Induced Kernel}
\label{Appendix: proof 1}
We first present proofs characterizing the SSM-induced kernel. We begin by recalling our definition of SSM-induced kernel in Section~\ref{subsec: ssm-induced kernel}.
\begin{restatable}[Positive Semi-Definiteness]{lemma}{PositiveSemiDefiniteness}
    \label{lemma: Positive Semi-Definiteness}
The SSM-induced kernel $K_{\mathrm{SSM}}(\pmb{u},\pmb{u}')$ is a valid kernel.
\end{restatable}
We first show our kernel is valid by  Lemma~\ref{lemma: Positive Semi-Definiteness}.
\begin{proof}[Proof of Lemma~\ref{lemma: Positive Semi-Definiteness}]
    By definition, for any set of input sequences $\{\pmb{u}^1,\cdots,\pmb{u}^P\}$, the Gram matrix $\mathbf{K}_{\mathrm{SSM}}$ has entries $\mathbf{K}_{\mu\nu} = K_{\mathrm{SSM}}(\pmb{u}^\mu ,\pmb{u}^\nu) = \frac{1}{L} (\pmb{T}_h \pmb{u}^\mu)^\top (\pmb{{T}_h} \pmb{u}^\nu)$.
    Let $\mathbf{U}=[\pmb{u}^1,\cdots, \pmb{u}^P]$ be the matrix of input sequences. The Gram matrix can be written in matrix form as $\mathbf{K} = \frac{1}{L} (\pmb{T}_h \mathbf{U})^\top (\pmb{T}_h \mathbf{U}).$
    For any non-zero vector $z\in\mathbb{R}^P$, we have:
    \begin{align}
    \pmb{z}^\top \mathbf{K} \pmb{z} = \frac{1}{L} \pmb{z}^\top (\pmb T_h \mathbf{U})^\top (\pmb T_h \mathbf{U}) \pmb{z} 
    = \frac{1}{L} \left\lVert(\pmb T_h \mathbf{U}) \pmb{z}\right\rVert_2^2 \ge 0.\notag
    \end{align}
    Therefore, since the Gram matrix $\mathbf{K}$ is positive semi-definite (and symmetric obviously), the kernel $K_{\mathrm{SSM}}(\pmb{u},\pmb{u}')$ is valid.
\end{proof}
Inserting the singular value decomposition (SVD) of Toeplitz matrix $\pmb T_h=\sqrt{L}\pmb{USV}^\top$ into Definition~\ref{def: ssm induced kernel}, we obtain $K_{\mathrm{SSM}}(\pmb{u},\pmb{u}')= (\pmb{V}^\top \pmb{u})^\top \pmb{S}^2 (\pmb{V}^\top \pmb{u}')$.
Then, we prove rows of  $\pmb{V}^\top \pmb u$ form a set of eigenfunctions of the kernel.
\begin{restatable}[Orthonormal Eigenfunctions]{lemma}{OrthonormalEigenfunctions}
    \label{lemma: orthogonal}
    For whitened input coordinates, rows of $\pmb{V}^\top \pmb{u}$ form a set of orthonormal eigenfunctions for the SSM-induced kernel.
\end{restatable}
\begin{proof}[Proof of Lemma~\ref{lemma: orthogonal}]
    Let $\pmb{v}^\top_i$ denote the $i$th row of the right singular matrix $\pmb{V}^\top$. For any integer $i,j\in [1,L]$, \[\int (\pmb{v}^\top_i\pmb{u})\cdot(\pmb{v}^\top_j\pmb{u})p(\pmb{u})d\pmb{u}=\mathbb{E}_{\pmb{u}}\left[\left(\pmb{V}^\top\pmb{uu}^\top\pmb{V}\right)_{ij}\right]=\left(\pmb{V}^\top\mathbb{E}_{\pmb{u}}[\pmb{u}\pmb{u}^\top]\pmb{V}\right)_{ij}=\delta_{ij}.\]
\end{proof}
The finite-dimensional decomposition above, together with Lemma~\ref{lemma: orthogonal}, establishes Theorem~\ref{Theorem: eigen}.
\EigenDecomposition*
\subsection{Frequency-Domain Characterization of the SSM Spectrum}
\label{Appendix: relation}

In this part, we analyze the relation between singular values of the SSM's Toeplitz matrix (also eigenvalues of SSM-induced kernel) and the frequency response of SSM (Theorem~\ref{thm:frequency-response-spectrum}). 
\FrequencyResponseSpectrum*
\begin{proof}
We begin by introducing a lemma from \citet{gray2006toeplitz}.
\begin{lemma}[Lemma 4.6 of \citet{gray2006toeplitz}]
\label{appendix lemma: asympotic}
    Let $\pmb{T}_n(f)=\{t_{k-j}\}_{j,k=0,\cdots, n-1}$ where 
    \[\sum_{k=-\infty}^\infty \lvert t_k\rvert<\infty,\] and \[
        f(\lambda)=\sum_{k=-\infty}^\infty t_k e^{ik\lambda}, \quad\hat{f}_n(\lambda)=\sum_{k=-(n-1)}^{n-1}t_k e^{ik\lambda}.
    \]
    Define the circulant matrices $\pmb C_n(f)$ with top row $(c_0^{(n)},\cdots,c_{n-1}^{(n)})$, and $\pmb C_n(\hat{f})$ with top row $(\hat{c}_0^{(n)},\cdots,\hat{c}_{n-1}^{(n)})$ where \[c_k^{(n)}=\frac{1}{n}\sum_{j=0}^{n-1}f(2\pi j/n)e^{2\pi ijk/n},\quad \hat{c}_k^{(n)}=\frac{1}{n}\sum_{j=0}^{n-1}\hat f_n(2\pi j/n)e^{2\pi ijk/n}.\] Then $\pmb C_n(f)\sim\pmb C_n(\hat{f}_n)\sim\pmb T_n$ as $n\to\infty$.
\end{lemma}
Lemma~\ref{appendix lemma: asympotic} states that as $n\to \infty$, $\pmb T_n$ is asymptotically equivalent to a circulant matrix $\pmb C_n(f)$ or $\pmb{C}_n(\hat f)$. Based on the fact that the eigenvalues of a circulant matrix are given by the discrete Fourier transform of its top row $(c^{(n)}_0,\cdots,c_{n-1}^{(n)})$,
\begin{align}
    \lambda_k = \sum_{j=0}^{n-1}c_j^{(n)}e^{-2\pi ijk/n},\notag
\end{align}
we obtain
\begin{align}
    \lambda_k(\pmb C_n(f))=\sum_{j=0}^{n-1}\frac{1}{n}\sum_{l=0}^{n-1}f(2\pi l/n)e^{2\pi ijl/n}e^{-2\pi ijk/n}=f(2\pi k/n)=\sum_{p=-\infty}^\infty t_p e^{ip\cdot 2\pi k/n}.\notag
\end{align}
Since $\pmb C_n$ is circulant, its singular values $s_k$ are equal to the absolute value of its eigenvalues $\lambda_k$. Thus we arrive at
\begin{align}
    s_k (\pmb T_n)\sim s_k(\pmb C_n)=\lvert\lambda_k(\pmb{C}_n)\rvert=\left\lvert \sum_{p=-\infty}^\infty t_p e^{ip\cdot 2\pi k/n}\right\rvert.\notag
\end{align}
For the SSM's Toeplitz matrix, $t_p=h_{-p}$ (see Section~\ref{subsec: ssm}). Then, $t_{p>0}=0, t_{p\le0}=h_{-p}$ is the impulse response kernel of SSM. Then, 
\begin{align}
    s_k=\left\lvert \sum_{p=-\infty}^\infty t_p e^{ip\cdot 2\pi k/n}\right\rvert=\left\lvert\sum_{p=0}^\infty h_p e^{-ip\omega_k}\right\rvert=\lvert H(\omega_k)\rvert,\notag
\end{align}
where $\omega_k=\frac{2\pi k}{n}(k=0,\cdots,n-1)$ and $H(\omega)$ is exactly the frequency response (or transfer function) of the SSM (see Section~\ref{subsec: ssm}). We summarize this result in Theorem~\ref{theorem: relation}. 
\begin{theorem}[Asymptotic Behavior of Singular Values]
\label{theorem: relation}
    For large $n$, the $k$-th singular value $s_k$ of the SSM's Toeplitz matrix $\pmb T_h\in\mathbb{R}^{n\times n}$ is asymptotically approximated by the frequency response $\lvert H(\omega_k)\rvert$, where $\omega_k = \frac{2\pi k}{n}$ for $k=0,\dots,n-1$.
\end{theorem}
With Theorem~\ref{theorem: relation}, it is straightforward to prove Theorem~\ref{thm:frequency-response-spectrum}. 
\end{proof}
Similar to the derivation of singular values, we can analogously show that the $k$-th row of $\pmb V^\top$ corresponds to the Fourier basis with frequency $\omega_k$.
\begin{theorem}[Asymptotic Behavior of Right Singular Vectors]
\label{theorem: relation basis}
    For large $n$, the $k$-th row of the right singular matrix $\pmb{V}^\top$ of the SSM's Toeplitz matrix $\pmb T_h\in\mathbb{R}^{n\times n}$ is asymptotically approximated by the Fourier basis with frequency $\omega_k$ for $k=0,\dots,n-1$.
\end{theorem}
It is worth noting that the singular values and the right singular matrix are not sorted here; the index $k$ refers to the corresponding frequency $\omega_k = \frac{2\pi k}{n}$.

While we assume a unit input covariance for simplicity, our main results can be generalized. For a general input covariance $\mathbb{E}_{\pmb{u}}[\pmb{uu}^\top]=\pmb{\Sigma}$, we reparameterize the input as $\pmb{x}=\pmb{\Sigma}^{-\frac{1}{2}}\pmb{u}$, which ensures $\mathbb{E}[\pmb{xx}^\top]=\pmb{I}_L.$ Then, the original Toeplitz map $\pmb{T}_h\pmb{u}$ is equivalent to $\tilde{\pmb{T}}\pmb{x}$ where $\tilde{\pmb{T}}=\pmb{T}_h\pmb{\Sigma}^{\frac{1}{2}}=\sqrt{L}\tilde{\pmb{U}}\tilde{\pmb{S}}\tilde{\pmb{V}}^\top$. This modified SVD result determines the eigen-decomposition by 
$\eta_\rho=\tilde{s}_\rho^2,\phi_\rho(\pmb{u})=\left(\tilde{\pmb{V}}^\top\pmb{\Sigma}^{-\frac{1}{2}}\pmb{u}\right)_\rho$, for $\rho=1,\cdots,L.$
While $\tilde{s}_\rho$ does not admit a closed-form expression, one can demonstrate that $s_\rho$ is strongly associated with the degree of alignment between the SSM and the data (see below). Our analysis indicates that greater alignment leads to larger values of $s_\rho$, thereby accelerating the learning dynamics of $\phi_\rho$.

In the unwhitened data case, $\tilde{\pmb T}$ defined above can be expressed as $\tilde{\pmb T}=\pmb{USV}^\top \pmb{QDQ}^\top$, where $\pmb{QDQ}^\top=\pmb{\Sigma}^{\frac{1}{2}}$. Thus, the SVD of $\tilde{\pmb T}$ is determined by the SVD of $\pmb M=\pmb S(\pmb V^\top\pmb{Q})\pmb D$, where $\tilde{\pmb T}=\pmb{UU_M\Sigma_MV_M}^\top\pmb{Q}^\top$ for $\pmb{M}=\pmb{U_M\Sigma_MV_M}^\top$. Since both $\pmb V^\top$ and $\pmb Q$ are orthogonal, $\pmb V^\top \pmb Q$ is itself an orthogonal change of basis: it maps the Fourier basis $\pmb V^\top$ into the eigenbasis of the data covariance (columns of     $\pmb Q$).
Therefore, the singular values of $\pmb M$, and hence of $\tilde{\pmb T}$, are governed by how the frequency response encoded in $\pmb S$ aligns with the spectral distribution of the high-variance directions encoded in $\pmb D$. In particular, a large singular value arises when a dominant data mode (large $\pmb D_j$) has most of its energy concentrated in frequencies where the LTI operator exhibits strong gain, resulting in faster learning at the corresponding basis. Actually, our spectral matching method in Section~\ref{subsec: tdi} is inspired by this finding.
\subsection{Theoretical Analysis of Task-Dependent Initialization}
\label{Appendix: proof 2}
Finally, we present proofs showing that minimizing the spectral matching loss promotes task-model spectral alignment by increasing a cross-spectrum alignment objective.

Let $S_\mathrm{model}=\lvert H(\omega)\rvert^2$ denote the model spectrum derived from the SSM's intrinsic frequency response and $\mathbf C_{\pmb{uY}}=\mathbb E_{\pmb u}[\pmb{uY}^\top]\approx\frac{1}{P}\sum_{\mu=1}^P\pmb u^\mu {\pmb Y^\mu}^\top$ denote  the input-output cross-covariance matrix, where $\pmb u$ and $\pmb Y$ are centered. The spectral matching loss is defined in Section~\ref{subsec: tdi}. 
\SpectralMatchingLoss*
We provide a two-part theoretical justification. We define the alignment objective $J$ as the Frobenius norm of the product of the Toeplitz matrix and the input-output cross-covariance matrix, $J\equiv \frac{1}{L}\lVert \pmb T_h\mathbf C_{\pmb{uY}}\rVert_F^2$. First, we show that maximizing the objective $J$ promotes a rapid increase in $C(\rho)$ (Lemma~\ref{lemma: maximize J}). Then we show that minimizing spectral matching loss $\mathcal{L}_\mathrm{spec}$ is an effective strategy for maximizing the objective $J$ (Lemma~\ref{lemma: minimize spectrum loss}). 
\begin{restatable}{lemma}{MaximizeJ}
    \label{lemma: maximize J}
    Assume $\eta_1\geq\eta_2\geq\cdots\geq\eta_L\geq0$. The total task-relevant projection energy $E=\sum_{\rho=1}^{L}\overline w_\rho^2$ is invariant. The alignment objective satisfies
    \[
    J\equiv\frac{1}{L}\lVert \pmb T_h\mathbf C_{\pmb{uY}}\rVert_F^2=\sum_{\rho=1}^{L}\eta_\rho\overline w_\rho^2.
    \]
    For fixed $\{\eta_\rho\}_{\rho=1}^{L}$, increasing $J$ favors concentrating task-relevant projection power in higher-eigenvalue kernel modes, and therefore favors a faster-rising cumulative-power curve $C(\rho)$. More precisely, any redistribution of projection power from a lower-eigenvalue mode to a higher-eigenvalue mode increases $J$ and promotes rapid increase in $C(\rho)$.
\end{restatable}
\begin{proof}[Proof of Lemma~\ref{lemma: maximize J}]
    Since the eigenfunctions of the SSM-induced kernel are linear, we write
    $\phi_\rho(\pmb u)=\pmb\alpha_\rho\pmb u$.
    Let $\overline w_{\rho,j}$ denote the projection coefficient of the
    $j$-th component of $\pmb Y$ onto $\phi_\rho$:
    \[
    \overline w_{\rho,j}
    =
    \mathbb E_{\pmb u}[\phi_\rho(\pmb u)\pmb Y_j]
    =
    \pmb\alpha_\rho\mathbb E_{\pmb u}[\pmb u\,\pmb Y_j]
    =
    \pmb\alpha_\rho\mathbf C_j
    =
    \phi_\rho(\mathbf C_j),
    \]
    where $\mathbf C_j$ is the $j$-th column of $\mathbf C_{\pmb{uY}}$. Therefore,
    \[E=\sum_{\rho=1}^{L}\overline w_\rho^2=\sum_{\rho=1}^{L}\sum_{j=1}^{d}\phi_\rho(\mathbf C_j)^2=\sum_{j=1}^{d}\|\mathbf C_j\|_2^2=\|\mathbf C_{\pmb{uY}}\|_F^2 \,\,\text{is invariant},
    \]
    \[
    \begin{aligned}
    J\equiv\frac{1}{L}\lVert \pmb T_h\mathbf C_{\pmb{uY}}\rVert_F^2
    =\sum_{\rho=1}^{L} \eta_\rho \sum_{j=1}^{d}\phi_\rho(\mathbf C_j)^2
    =\sum_{\rho=1}^{L}\sum_{j=1}^{d}\eta_\rho\overline w_{\rho,j}^2
    =\sum_{\rho=1}^{L}\eta_\rho\overline w_\rho^2.
    \end{aligned}
    \]
    The last step follows from the definition in Section~\ref{subsec: kr and rkhs}.

    To characterize how $J$ depends on the allocation of projection power, consider transferring an amount $\delta>0$ from mode $j$ to mode $i$, with $i<j$ and $\eta_i>\eta_j$, while keeping the total projection energy $E$ fixed. Then
    \[J'-J=\eta_i\delta-\eta_j\delta=\delta(\eta_i-\eta_j)>0.\]
    Meanwhile, since the added power enters the cumulative sum at mode $i$ and is removed only at mode $j$,
    \[C'(k)-C(k)=
    \begin{cases}
    0, & k<i,\\[2mm]
    \delta/E, & i\leq k<j,\\[2mm]
    0, & k\geq j.
    \end{cases}
    \]
    Hence $C'(k)\geq C(k)$ for all $k$, with strict inequality for $i\leq k<j$.

    Therefore, for a fixed eigenspectrum and fixed total projection energy, redistributing task-relevant projection power toward higher-eigenvalue modes increases $J$ and makes the cumulative-power curve rise faster. This establishes the preference encoded by maximizing $J$. 
    
    Although the eigenspectrum itself may change during spectral matching, for any resulting eigenvalue multiset the rearrangement inequality shows that $J$ is maximized, with respect to the pairing of modes and task-relevant projection powers, when larger $\eta_\rho$ are paired with larger $\overline w_\rho^2$, which is consistent with the transfer argument above.
\end{proof}
\begin{restatable}{lemma}{MinimizeL}
    \label{lemma: minimize spectrum loss}
    For the alignment objective $J$, there exists a constant $\kappa>0$ such that $J=\kappa\langle S_{\mathrm{model}},S_{\mathrm{task}}\rangle$. Therefore,
    \[ \mathcal L_{\mathrm{spec}} = 2- \frac{2J} {\kappa\|S_{\mathrm{model}}\|_2\|S_{\mathrm{task}}\|_2}.\]
    Hence minimizing $\mathcal L_{\mathrm{spec}}$ is exactly equivalent to maximizing the normalized alignment objective $\frac{J}{\|S_{\mathrm{model}}\|_2\|S_{\mathrm{task}}\|_2}$. If $\|S_{\mathrm{model}}\|_2$ is fixed, this is also equivalent to maximizing $J$.
\end{restatable}
\begin{proof}[Proof of Lemma~\ref{lemma: minimize spectrum loss}]
    Expanding the general alignment objective $J=\frac{1}{L}\lVert \pmb T_h\mathbf{C}_{\pmb{uY}}\rVert_F^2$ as the sum of squared Euclidean norms, $J=\frac{1}{L}\sum_{j=1}^d\lVert\pmb T_h\mathbf{C}_j\rVert_2^2$ (recall that $d$ is the dimension of labels). By Parseval's Theorem, the norm of a vector is proportional to the norm of its Fourier transform. Thus, \[
    \lVert\pmb T_h\mathbf{C}_j\rVert_2^2\propto \sum_\omega\lvert \mathcal{F}(\pmb T_h\mathbf{C}_j)(\omega)\rvert^2=\sum_\omega \left\lvert \mathcal{F}(h*\mathbf{C}_j)\right\rvert^2=\sum_\omega \lvert H(\omega)\rvert^2\left|\mathcal{F}\!\left[\mathbf{C}_{\pmb u \pmb Y}^{(:,j)}\right](\omega)\right|^2,
    \] where the last step follows from Convolution Theorem. Substituting this back into the expression for $J$, we arrive at the frequency-domain representation, \[
    J\propto\sum_\omega \lvert H(\omega)\rvert^2\sum_{j=1}^d\left|\mathcal{F}\!\left[\mathbf{C}_{\pmb u \pmb Y}^{(:,j)}\right](\omega)\right|^2\propto\sum_\omega S_\mathrm{model}(\omega)S_\mathrm{task}(\omega)=\langle S_{\mathrm{model}},S_{\mathrm{task}}\rangle.
    \]This equation shows that $J$ is proportional to the inner product between the model's power spectrum and the task's total power spectrum. Combining this relation with Definition~\ref{def:spectral-matching-loss},
    \[
    \begin{aligned}
    \mathcal L_{\mathrm{spec}}
    = \left\|\frac{S_{\mathrm{model}}}{\|S_{\mathrm{model}}\|_2} - \frac{S_{\mathrm{task}}}{\|S_{\mathrm{task}}\|_2} \right\|_2^2 = 2- 2\frac{\langle S_{\mathrm{model}},S_{\mathrm{task}}\rangle}
    {\|S_{\mathrm{model}}\|_2\|S_{\mathrm{task}}\|_2}=
    2-\frac{2J} {\kappa\|S_{\mathrm{model}}\|_2\|S_{\mathrm{task}}\|_2}.
    \end{aligned}
    \]
    This proves the result.
\end{proof}
Based on Lemma~\ref{lemma: maximize J}, Lemma~\ref{lemma: minimize spectrum loss}, and the kernel generalization theory discussed in Section \ref{subsec: kr and rkhs}, we derive Theorem~\ref{theorem: spectrum matching}.
\FINAL*

\subsection{Parameter-Level Analysis of SSM Frequency Bias}
\label{Appendix: s4d}
In this section, we will analyze the relation between frequency bias of an SSM and its parameters. 

First, we restate the definition of frequency bias in \citet{yu2024tuningfrequencybiasstate}: frequency bias of an SSM means that the frequency responses of the systems have more variation in the low-frequency area than the high-frequency area.

The frequency response $H(\omega)$ of an SSM is actually the discrete-time Fourier 
transform of its impulse response $h$,\[
    H(\omega)=\operatorname{DTFT}(h)=\sum_{n=0}^\infty h_ne^{-i\omega n},\quad\omega\in(0,2\pi).\]
Substituting $ h_n = \overline{\pmb{C}}\overline{\pmb{A}}^n\overline{\pmb{B}}$, 
\begin{align*}
    H(\omega)=\sum_{n=0}^\infty \overline{\pmb{C}}\overline{\pmb{A}}^n\overline{\pmb{B}}e^{-i\omega n}=\overline{\pmb C}(\pmb {I}-\overline{\pmb A}e^{-i\omega})^{-1}\overline{\pmb B}.
\end{align*}
We substitute $\pmb A, \pmb B, \pmb C,\Delta$ into $\overline{\pmb A}, \overline{\pmb B},\overline{\pmb C}$, and set $\pmb B=\pmb 1\in\mathbb{R}^{n\times1}$ (fixed $\pmb B$ in S4D training), we derive
\begin{align*}
    H(\omega)=\sum_k \pmb C_k \left(e^{\Delta \pmb A_k}-1\right)\frac{1}{\pmb A_k}\frac{1}{1-e^{-i\omega+\Delta\pmb A_k}},
\end{align*}
where $\pmb{A}_k$ denotes the $k$-th diagonal element of $\pmb A$ and $\pmb C_k$ denotes the $k$-th element of $\pmb C$. This result is similar to equation (3) in \citet{yu2024tuningfrequencybiasstate}.
$\lvert H(\omega)\rvert$ attains an extremum when the denominator reaches its extremum, which is given by \[\frac{d}{d\omega}\operatorname{Re}e^{-i\omega+\Delta(\operatorname{Re}\pmb A_k + i\operatorname{Im} \pmb A_k)}=0.\] Thus, $\omega=\Delta\operatorname{Im}\pmb A_k$ gives the location of the local extremum of $\lvert H(\omega)\rvert$. Additionally, the corresponding $\pmb C_k$ and $\operatorname{Re}\pmb A_k$ determine the magnitude at $\omega_k=\Delta \operatorname{Im}\pmb A_k$.
\begin{theorem}
    \label{theorem: s4d}
    Let $\lvert H(\omega)\rvert$ be the frequency response of an S4D-like SSM with parameters $\pmb A\in\mathbb{C}^{n\times n}, \pmb C\in\mathbb{C}^{1\times n}, \Delta\in\mathbb{R}$. Then, the location of its extremum is determined by $\omega_k=\Delta \operatorname{Im}\pmb A_k$, whereas the magnitude of the extremum is determined by $\operatorname{Re} \pmb A$ and $\pmb C$.
\end{theorem}
This result can successfully explain the frequency bias of the SSM. We now provide an estimate for S4D. Since the original S4D initialization gives\[
 \pmb A_k = -0.5+i \pi\cdot k,\quad k=1,\cdots,\lfloor n/2\rfloor,
\]
the extremum points are then determined by $\omega_k=\pi\Delta k$. Typically, the number of states, $n$, is 64, and $\Delta$ is sampled from a log-uniform distribution between 0.001 and 0.1. Thus, the proportion of the largest $\omega_k$ relative to $\pi$,  $\omega_n/\pi=\Delta\cdot\lfloor n/2\rfloor$ satisfies a log-uniform distribution between 0.032 and 3.2. Thus, $P(\omega_n/\pi<0.1)=0.25, P(\omega_n/\pi<0.3)=0.49, P(\omega_n/\pi<0.6)=0.64$. This result shows that with the initialization of S4D, the informative frequency response, i.e., frequency area with more variation, has a high probability to concentrate in a narrow, low-frequency area, causing a low frequency bias of SSM. The first approach to tuning the frequency bias in \citet{yu2024tuningfrequencybiasstate}, involving the simultaneous multiplication of all frequencies by a coefficient, is equivalent to adjusting the range of $\Delta$.

\section{Task Spectrum Estimation by Fisher Log-Power}
\label{Appendix:fisher-log-power}
For a sequence-to-class dataset
$\mathcal{D}=\{(\pmb u^\mu,\pmb y^\mu)\}_{\mu=1}^P$, where
$\pmb y^\mu\in\{0,1\}^d$ is the one-hot label and $\pmb u^\mu$ has sequence
length $L$, we define
\[
    \widehat{\pmb u}^{\mu}(\omega_k)
    =
    \mathrm{rFFT}(\pmb u^\mu)_k,
    \qquad
    k=0,\dots,\lfloor L/2\rfloor .
\]
The Fisher log-power method forms
\[
    z^\mu(\omega_k)
    =
    \log\left(
        \left|\widehat{\pmb u}^{\mu}(\omega_k)\right|^2+\epsilon
    \right).
\]
For class $j$, let
\[
    \pi_j=\mathbb{E}[y_j],
    \qquad
    \mu_j(\omega_k)
    =
    \mathbb{E}\!\left[
        z(\omega_k)\mid y_j=1
    \right],
\]
and
\[
    \sigma_j^2(\omega_k)
    =
    \operatorname{Var}\!\left(
        z(\omega_k)\mid y_j=1
    \right).
\]
The class-prior-weighted mean is
\[
    \bar\mu(\omega_k)
    =
    \sum_{j=1}^d \pi_j\mu_j(\omega_k).
\]
The Fisher log-power spectrum used in the experiments is
\[
    S_{\mathrm{FLP}}(\omega_k)
    =
    \frac{
        \sum_{j=1}^d
        \pi_j
        \left(
            \mu_j(\omega_k)-\bar\mu(\omega_k)
        \right)^2
    }{
        \sum_{j=1}^d
        \pi_j\sigma_j^2(\omega_k)+\lambda
    },
\]
where $\epsilon>0$ stabilizes the logarithm and $\lambda>0$ is a small
regularizer. In the implementation, $\epsilon=10^{-6}$ and
$\lambda=10^{-4}$ by default.

\begin{algorithm}[htbp]
\caption{Fisher log-power task spectrum}
\label{alg:fisher-log-power}
\begin{algorithmic}[1]
\REQUIRE Dataset $\mathcal{D}=\{(\pmb u^\mu,\pmb y^\mu)\}_{\mu=1}^P$,
number of classes $d$, stabilizer $\epsilon$, Fisher regularizer $\lambda$
\STATE Initialize class counts $N_j\leftarrow 0$
\STATE Initialize accumulators $A_j(\omega)\leftarrow 0$ and
$B_j(\omega)\leftarrow 0$
\FOR{each minibatch $(\pmb u,\pmb y)$}
    \STATE Compute $\widehat{\pmb u}(\omega)=\mathrm{rFFT}(\pmb u)$
    \STATE Compute
    $z(\omega)=\log(|\widehat{\pmb u}(\omega)|^2+\epsilon)$
    \FOR{each class $j=1,\dots,d$}
        \STATE Update $N_j\leftarrow N_j+\sum_\mu y_j^\mu$
        \STATE Accumulate
        $A_j(\omega)\leftarrow
        A_j(\omega)+\sum_\mu y_j^\mu z^\mu(\omega)$
        \STATE Accumulate
        $B_j(\omega)\leftarrow
        B_j(\omega)+\sum_\mu y_j^\mu z^\mu(\omega)^2$
    \ENDFOR
\ENDFOR
\STATE $\pi_j\leftarrow N_j/\sum_{j'}N_{j'}$
\STATE $\mu_j(\omega)\leftarrow A_j(\omega)/N_j$
\STATE $\sigma_j^2(\omega)\leftarrow B_j(\omega)/N_j-\mu_j(\omega)^2$
\STATE $\bar\mu(\omega)\leftarrow \sum_j\pi_j\mu_j(\omega)$
\STATE $B_{\mathrm{between}}(\omega)
\leftarrow
\sum_j\pi_j(\mu_j(\omega)-\bar\mu(\omega))^2$
\STATE $B_{\mathrm{within}}(\omega)
\leftarrow
\sum_j\pi_j\sigma_j^2(\omega)$
\STATE \textbf{return} the Fisher log-power spectrum
$S_{\mathrm{FLP}}(\omega)
=
\frac{
    B_{\mathrm{between}}(\omega)
}{
    B_{\mathrm{within}}(\omega)+\lambda
}.$

\end{algorithmic}

\end{algorithm}

\paragraph{Connection to the cross spectrum.}
The theoretical analysis uses the cross spectrum based on the centered input-label
cross-covariance. For each class $j$, define
\[
    \mathbf C_{\pmb{uy},j}
    \equiv \mathbf{C}_{\pmb u \pmb Y}^{(:,j)}=
    \mathbb{E}\left[
        \left(\pmb{u}-\bar {\pmb u} \right)
        \left(y_j-\pi_j\right)
    \right],
    \qquad
    \pi_j=\mathbb{P}(y_j=1).
\]
Since $y_j$ is one-hot, this can be written as
\[
    \mathbf  C_{\pmb{uy},j}
    =
    \pi_j
    \left(
        \mathbb{E}\!\left[\pmb u\mid y_j=1\right]
        -
        \mathbb{E}\!\left[\pmb u\right]
    \right).
\]
Thus the cross-covariance measures class-dependent deviations of the input
signal from its marginal mean. The corresponding cross spectrum is
\[
    S_{\mathrm{cross}}(\omega)
    =
    \frac{1}{d}
    \sum_{j=1}^{d}
    \left|
        \mathcal{F}[\mathbf{C}_{\pmb{uy},j}](\omega)
    \right|^2 .
\]
Empirically, this is estimated by
\[
    \widehat S_{\mathrm{cross}}(\omega)
    =
    \frac{1}{d}
    \sum_{j=1}^{d}
    \left|
        \mathrm{rFFT}\left(
            \frac{1}{P}
            \sum_{\mu=1}^{P}
            (\pmb u^\mu-\bar{\pmb u})
            (y_j^\mu-\pi_j)
        \right)(\omega)
    \right|^2 .
\]
Therefore, $S_{\mathrm{cross}}$ measures frequencies at which the input signal
has class-dependent mean structure.

Fisher log-power follows the same label-conditioned spectral principle, but
uses the phase-invariant log-power statistic
\[
    \phi_\omega(\pmb u)
    =
    \log\left(
        |\widehat{\pmb u}(\omega)|^2+\epsilon
    \right)
\]
instead of the signed Fourier coefficient of $\pmb u$.
Its between-class numerator is
\[
    B_{\mathrm{FLP}}(\omega)
    =
    \sum_{j=1}^{d}
    \pi_j
    \left(
        \mathbb{E}\!\left[
            \phi_\omega(\pmb u)\mid y_j=1
        \right]
        -
        \mathbb{E}\!\left[
            \phi_\omega(\pmb u)
        \right]
    \right)^2 .
\]
Thus, $S_{\mathrm{cross}}$ and Fisher log-power are both frequency-wise
label-dependence measures. The cross spectrum measures phase-sensitive
linear dependence between the input signal and the label, whereas Fisher
log-power measures class separability of spectral-energy features.

They share several useful properties. Both are nonnegative spectra over
frequencies. Both assign low importance to frequencies whose relevant spectral
statistic has weak dependence on the label, even if the marginal spectral energy
at those frequencies is large. Both are estimated from empirical moments
involving the one-hot labels and can be normalized into an importance
distribution over frequencies.

\begin{figure}[htbp]
    \centering
    \subfigure[$r=0.001$, 54 examples]{\includegraphics[width=0.3\textwidth]{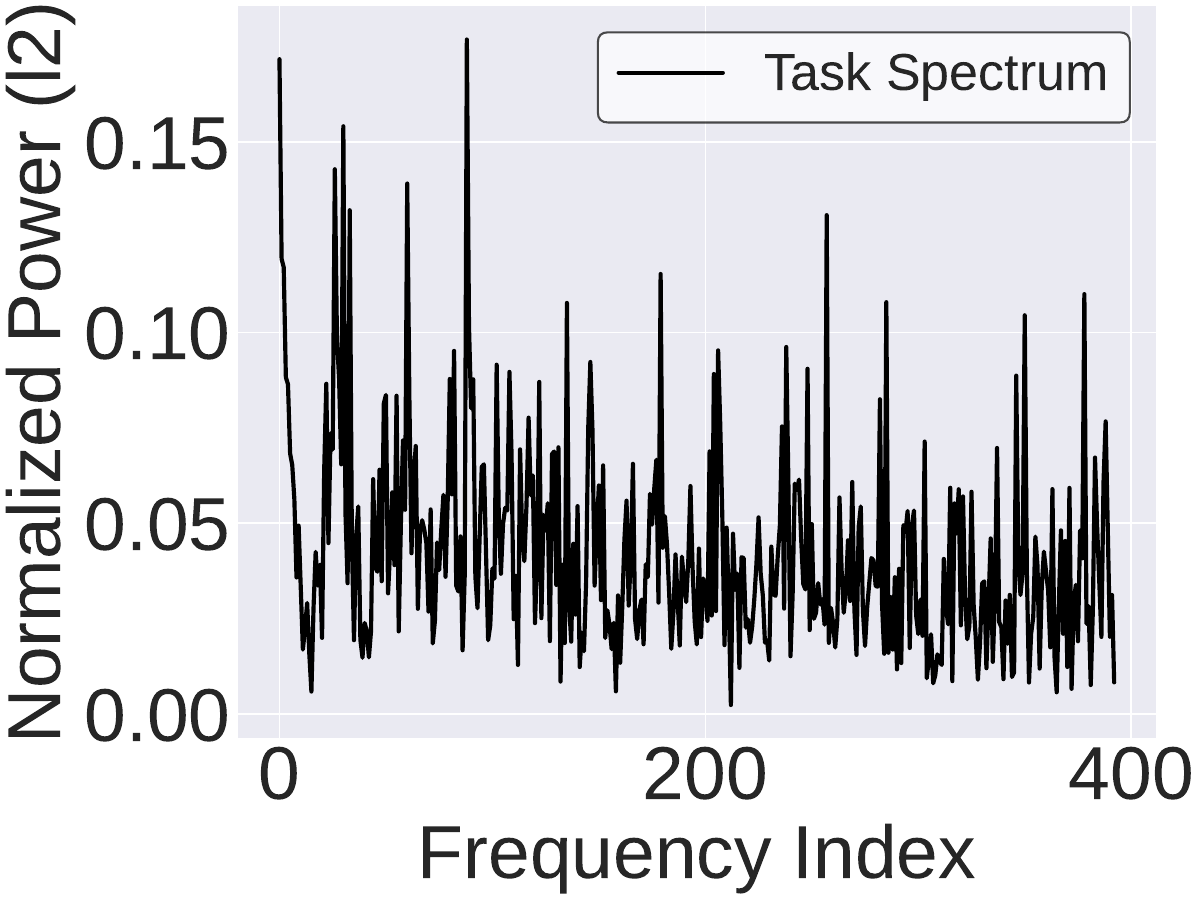}}
    \subfigure[$r=0.005$, 270 examples]{\includegraphics[width=0.3\textwidth]{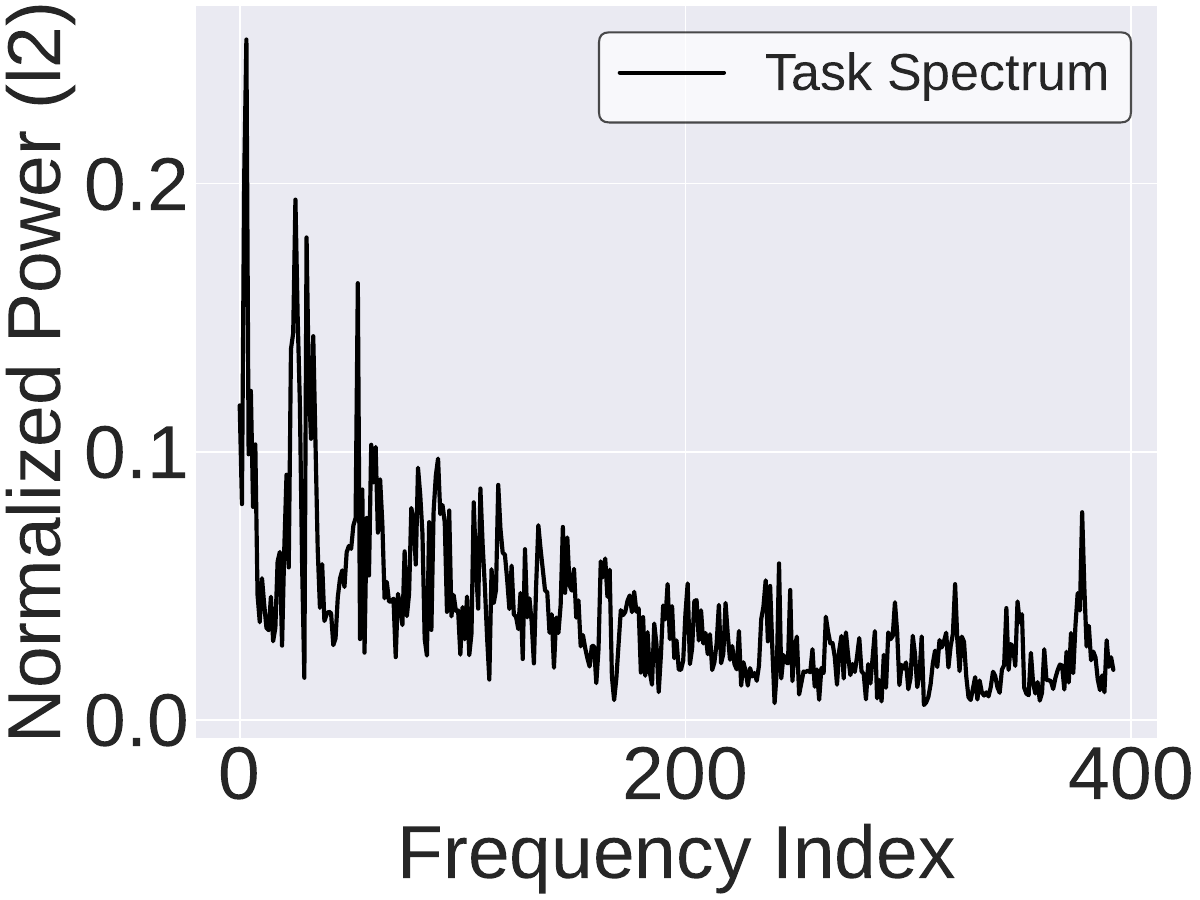}}
    \subfigure[$r=0.010$, 540 examples]{\includegraphics[width=0.3\textwidth]{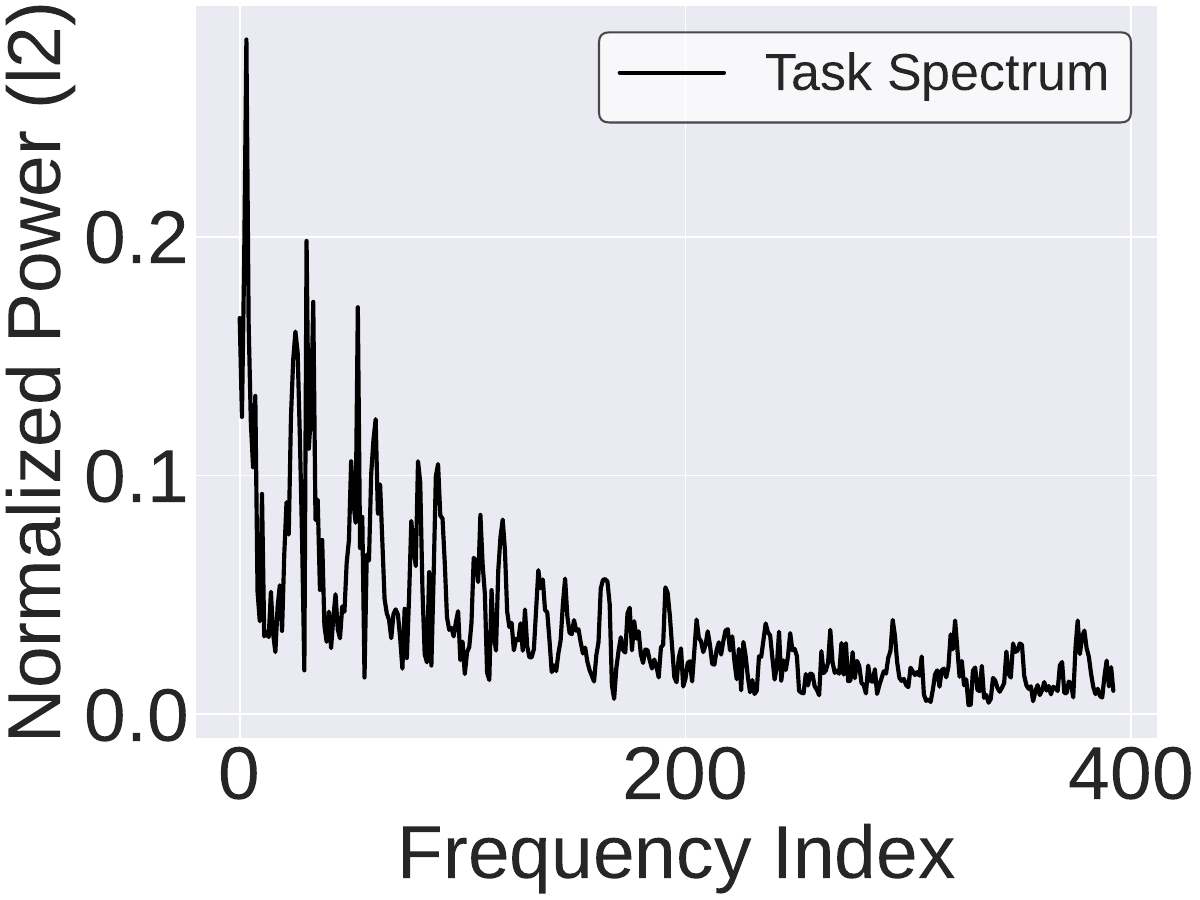}}\\
    \subfigure[$r=0.020$, 1080 examples]{\includegraphics[width=0.3\textwidth]{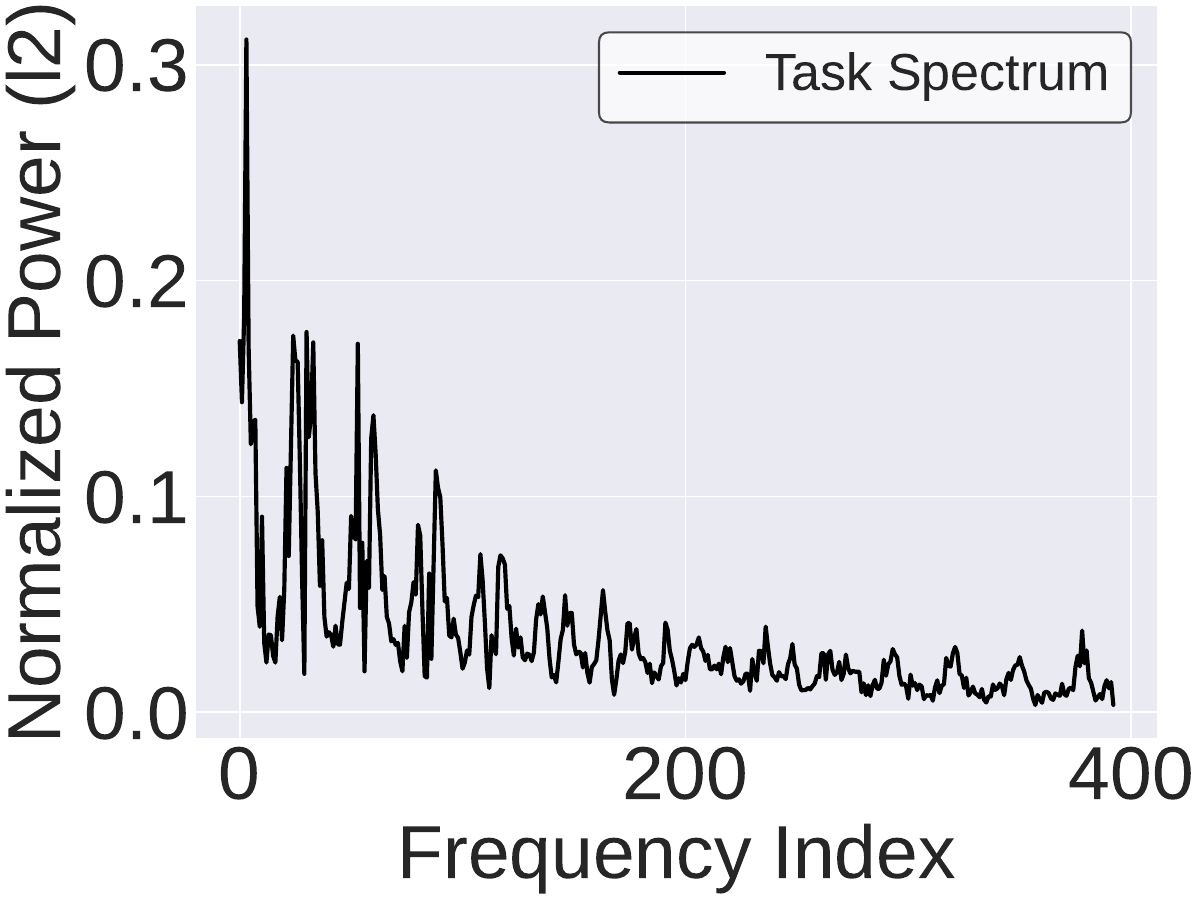}}
    \subfigure[$r=0.100$, 5400 examples]{\includegraphics[width=0.3\textwidth]{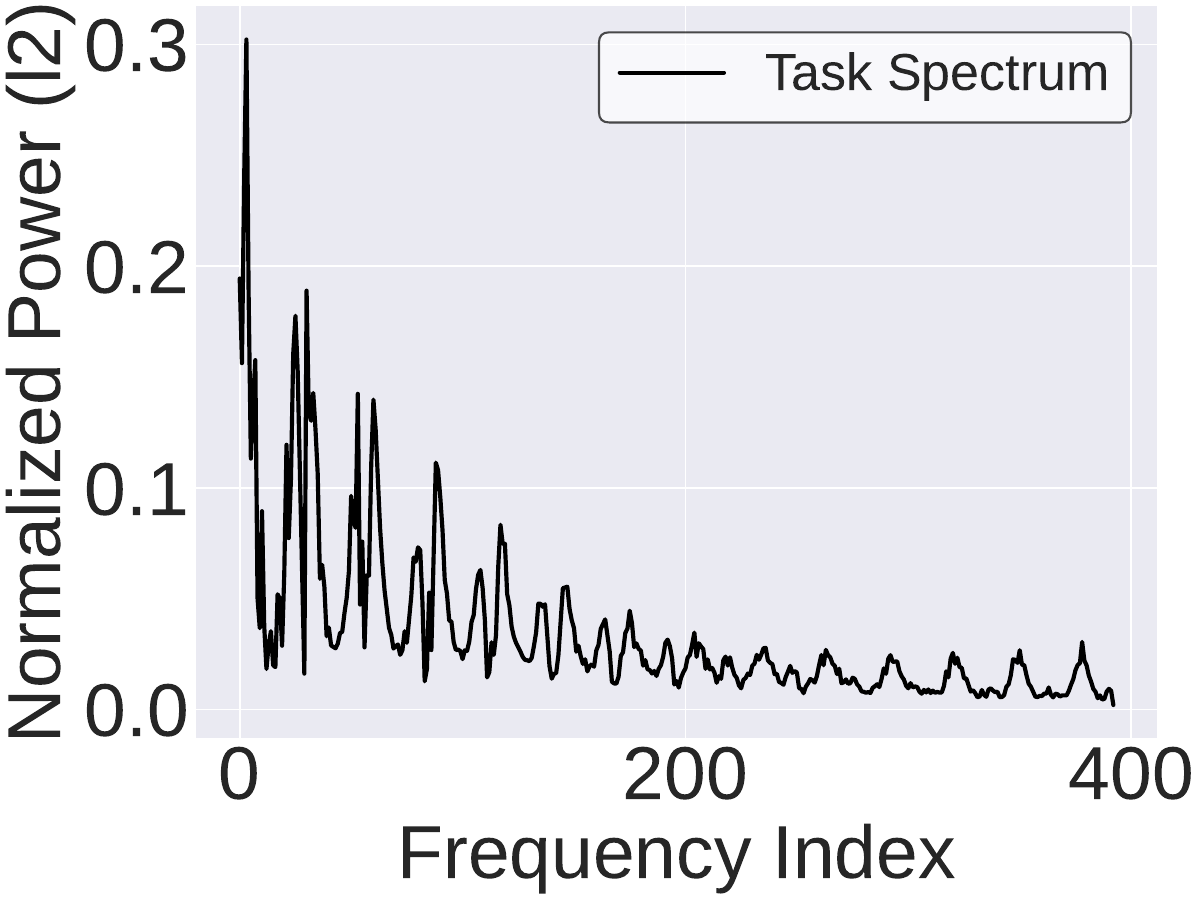}}
    \subfigure[$r=1.000$, 54000 examples]{\includegraphics[width=0.3\textwidth]{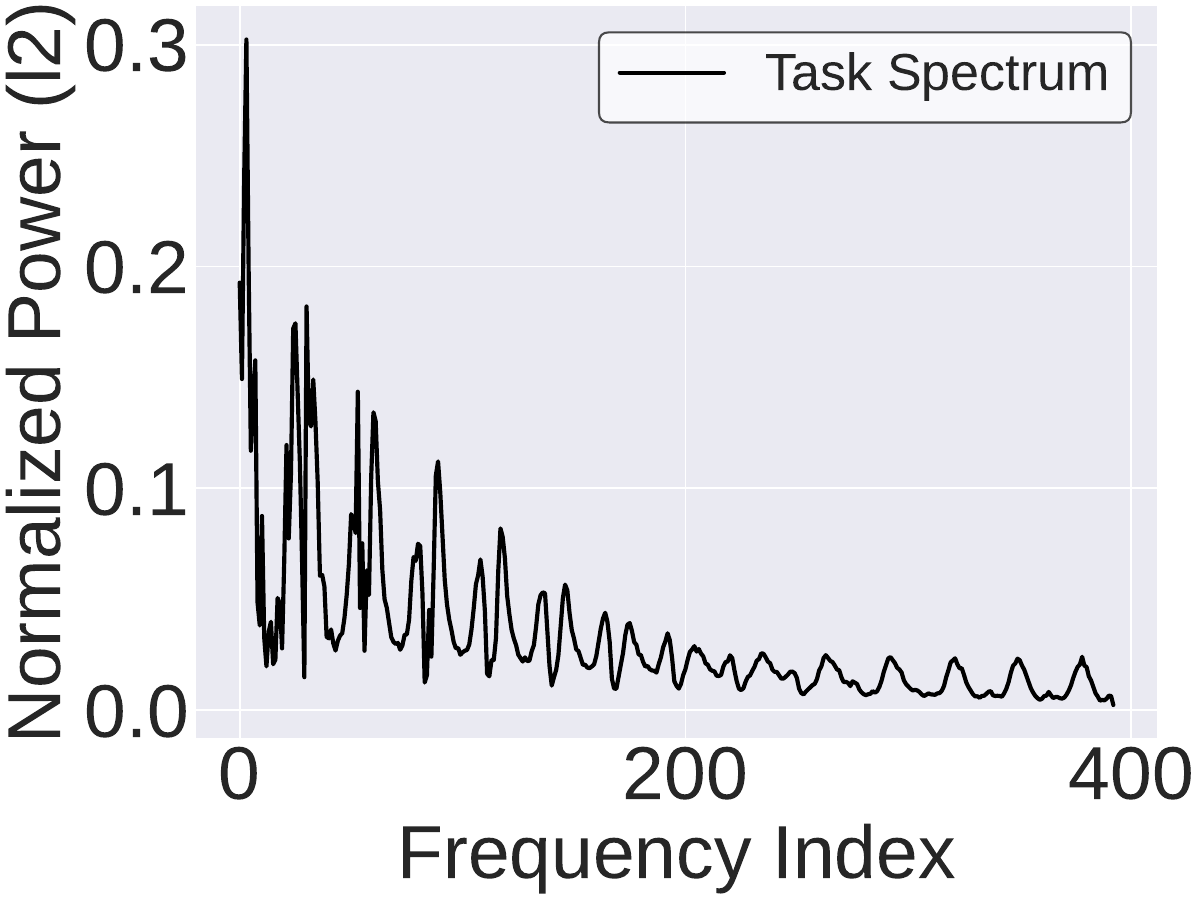}}
    \caption{\textbf{Stability of the Fisher log-power task spectrum under subsampling.} We compute the Fisher log-power spectrum on sequential MNIST using six training-subset ratios while keeping the preprocessing and frequency grid fixed.The estimated spectra become relatively stable once the number of examples reaches the order of \(10^2\): the dominant spectral peaks remain largely consistent across larger subsets, while additional examples mainly reduce small fluctuations rather than changing the main task-relevant frequency bands.}
    \label{fig: appendix spectrum}
\end{figure}

The additional Fisher normalization is
\[
    S_{\mathrm{FLP}}(\omega)
    =
    \frac{
        B_{\mathrm{FLP}}(\omega)
    }{
        W_{\mathrm{FLP}}(\omega)+\lambda
    },
    \qquad
    W_{\mathrm{FLP}}(\omega)
    =
    \sum_{j=1}^{d}
    \pi_j
    \operatorname{Var}\!\left(
        \phi_\omega(\pmb u)\mid y_j=1
    \right).
\]
Hence Fisher log-power can be viewed as a noise-normalized and phase-invariant surrogate of the cross-spectrum criterion. It preserves the key idea used in the theory, namely selecting frequencies whose statistics are predictive of the one-hot label, while discounting frequencies with large within-class variability.
We do not claim that $S_{\mathrm{FLP}}$ is equivalent to $S_{\mathrm{cross}}$; rather, it is a practical surrogate tailored to classification tasks where discriminative information is often carried by spectral energy rather than signed phase-sensitive coefficients.
\paragraph{Stability of Fisher log-power under subsampling.}
Because TDI relies on estimating a task-relevant spectrum from finite training data, we examine whether the Fisher log-power statistic is stable under different subsampling ratios.  Using sequential MNIST as a representative classification task, we compute the Fisher log-power spectrum from training subsets with ratios \(r\in\{0.001, 0.005, 0.010,0.020,0.100,1.000\}\) (Figure~\ref{fig: appendix spectrum}), while keeping the preprocessing and frequency grid fixed.  Empirically, the Fisher log-power spectrum becomes relatively stable once the number of examples reaches the order of \(10^2\): the dominant peaks remain consistent, while additional samples mainly reduce small fluctuations rather than changing the main task-relevant frequency bands.  This suggests that Fisher log-power provides a reasonably stable estimate of task-relevant spectral structure even from limited labeled data. 

We emphasize that this diagnostic does not imply perfect spectrum estimation in all low-data settings; rather, it supports the use of Fisher log-power as a practical classification-oriented surrogate for \(S_{\mathrm{task}}\).

\paragraph{Bootstrap stability diagnostic.}
\label{app:bootstrap-stability}
We complement the qualitative spectrum comparison above with a quantitative diagnostic of Fisher log-power estimation stability. The diagnostic requires labeled training data but no model training.

We evaluate a grid of candidate sample sizes. At each size, we draw five independent class-stratified subsets. For each subset, we compute its Fisher log-power spectrum and generate 50 stratified bootstrap resamples, recomputing the spectrum for each resample with the same preprocessing and frequency grid. We then calculate the correlation between each bootstrap spectrum and the original subset spectrum. A subset is considered stable if the 2.5th percentile of these correlations is at least $0.95$. The stability transition is the first evaluated sample size at which all five subsets satisfy this criterion and continue to satisfy it at every subsequent evaluated size. These transitions are coarse, grid-dependent estimates, rather than exact sample-size thresholds.

Table~\ref{tab:bootstrap-stability} summarizes the observed transitions. The first stable training ratio varies substantially across datasets, so a common training-ratio threshold is inappropriate. PathFinder does not reach the stability criterion on the evaluated grid. For ListOps, numerical token IDs lack an intrinsic spectral interpretation, so we do not apply this diagnostic.

\begin{table}[htbp]
\centering
\small
\caption{
\textbf{Bootstrap stability of the Fisher log-power spectrum.}
The first stable ratio satisfies the bootstrap criterion for all five subsets and all subsequent evaluated sizes. Ratios refer to fractions of each dataset's original training split. Stability does not imply a positive TDI accuracy gain. }
\label{tab:bootstrap-stability}
\begin{tabular}{@{}p{0.30\linewidth}p{0.18\linewidth}p{0.44\linewidth}@{}}
\toprule
Dataset & First stable ratio & Observation \\
\midrule
Freq. Cls. & $2.15\%$
& Identifiable transition \\
sMNIST & $2.15\%$
& Identifiable transition \\
Speech Commands & $4.64\%$
& Identifiable transition \\
SC10 & $4.64\%$
& Identifiable transition \\
CIFAR-10 & $21.5\%$
& Negative TDI gains on both sides of the transition \\
PathFinder & Not reached
& No transition identified on the evaluated grid \\
ListOps & Not applicable
& Token IDs lack an intrinsic spectral interpretation \\
\bottomrule
\end{tabular}
\end{table}

For each of the five datasets with an identifiable transition, we compare the matched TDI-minus-baseline accuracy gain at the grid point immediately before the transition with that at the grid point immediately after it. The gain decreases in all five cases, with the mean across datasets changing from $+9.00$ to $-0.95$ percentage points. This observation motivates a conservative empirical screen: beyond the dataset-specific stability transition, we would not apply TDI by default. Below or near the transition, TDI remains a candidate only if the estimated spectrum is informative and the default initialization is spectrally mismatched; instability alone is not a reason to apply TDI.

The diagnostic measures spectral-estimation stability, not initialization mismatch or the sign of the downstream gain. In particular, CIFAR-10 has negative TDI gains on both sides of its transition, demonstrating that stability is insufficient to ensure a benefit. For PathFinder and ListOps, we abstain from making a TDI applicability recommendation using this diagnostic. The observed transitions therefore should not be interpreted as guaranteed benefit--harm crossovers or universal measures of task difficulty.

\paragraph{Variable-length sequences.}
Our theoretical analysis assumes a fixed sequence length $L$. For variable-length numerical sequences with a common sampling interval, task spectra could be estimated on each sequence's native frequency grid and interpolated onto a shared frequency grid before aggregation, using consistent spectral normalization across lengths and excluding padded positions. This provides a possible extension of the spectrum estimator; the fixed-length kernel analysis does not directly establish guarantees for variable-length inputs.

\section{Task-Dependent Initialization Algorithms}
\label{Appendix: tdi-implementation}
TDI follows the same high-level principle across all experiments: estimate a task-relevant spectrum $S_{\mathrm{task}}$, initialize the SSM so that its model spectrum $S_{\mathrm{model}}$ is aligned with $S_{\mathrm{task}}$, and then train the model with the standard supervised objective. 

The implementation differs only in the SSM parameterization and in whether the spectrum-aware initialization is further refined by gradient descent. A unified TDI algorithm is summarized in Algorithm~\ref{alg:tdi-unified}. We use three TDI variants in our experiments, summarized in Table~\ref{tab:tdi-variants}. Kernel regression experiments use a complex diagonal SSM with peak initialization followed by gradient refinement. The one-layer RNN experiments use a real-valued matrix SSM, where peak initialization is implemented by real resonant blocks followed by gradient refinement.
Deep SSM experiments use a complex diagonal SSM with direct construction, applied to the first SSM layer for efficiency. All variants share the same spectral matching objective; they differ only in how task-relevant frequencies are converted into valid SSM parameters. It is worth noting that TDI is a one-time initialization step and introduces no inference-time overhead; in deep SSM experiments we use the constructive peak initialization without gradient refinement.

\begin{table}[t]
\centering
\caption{\textbf{TDI variants used in different experiments.}}
\label{tab:tdi-variants}
\resizebox{\textwidth}{!}{%
\begin{tabular}{lll}
\toprule
Experiment & SSM parameterization & TDI variant \\
\midrule
Kernel regression & Complex diagonal SSM & Peak initialization + gradient refinement (Algorithm~\ref{alg:tdi-complex}) \\
One-layer RNN & Real-valued matrix SSM & Peak initialization + gradient refinement (Algorithm~\ref{alg:tdi-real})\\
Deep SSM & Complex diagonal SSM & Direct construction (Algorithm~\ref{alg:tdi-complex})\\
\bottomrule
\end{tabular}
}
\end{table}
Throughout the algorithms, selected frequencies $f_j$ are represented as normalized rFFT frequencies in cycles per input step, i.e., $f_j\in[0,1/2]$. If raw FFT bin indices $b_j$ are selected, we convert them by $f_j=b_j/L$.
\begin{algorithm}[t]
\caption{Unified Task-Dependent Initialization (TDI)}
\label{alg:tdi-unified}
\begin{algorithmic}[1]
\REQUIRE Training data $\mathcal D_{\mathrm{train}}$, parameterization $p\in\{\textsc{ComplexDiag},\textsc{RealMatrix}\}$, TDI variant $v$, task-spectrum estimator $\operatorname{TaskSpectrum}$
\ENSURE Initialized SSM parameters

\STATE Estimate and normalize the task spectrum:
\[
    S_{\mathrm{task}}
    \leftarrow
    \mathcal N\!\left(
        \operatorname{TaskSpectrum}(\mathcal D_{\mathrm{train}})
    \right).
\]

\IF{$p=\textsc{ComplexDiag}$}
    \STATE Initialize the complex diagonal SSM using Algorithm~\ref{alg:tdi-complex} with $S_{\mathrm{task}}$ and variant $v\in\{\textsc{PeakGD},\textsc{Construct}\}$.
\ELSIF{$p=\textsc{RealMatrix}$}
    \STATE Initialize the real-valued SSM/RNN using Algorithm~\ref{alg:tdi-real} with $S_{\mathrm{task}}$ and variant $v=\textsc{PeakGD}$.
\ENDIF

\STATE Train the model with the standard supervised objective and the same downstream training protocol as the baseline.
\end{algorithmic}
\end{algorithm}
\begin{algorithm}[t]
\caption{TDI for Complex Diagonal SSMs}
\label{alg:tdi-complex}
\begin{algorithmic}[1]
\REQUIRE Task spectrum $S_{\mathrm{task}}$, default parameters $(\Delta_0,\pmb A_0,\pmb B_0,\pmb C_0)$, number of heads $H$, state size $N$, sequence length $L$, variant $v\in\{\textsc{PeakGD},\textsc{Construct}\}$, refinement steps $T$
\ENSURE Complex diagonal SSM parameters $\{(\Delta^{(h)},\pmb A^{(h)},\pmb B^{(h)},\pmb C^{(h)})\}_{h=1}^{H}$

\STATE Normalize $S_{\mathrm{task}}\leftarrow \mathcal N(S_{\mathrm{task}})$.
\STATE Select dominant frequencies $\{f_j\}$ and strengths $\{r_j\}$ from $S_{\mathrm{task}}$.

\FOR{$h=1$ to $H$}
    \STATE Initialize $(\Delta,\pmb A,\pmb B,\pmb C)$ from the default SSM initialization $(\Delta_0,\pmb A_0,\pmb B_0,\pmb C_0)$.

    \STATE \textbf{Peak initialization.}
    \STATE Assign each state $j$ a task frequency $f_j$ and strength $r_j$.
    \STATE Set $\Delta\leftarrow \Delta_0$.
    \STATE Set damping
       $ \gamma_j
        \leftarrow
        \gamma_{\min}
        +
        (\gamma_{\max}-\gamma_{\min})(1-r_j)$.
    \STATE Set complex diagonal modes
       $ A_j
        \leftarrow
        -\gamma_j - i\,2\pi f_j/\Delta$.
    \STATE Set gains
        $B_j,C_j
        \leftarrow
        g_{\min}+(g_{\max}-g_{\min})r_j$.

    \IF{$v=\textsc{PeakGD}$}
        \STATE \textbf{Gradient refinement.}
        \STATE Reparameterize $\Delta=\exp(\theta_{\Delta})$ and $\pmb A=-\exp(\theta_{\Re})-i\theta_{\Im}$.
        \STATE Initialize Adam on $(\theta_{\Re},\theta_{\Im},\pmb C,\theta_{\Delta})$; keep $\pmb B$ fixed.
        \FOR{$t=1$ to $T$}
            \STATE Compute impulse response
                $\pmb k
                \leftarrow
                \operatorname{S4DImpulseResponse}(\Delta,\pmb A,\pmb B,\pmb C,L)$.
            \STATE Compute normalized model spectrum
                $S_{\mathrm{model}}
                \leftarrow
                \mathcal N\!\left(
                    |\operatorname{FFT}(\pmb k)|^2
                \right)$.
            \STATE Compute spectral matching loss
                $\mathcal L_{\mathrm{spec}}
                =
                \|S_{\mathrm{model}}-S_{\mathrm{task}}\|_2^2$.
            \STATE Update $(\theta_{\Re},\theta_{\Im},\pmb C,\theta_{\Delta})$ by one Adam step.
        \ENDFOR
        \STATE Recover $\Delta$, $\pmb A$, and $\pmb C$ from the optimized parameters.
    \ELSIF{$v=\textsc{Construct}$}
        \STATE Use the peak-initialized parameters directly without gradient refinement.
    \ENDIF

    \STATE Store $(\Delta^{(h)},\pmb A^{(h)},\pmb B^{(h)},\pmb C^{(h)})\leftarrow(\Delta,\pmb A,\pmb B,\pmb C)$.
\ENDFOR

\STATE Stack head-wise parameters into SSM tensors.
\RETURN Complex diagonal SSM parameters.
\end{algorithmic}
\end{algorithm}
\begin{algorithm}[t]
\caption{TDI for Real-Valued SSMs/RNNs}
\label{alg:tdi-real}
\begin{algorithmic}[1]
\REQUIRE Task spectrum $S_{\mathrm{task}}$, initial parameters $(\Delta_0,\pmb A_0,\pmb B_0)$, state size $N$, sequence length $L$, refinement steps $K$
\ENSURE Initialized parameters $(\Delta, \pmb A,\pmb B,\pmb C)$

\STATE Normalize $S_{\mathrm{task}}\leftarrow \mathcal N(S_{\mathrm{task}})$.
\STATE Set output vector $\pmb C\leftarrow \mathbf 1_{1\times N}$.

\STATE \textbf{Peak initialization.}
\STATE Initialize $\Delta\leftarrow \Delta_0$, $\pmb A\leftarrow -0.5\mathbf I$, $\pmb B\leftarrow \pmb 0$.
\STATE Select top-$M$ frequency bins $\{f_j\}_{j=1}^{M}$ from $S_{\mathrm{task}}$, where $M=\min(\lfloor N/2\rfloor,|S_{\mathrm{task}}|)$.
\STATE Save $\pmb B_{\mathrm{init}}\leftarrow \pmb B_0$.
\FOR{$j=1$ to $M$}
    \STATE Set angular frequency
        $\omega_j\leftarrow 2\pi f_j/\Delta$.
    \STATE Set damping
        $\alpha_j
        \leftarrow
        \begin{cases}
        0.1, & j\le 6,\\
        0.5, & j>6.
        \end{cases}$
    \STATE Insert the real resonant block
        $\begin{bmatrix}
        -\alpha_j & \omega_j\\
        -\omega_j & -\alpha_j
        \end{bmatrix}$
    into rows/columns $2j-1:2j$ of $\pmb A$.
    \STATE Set $B_{2j-1}\leftarrow B_{\mathrm{init},2j-1}$.
\ENDFOR

\STATE \textbf{Gradient refinement.}
\STATE Initialize Adam on $(\Delta, \pmb A,\pmb B)$.
\FOR{$i=1$ to $K$}
    \STATE Discretize by ZOH:
    $(\bar{\pmb A},\bar{\pmb B})
        \leftarrow
        \operatorname{ZOH}(\Delta, \pmb A,\pmb B)$.
    \STATE Compute impulse response
        $\pmb k
        \leftarrow
        \operatorname{ImpulseResponse}(\bar{\pmb A},\bar{\pmb B},\pmb C,L)$.
    \STATE Compute normalized model spectrum
        $S_{\mathrm{model}}
        \leftarrow
        \mathcal N\!\left(
            |\operatorname{FFT}(\pmb k)|^2
        \right)$.
    \STATE Compute spectral matching loss
        $\mathcal L_{\mathrm{spec}}
        =
        \|S_{\mathrm{model}}-S_{\mathrm{task}}\|_2^2$.
    \STATE Update $(\Delta, \pmb A,\pmb B)$ by one Adam step.
\ENDFOR

\RETURN $(\Delta, \pmb A,\pmb B,\pmb C)$.
\end{algorithmic}
\end{algorithm}

\section{Additional Experimental Details}
\label{app:experimental_details}

\subsection{Empirical Validation of Kernel Eigen-Decomposition}
\label{app:exp1_details}

\paragraph{Dataset and preprocessing.}
We use 15,000 randomly selected MNIST images~\citep{MNIST} to empirically validate the eigendecomposition of the SSM-induced kernel. Since Theorem~\ref{Theorem: eigen} is stated in whitened input coordinates, we apply whitening before constructing the empirical kernel. After whitening, the effective dimensionality of the data is reduced, and we keep the top 550 dimensions in our main experiment. This preprocessing allows us to isolate the spectral bias induced by the SSM convolution operator from correlations in the input distribution.
\paragraph{Empirical kernel eigendecomposition.}

To obtain empirical eigenvalues and eigenfunctions, we diagonalize the Gram matrix computed on the full evaluation set. Specifically, for samples
$\{\pmb{u}^{\mu}\}_{\mu=1}^{M}$, we construct
\[
    \mathbf{K}_{\mu\nu}
    =
    K_{\mathrm{SSM}}(\pmb{u}^{\mu}, \pmb{u}^{\nu}),
    \qquad
    \mu,\nu=1,\ldots,M,
\]
where $M$ denotes the number of samples used to estimate the empirical kernel operator. We then diagonalize
\[
    \mathbf{K} = M \mathbf{\Phi}\mathbf{\Lambda}\mathbf{\Phi}^{\top},
\]
where the diagonal entries of $\mathbf{\Lambda}$ are the empirical kernel eigenvalues and the columns of $\mathbf{\Phi}$ are the corresponding empirical eigenvectors on the sampled data. The factor $M$ ensures consistency with the empirical data distribution
    $p_M(\pmb{u}) = \frac{1}{M}\sum_{\mu=1}^{M} \delta(\pmb{u}-\pmb{u}^{\mu})$.
Under this empirical distribution, the empirical eigenfunctions are represented by $\sqrt{M}\mathbf{\Phi}$, since
\[
    \frac{1}{M}
    \sum_{\mu=1}^{M}
    \left(\sqrt{M}\mathbf{\Phi}_{\mu i}\right)
    \left(\sqrt{M}\mathbf{\Phi}_{\mu j}\right)
    =
    \delta_{ij}.
\]
Thus, the Gram-matrix eigendecomposition provides empirical counterparts to the eigenvalues and eigenfunctions predicted by Theorem~\ref{Theorem: eigen}.

\paragraph{Additional results on unwhitened inputs.}
In addition to the whitened setting used in the main text, we also evaluate the same diagnostics on unwhitened MNIST inputs. The unwhitened setting no longer isolates the SSM operator alone: the effective feature operator becomes $\pmb{T}_h \Sigma^{1/2}$, so the resulting spectrum reflects both the SSM convolution and the input covariance structure. Nevertheless, the empirical kernel eigendecomposition remains consistent with the corresponding effective operator.

\begin{figure}[htbp]
\centering
\subfigure{%
\begin{overpic}[width=0.26\textwidth]{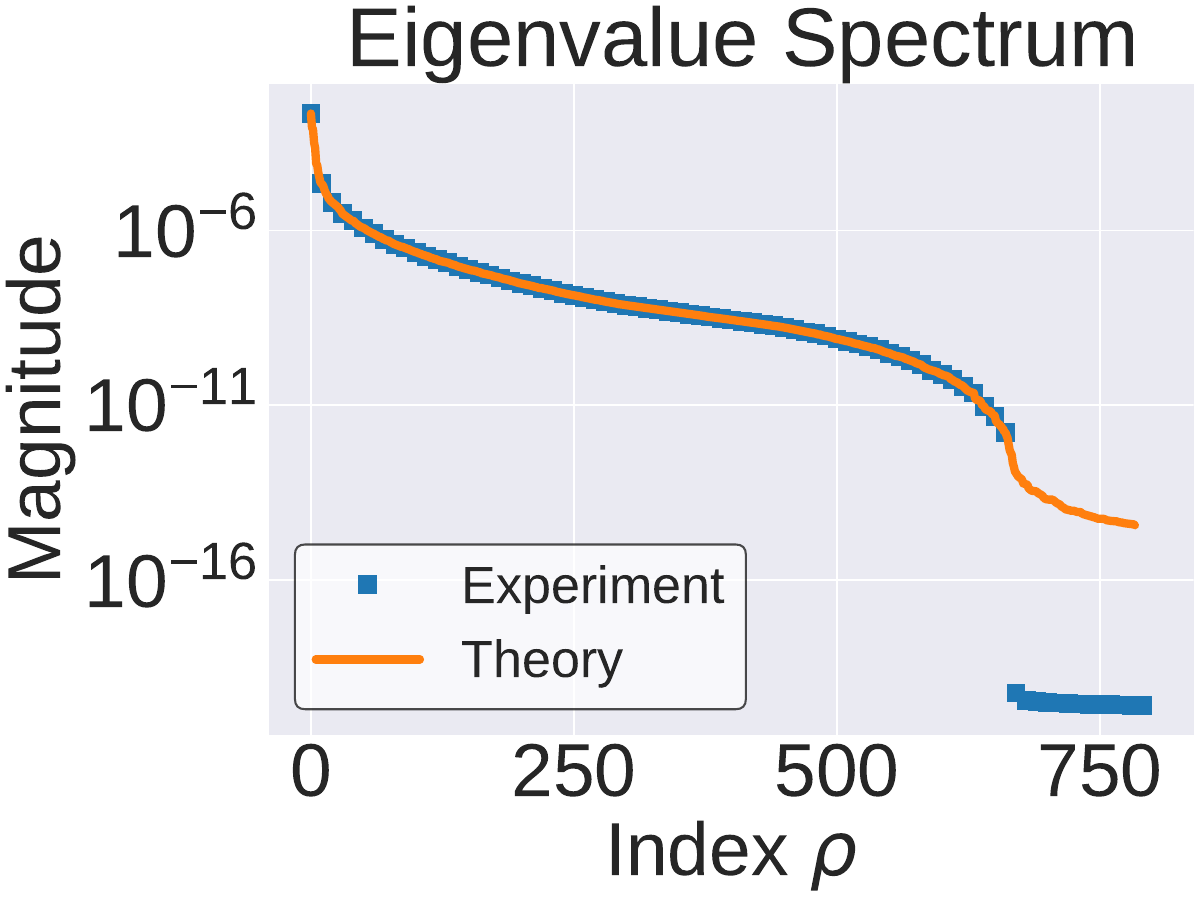}
  \put(1,80){\colorbox{white}{(a)}}
\end{overpic}%
}\hspace{-0.6em}%
\subfigure{%
\begin{overpic}[width=0.26\textwidth]{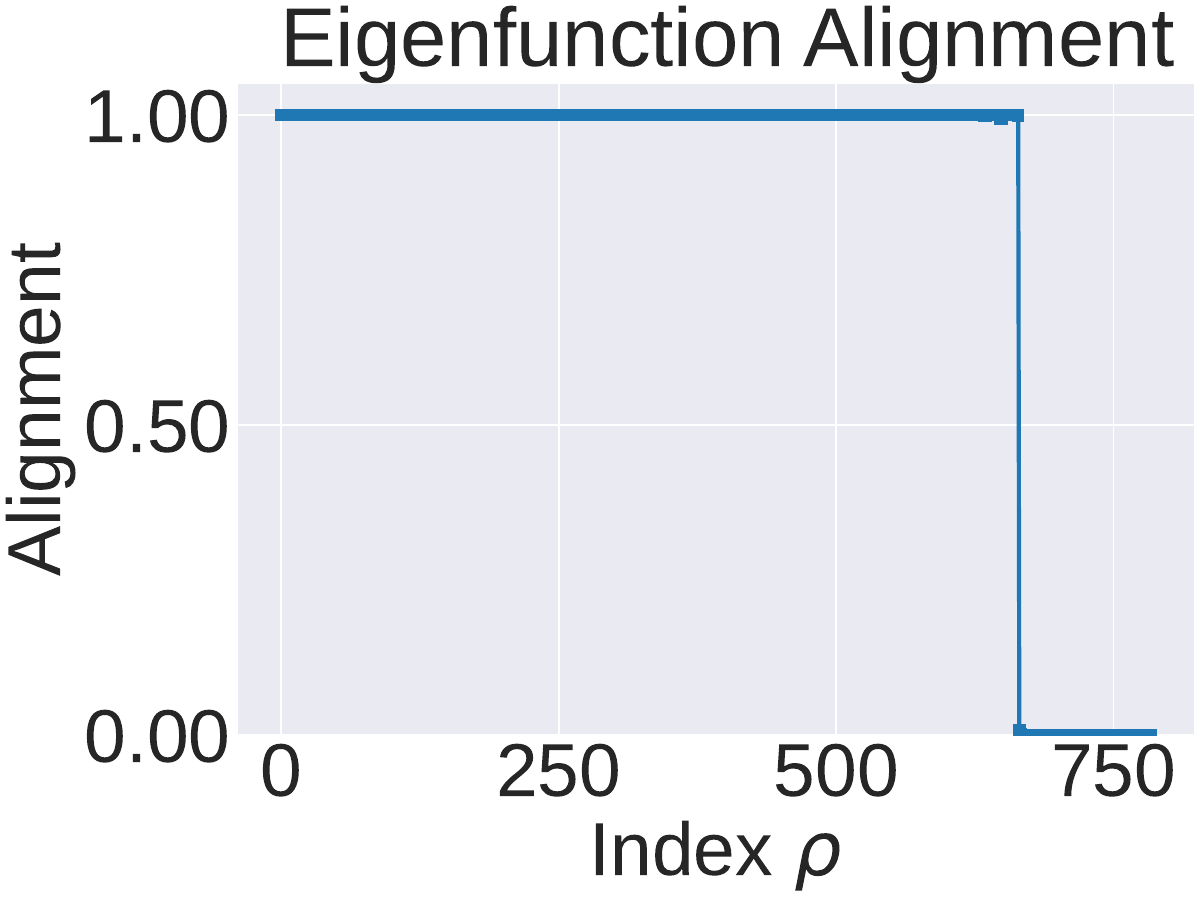}
  \put(1,80){\colorbox{white}{(b)}}
\end{overpic}%
}\hspace{-0.6em}%
\subfigure{%
\begin{overpic}[width=0.26\textwidth]{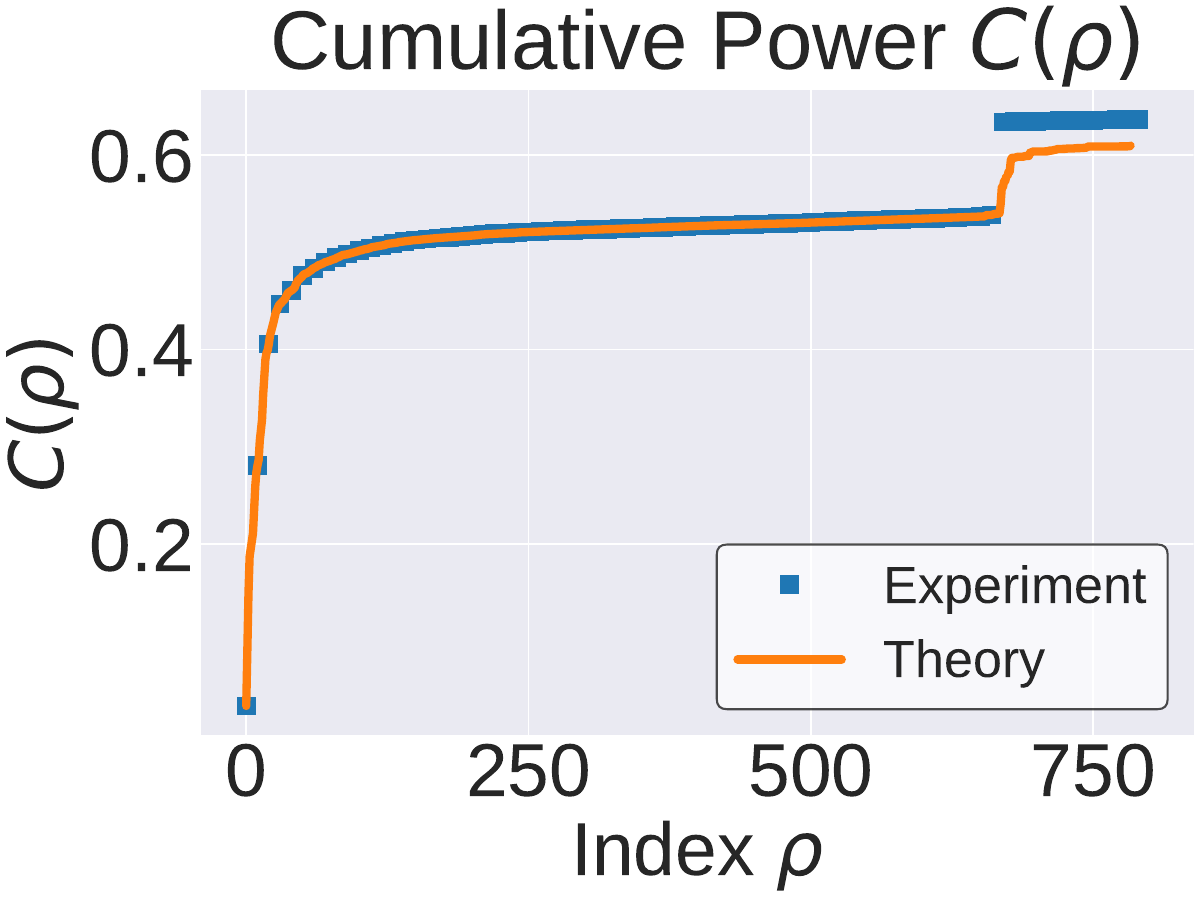}
  \put(1,80){\colorbox{white}{(c)}}
\end{overpic}%
}\hspace{-0.6em}%
\subfigure{%
\begin{overpic}[width=0.26\textwidth]{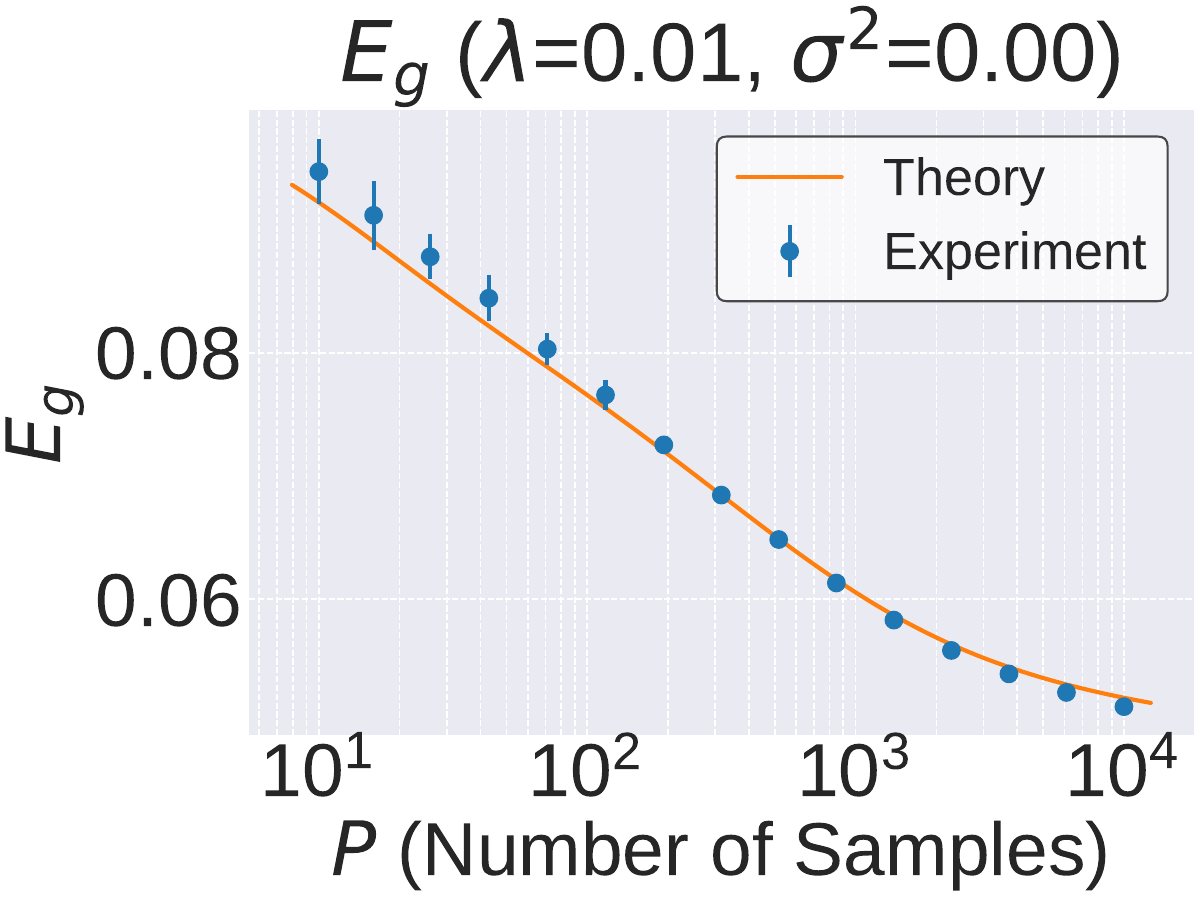}
  \put(1,80){\colorbox{white}{(d)}}
\end{overpic}%
}
\caption{Sanity check of the SSM-induced kernel eigendecomposition.
(a) The empirical eigenvalue spectrum matches the theoretical prediction
$\eta_\rho=s_\rho^2$ obtained from the singular values of the Toeplitz convolution operator.
(b) The empirical and predicted eigenfunctions are well aligned across most modes.
(c) The predicted eigenbasis accurately captures the cumulative task power $C(\rho)$.
(d) Kernel regression generalization predicted from the eigenspectrum matches empirical KRR performance across sample sizes. Error bars indicate standard
deviations over 50 independent runs.
}
\label{fig:sanity_unwhitened}
\end{figure}

Figure~\ref{fig:sanity_unwhitened} reports the eigenvalue spectrum, eigenfunction agreement, squared target weights, cumulative power, and kernel regression generalization on unwhitened inputs. Compared with the
whitened setting, the cumulative power $C(\rho)$ often rises faster because the input covariance concentrates task-relevant variation into fewer effective directions. This can lead to better empirical generalization, but it also mixes the inductive bias of the SSM with the structure of the input distribution. Therefore, we use whitened inputs in the main text to cleanly validate Theorem~\ref{Theorem: eigen}.

\paragraph{Finite-sample rank and empirical tail.}
Theorem~\ref{Theorem: eigen} states that the SSM-induced kernel has rank at most
$L$, because the feature map
$\Phi_{\mathrm{SSM}}(\pmb{u})=\pmb{T}_h\pmb{u}$ lies in an
$L$-dimensional feature space. In the empirical Gram-matrix eigendecomposition,
however, eigenvalues beyond the theoretical rank are not exactly zero. This
small tail arises from finite-sample estimation, numerical precision, and
preprocessing operations such as whitening and dimensionality reduction.

Although the tail eigenvalues are small, their cumulative contribution to the
empirical target expansion is not negligible. In the kernel generalization
prediction, the leading $L$ modes are described by the theoretical
SSM-induced eigenspectrum, while the residual empirical tail behaves like an
additional weak-noise component. Concretely, in Figure~\ref{fig:sanity}(d), the
theoretical curve is computed from the predicted leading spectrum together with
a constant offset accounting for the cumulative contribution of the empirical
tail. This offset is non-trivial even though each individual tail eigenvalue is
small, because it aggregates over many weak directions. This empirical tail does not contradict Theorem~\ref{Theorem: eigen}; rather,
it reflects the difference between the ideal finite-dimensional kernel operator
and its finite-sample numerical estimate. Similar finite-sample spectral tails
and their effect on kernel learning curves have also been observed in prior
kernel eigenlearning analyses~\citep{canatar2021spectral,Canatar2022BandwidthEG,
simon2023eigenlearningframeworkconservationlaw}.

\subsection{Kernel Regression Experimental Details}
\label{appendix: exp2 details}
\begin{figure}[htbp]
    \centering
    \includegraphics[width=0.9\linewidth]{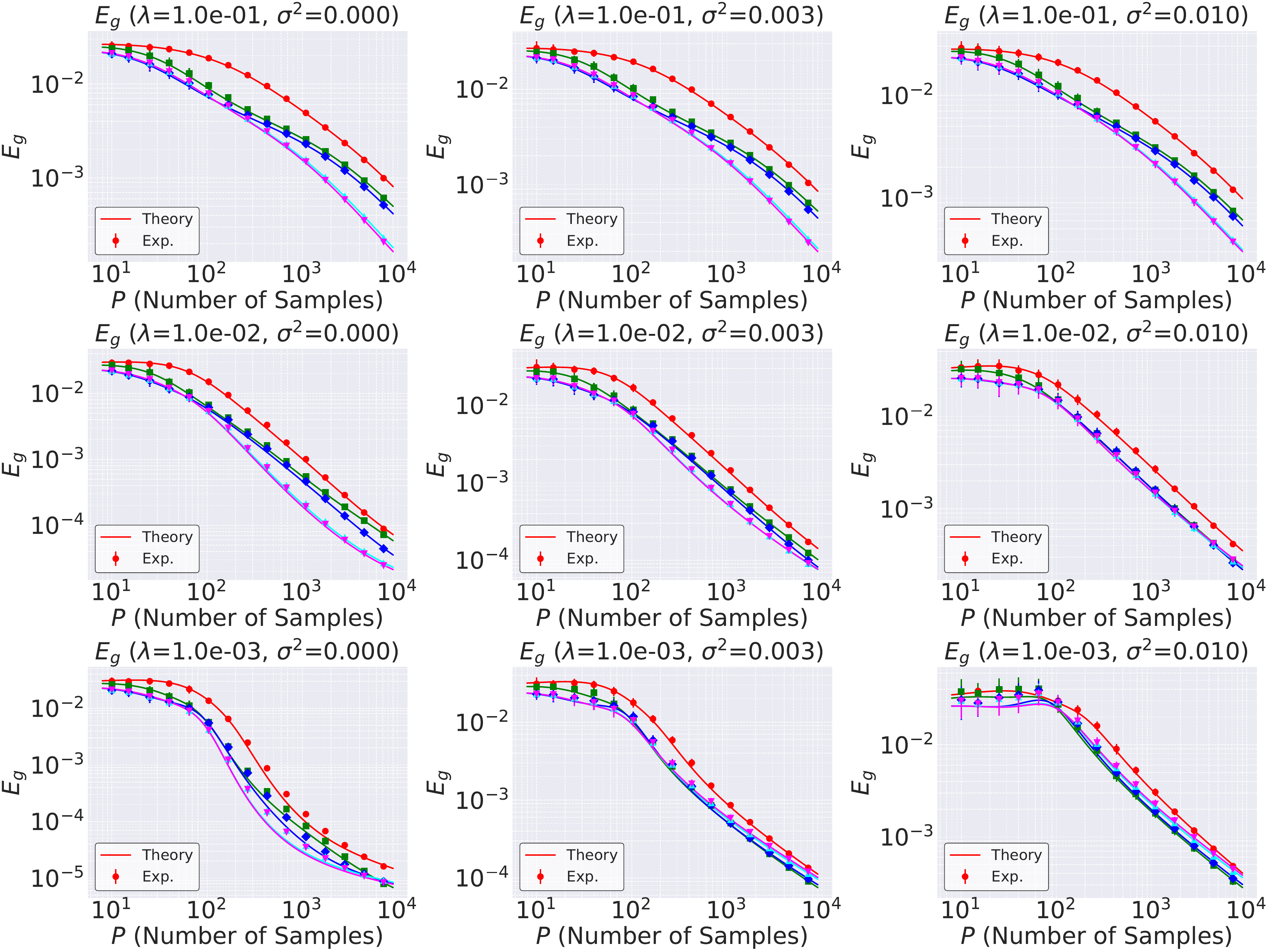}
    \caption{Learning curves of \textit{matched} case under more parameter settings.}
    \label{fig: exp2 match appendix}
\end{figure}

\begin{figure}[htbp]
    \centering
    \includegraphics[width=0.9\linewidth]{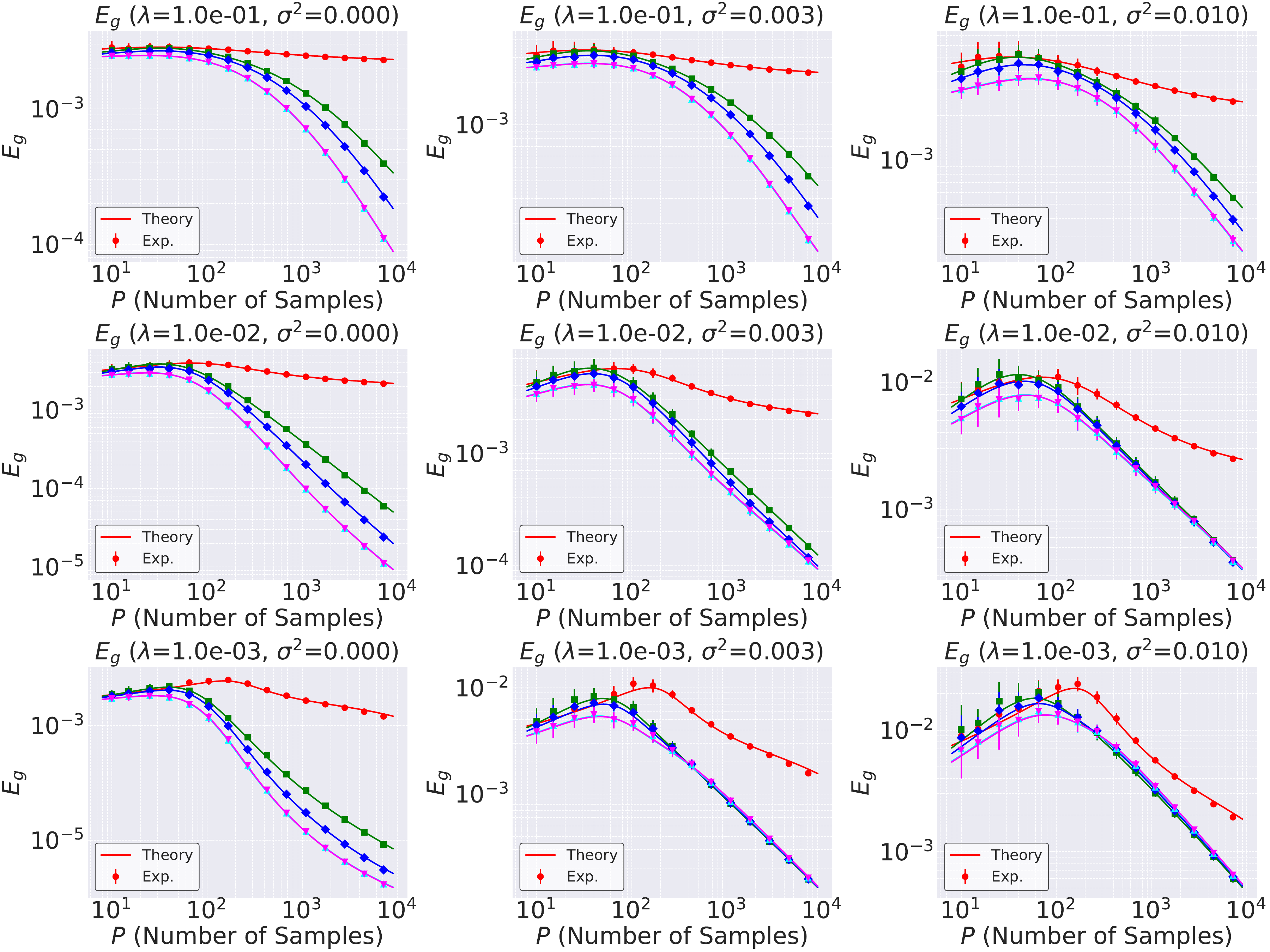}
    \caption{Learning curves of \textit{mismatched} case under more parameter settings.}
    \label{fig: exp2 mismatch appendix}
\end{figure}

\paragraph{Synthetic task generation.}
For the synthetic low- and high-frequency tasks, we generate labels based on MNIST images as follows. Let $\mathbf U\in\mathbb{R}^{P\times L}$ denote the mean-centered MNIST data matrix, with each image reshaped into a vector. We define a temporal grid $t\in[0,1]^L$ and construct frequency patterns corresponding to low- or high-frequency components,
\begin{align}
p_\text{low}(t)&=\cos(2\pi\cdot 20t)+\cos(2\pi\cdot40t),\notag\\
p_\text{high}(t)&= \cos(2\pi\cdot300t)+\sin(2\pi\cdot350t)\notag,
\end{align}
which are $L_2$-normalized and used to generate labels as $\mathbf{Y}=\mathbf{U}p$. This procedure ensures that the synthetic labels inherit the desired frequency structure while preserving the input correlations of MNIST.
\paragraph{Design of low- and high-frequency biased SSMs.}
In this experiment, the SSMs are parameterized as S4D~\citep{gu2022parameterizationinitializationdiagonalstate}, including $\pmb A=-\exp{\log{(-\operatorname{Re}\pmb A)}}+i\operatorname{Im}\pmb A$, output matrix $\pmb C$, and time step $\Delta=\exp{\log{\Delta}}$, where $\log{(-\operatorname{Re}\pmb A)}$, $\operatorname{Im}\pmb A$, $\pmb{C}$, and $\log{\Delta}$ are trainable parameters. 
The parameters of the low-frequency biased SSM are set by
\[
    \Delta = \frac{0.5}{N},\quad \operatorname{Im}\pmb A=\pi\cdot\{0, 1, \cdots,\lfloor n/2\rfloor\},
\]
and those of the high-frequency biased SSM are
\[
    \Delta=\frac{1}{N},\quad \operatorname{Im}\pmb A = \pi\cdot\{\lfloor n/2\rfloor,\cdots, n-1\},
\]
where 
\[
    \pmb C\sim\mathcal{N}(0,\pmb I),\quad \operatorname{Re}\pmb A = \{-0.5,-0.5,\cdots,-0.5\}
\]
are fixed.
\paragraph{Learning curves under more parameter settings.} We also provide additional figures for Section~\ref{subsec: exp2}, including learning curves under more parameter settings, shown in Figure~\ref{fig: exp2 match appendix} and Figure~\ref{fig: exp2 mismatch appendix} 
\subsection{Trainable One-Layer SSM Experimental Details}
\label{app:exp3_details}
This appendix summarizes the experimental setup used for the Standard RNN, the one-layer SSM baseline, and TDI. All methods were evaluated under the same data splits, training ratios, random seeds, and optimization protocol unless noted otherwise. The complete final accuracy and early-epoch summary table for Standard RNN, the SSM baseline, and TDI is provided in Appendix~\ref{app:one_layer_full_results}.

\paragraph{Model Structures.}
\subparagraph{Standard RNN.}
The Standard RNN branch uses a single-layer recurrent encoder followed by a two-layer MLP readout:
\begin{equation}
    h_t = \tanh\!\left(W_{xh} x_t + W_{hh} h_{t-1}\right), \qquad
    \hat{y} = \mathrm{MLP}(h_T).
\end{equation}
In the code, this is implemented with \texttt{nn.RNN} and a readout of the form
\texttt{Linear(128 $\to$ 64) + ReLU + Linear(64 $\to$ \#classes)}.
The hidden size is 128 and the number of recurrent layers is 1.
\subparagraph{Linear SSM baseline.}
The baseline uses the SSM implementation with a linear state update and a two-layer MLP readout:
\begin{equation}
    h_t = W h_{t-1} + W_{\mathrm{in}} x_t, \qquad
    \hat{y} = \mathrm{MLP}(h_T).
\end{equation}
SSM is initialized from the HiPPO \texttt{foud}~\citep{gu2020hippo} parameterization, discretized with zero-order hold, and then trained end-to-end.
In this setting, both \texttt{W} and \texttt{W\_in} are trainable.
\subparagraph{TDI.}
TDI uses the same architecture as the SSM baseline, but the state-space parameters are first adapted with task-dependent initialization before the supervised training stage. The implementation keeps the same hidden size, readout head, and optimization settings as the baseline. For details of TDI algorithm, see Appendix~\ref{Appendix: tdi-implementation}.

\paragraph{Dataset Statistics.}
All inputs are treated as sequences with input dimension 1. For MNIST, the image is flattened row-wise into a length-784 sequence; the remaining datasets are already time-series or are generated as one-dimensional sequences. The details of these datasets are summarized in Table~\ref{tab:appendix-dataset-statistics}.
\begin{table*}[t]
\centering
\caption{\textbf{Dataset statistics used in the one-layer trainable SSM experiments.} The reported train/test counts correspond to the full dataset split before applying any training ratio.}
\label{tab:appendix-dataset-statistics}
\small
\begin{tabular}{lrrrrr}
\toprule
Dataset & Train & Test & Classes & Seq. len. & Input dim. \\
\midrule
Binary Freq. & 40000 & 10000 & 2 & 256 & 1 \\
ECG5000 & 500 & 4500 & 5 & 140 & 1 \\
FordA & 3601 & 1320 & 2 & 500 & 1 \\
FordB & 3636 & 810 & 2 & 500 & 1 \\
MNIST & 60000 & 10000 & 10 & 784 & 1 \\
TwoPatterns & 1000 & 4000 & 4 & 128 & 1 \\
Wafer & 1000 & 6164 & 2 & 152 & 1 \\
\bottomrule
\end{tabular}%
\end{table*}
\paragraph{Binary Frequency Classification.}
We construct a synthetic binary sequence classification dataset in which the label is determined solely by the amplitude of a designated frequency component. Each example is a length-256 univariate sequence. The target sinusoid has frequency $3$ Hz and random phase, and its amplitude is sampled from a label-dependent interval to ensure class balance. Specifically, class 1 samples use amplitudes from $(0.6, 1.0)$, whereas class 0 samples use amplitudes from $(0.1, 0.4)$. We additionally superimpose distractor sinusoids at frequencies $1$, $7$, and $11$ Hz with random amplitudes and phases, together with Gaussian noise ($\sigma=0.1$). Each sequence is then normalized by its maximum absolute value.

\paragraph{Hyperparameters.}
We use random seeds $\{42,123,1024\}$ and training ratios
$\{0.01,0.02,0.04,0.08,0.16,0.32,0.64,1.00\}$ for all datasets. Other
hyperparameters are summarized in Table~\ref{tab:appendix-one-layer-ssm-hyperparameters}.
\begin{table*}[t]
\centering
\caption{
\textbf{Hyperparameters for the trainable one-layer SSM experiments.} The only
dataset-dependent training hyperparameter is the number of epochs.
}
\label{tab:appendix-one-layer-ssm-hyperparameters}
\small
\begin{minipage}[t]{0.35\textwidth}
\centering
\begin{tabular}{lc}
\toprule
Dataset & Epochs \\
\midrule
Binary Freq. & 15 \\
ECG5000 & 15 \\
FordA & 10 \\
FordB & 15 \\
MNIST & 15 \\
TwoPatterns & 15 \\
Wafer & 10 \\
\bottomrule
\end{tabular}
\end{minipage}
\hfill
\begin{minipage}[t]{0.64\textwidth}
\centering
\begin{tabular}{ll}
\toprule
Shared setting & Value \\
\midrule
Optimizer & AdamW \\
Learning rate & $10^{-3}$ \\
Weight decay & $10^{-2}$ \\
Gradient clipping & $1.0$ \\
Batch size & 128 \\
Hidden size & 128 \\
Spectral radius & 0.9 \\
Scheduler & cosine annealing, $\eta_{\min}=10^{-6}$ \\
\bottomrule
\end{tabular}
\end{minipage}
\end{table*}
\subsection{Deep SSM Experimental Details}
\label{app:deep_ssm_details}
This appendix summarizes the experimental setup used for the deep SSM baseline, TDI, and TDI-Frozen on Frequency Classification, sMNIST, pMNIST, PathFinder, CIFAR-10, ListOps, Speech Commands, and SC10. Unless otherwise noted, all methods were evaluated with the same data splits, training ratios, random seeds, optimizer, and learning-rate schedule.

\paragraph{Model Structures.}
\subparagraph{Deep SSM baseline.}
All deep SSM experiments use the S4D backbone with diagonal state-space parameterization, i.e., \texttt{diag-lin}~\citep{gu2022parameterizationinitializationdiagonalstate}. For real-valued sequence inputs, the model first applies a linear input projection to dimension $d_{\mathrm{model}}$, then processes the sequence with a stack of residual S4 blocks, and finally applies a sequence-classification decoder. For Frequency Classification, sMNIST, pMNIST, PathFinder, CIFAR-10, Speech Commands, and SC10, the decoder uses temporal mean pooling followed by a linear classifier. For ListOps, the pipeline uses an embedding encoder together with a length-aware last-token decoder inherited from the LRA text setup. In all experiments, only the first S4 layer is expanded to \texttt{n\_ssm}$=d_{\mathrm{model}}$, while the remaining layers keep the default setting.

\subparagraph{TDI.}
TDI uses the same deep SSM backbone as the baseline, but adapts the first S4 layer with task-dependent initialization before supervised training. In the implementation, the task-spectrum configuration is injected only into the first S4 layer, while all later layers use the baseline initialization and are trained normally. All TDI runs use Fisher log-power as the task-spectrum estimator and Algorithm~\ref{alg:tdi-complex} for initialization. For numerical inputs, we estimate the task spectrum before the time-shared pointwise linear encoder, which mixes channels without mixing distinct temporal frequencies. For ListOps, we instead estimate the spectrum from numerical token IDs before embedding. This is a heuristic that depends on the arbitrary token numbering: token IDs have no intrinsic metric or spectral semantics, and their spectrum should not be interpreted as the spectrum of the learned embeddings.

\subparagraph{TDI-Frozen.}
TDI-Frozen uses the same task-dependent initialization as TDI for the first S4 layer, but freezes the initialized first-layer S4 parameters during supervised training. All later layers, the decoder, and the remaining trainable parameters are trained normally. This provides a control for whether a fixed task-aligned spectral filter is sufficient, or whether allowing the TDI-initialized layer to
adapt during downstream training is important.

\paragraph{Dataset Statistics.}
All tasks are treated as sequences. sMNIST is the standard sequential MNIST task obtained by flattening each image row-wise into a length-784 sequence. The pMNIST script uses the same 784-pixel sequence together with a fixed permutation implemented by the codebase. CIFAR-10 refers to grayscale sequential CIFAR-10: each $32\times 32$ image is converted to grayscale and flattened into a length-1024 sequence. PathFinder uses $32\times32$ grayscale images and flattens them into length-1024 sequences. ListOps uses token sequences truncated to length 2048 with EOS appended. Speech Commands and SC10 use raw one-second audio waveforms of length 16000.

\begin{table*}[t]
\centering
\caption{\textbf{Dataset statistics used in the deep SSM experiments.} The reported train/validation/test counts correspond to the full split before applying any training subset ratio.}
\label{tab:appendix-deep-ssm-dataset-statistics}
\small
\begin{tabular}{lrrrrrr}
\toprule
Dataset & Train & Val & Test & Classes & Seq. len. & Input dim. \\
\midrule
Freq. Classification & 10000 & 1000 & 1000 & 10 & 1024 & 1 \\
sMNIST & 54000 & 6000 & 10000 & 10 & 784 & 1 \\
pMNIST & 54000 & 6000 & 10000 & 10 & 784 & 1 \\
PathFinder & 480001 & 59999 & 59999 & 2 & 1024 & 1 \\
CIFAR-10 & 45000 & 5000 & 10000 & 10 & 1024 & 1 \\
ListOps & 96000 & 2000 & 2000 & 10 & 2048 & -- \\
Speech Commands & 84843 & 9981 & 11005 & 35 & 16000 & 1 \\
SC10 & 24482 & 5246 & 5247 & 10 & 16000 & 1 \\
\bottomrule
\end{tabular}
\end{table*}

\paragraph{Frequency Classification.}
For Frequency Classification, we use the synthetic dataset where each sample is a length-1024 univariate signal. The class label determines a frequency band within the interval $[4,64]$ Hz, which is partitioned into 10 bins. For each example, the principal sinusoid frequency is sampled uniformly within the class-specific bin, its amplitude is sampled from $[0.8,1.2]$, and its phase is random. The signal is further modulated by a low-frequency envelope with modulation depth $0.15$ and modulation frequency sampled from $[0.25,2.0]$ Hz. We additionally add a second harmonic with scale $0.3$, a low-frequency drift term with scale $0.1$, and Gaussian noise with standard deviation $0.15$. Each sequence is finally normalized to zero mean and unit variance.

\paragraph{Hyperparameters.}
All runs use the five random seeds
$\{0,42,218,18,2218\}$.
The training ratios are logarithmically spaced between $10^{-3}$ and $1$. PathFinder, ListOps, Speech Commands, and SC10 use
\[
\{0.00100, 0.00215, 0.00464, 0.01000, 0.02154, 0.04642, 0.10000, 0.21544, 0.46416, 1.00000\},
\]
Frequency Classification drops the smallest point and uses the remaining 9 ratios, and sMNIST, pMNIST, and CIFAR-10 use the last 8 ratios
\[
\{0.00464, 0.01000, 0.02154, 0.04642, 0.10000, 0.21544, 0.46416, 1.00000\}.
\]

All experiments use AdamW with cosine annealing and $\eta_{\min}=10^{-6}$. No gradient clipping is applied. Other hyperparameters are summarized in Table~\ref{tab:appendix-deep-ssm-hparams}.
\begin{table*}[t]
\centering
\caption{\textbf{Hyperparameters used in the deep SSM experiments.} LR is learning rate and WD is weight decay. BN and LN refer to Batch Normalization and Layer Normalization, respectively; Pre-norm indicates whether normalization is applied before the S4 block.}
\label{tab:appendix-deep-ssm-hparams}
\small
\resizebox{\textwidth}{!}{%
\begin{tabular}{lccccccccccc}
\toprule
Dataset & Depth & Features $H$ & State Size $N$ & Norm & Pre-norm & Dropout & LR & Batch Size & Epochs & WD & $(\Delta_{\min}, \Delta_{\max})$ \\
\midrule
Freq. Classification & 2 & 64  & 64 & BN & True  & 0   & 0.005 & 64  & 10 & 0.01 & $(0.001, 0.1)$ \\
sMNIST                & 4 & 256 & 64 & LN & False & 0.1 & 0.01  & 128 & 30 & 0.05 & $(0.001, 0.1)$ \\
pMNIST                & 4 & 256 & 64 & LN & False & 0.1 & 0.01  & 128 & 30 & 0.05 & $(0.001, 0.1)$ \\
CIFAR-10                 & 4 & 256 & 64 & LN & False & 0.1 & 0.01  & 128 & 30 & 0.05 & $(0.001, 0.1)$ \\
Speech Commands       & 4 & 256 & 64 & BN & False & 0   & 0.01  & 16  & 30 & 0.05 & $(0.001, 0.1)$ \\
SC10                  & 4 & 256 & 64 & BN & False & 0   & 0.01  & 16  & 30 & 0.05 & $(0.001, 0.1)$ \\
ListOps               & 4 & 256 & 64 & BN & False & 0   & 0.01  & 50  & 30 & 0.05 & $(0.001, 0.1)$ \\
PathFinder            & 4 & 256 & 64 & BN & True  & 0   & 0.004 & 70  & 30 & 0.03 & $(0.001, 0.1)$ \\
\bottomrule
\end{tabular}
}
\end{table*}
\subsection{Compute Resources}
Kernel regression and one-layer SSM experiments were run on NVIDIA RTX 4090 GPUs. Deep SSM experiments were run on NVIDIA RTX 4090 and RTX 5090 GPUs. Each RTX 4090/5090 GPU has 24GB/32GB memory. Each training job used a single GPU and corresponded to one dataset, one training ratio, one random seed, and one model variant. The one-layer SSM experiments evaluated 7 datasets, 8 training ratios, and 3 random seeds, while the deep SSM experiments evaluated 8 datasets, 10 training ratios, and 5 random seeds. Kernel regression runs take negligible compute time. Typical one-layer SSM runs take roughly 1--5 GPU-hours, while deep SSM runs take roughly 1--30 GPU-hours depending on dataset and training ratio. Across all experiments reported in the paper, including baseline, TDI, and TDI-Frozen, the total compute budget was approximately 400--500 GPU-hours. TDI adds only a task-spectrum estimation and initialization step before supervised training and does not change the architecture or inference cost.
\subsection{Code Availability and Reproducibility}
The code for reproducing our experiments is publicly available at \url{https://github.com/Criticality-Cognitive-Computation-Lab/Aligning-Inductive-Bias-for-State-Space-Models}. The repository contains the implementation of TDI, baseline training code, configuration files, and scripts for reproducing the main experiments reported in Section~\ref{sec: experiments}. The included README describes the software environment, dataset preparation steps, and example commands for running the kernel regression, one-layer SSM, and deep SSM experiments.

\section{Additional Experimental Results}
\subsection{Complete Results for Trainable One-Layer SSMs}
In this section, we provide full per-dataset, per-ratio comparisons among the SSM baseline, TDI, and the Standard RNN baseline in Table~\ref{tab:appendix-complete-standard-baseline-tdi}.
\label{app:one_layer_full_results}
{
\captionsetup{font=normalsize}
\scriptsize
\setlength{\tabcolsep}{3.5pt}
\renewcommand{\arraystretch}{1.05}
\begin{scriptsize}
\begin{longtable}{llcccccc}
\caption{\textbf{Complete final performance for Standard RNN, the SSM baseline, and TDI.} The left block reports final test accuracy; the right block reports mean test accuracy over the first 5 epochs. Each cell reports mean $\pm$ standard deviation over 3 seeds. The best value within each block for each dataset-ratio row is bolded, and the second-best value is underlined.}\label{tab:appendix-complete-standard-baseline-tdi}\\
\toprule
& & \multicolumn{3}{c}{Final test accuracy} & \multicolumn{3}{c}{First-5 test accuracy} \\
\cmidrule(lr){3-5} \cmidrule(lr){6-8}
Dataset & Ratio & Std. RNN & Baseline & TDI & Std. RNN & Baseline & TDI \\
\midrule
\endfirsthead
\toprule
& & \multicolumn{3}{c}{Final test accuracy} & \multicolumn{3}{c}{First-5 test accuracy} \\
\cmidrule(lr){3-5} \cmidrule(lr){6-8}
Dataset & Ratio & Std. RNN & Baseline & TDI & Std. RNN & Baseline & TDI \\
\midrule
\endhead
Binary Freq. & 0.01 & $50.4 \pm 0.9$\% & \underline{$73.0 \pm 4.8$\%} & \boldmath\textbf{$79.1 \pm 0.9$\%}\unboldmath & $50.2 \pm 0.7$\% & \underline{$55.1 \pm 3.5$\%} & \boldmath\textbf{$58.6 \pm 7.3$\%}\unboldmath \\
Binary Freq. & 0.02 & $53.6 \pm 2.5$\% & \underline{$83.6 \pm 5.4$\%} & \boldmath\textbf{$84.2 \pm 1.3$\%}\unboldmath & $50.8 \pm 0.8$\% & \boldmath\textbf{$55.9 \pm 3.9$\%}\unboldmath & \underline{$54.1 \pm 4.1$\%} \\
Binary Freq. & 0.04 & $52.9 \pm 3.8$\% & \underline{$86.6 \pm 2.2$\%} & \boldmath\textbf{$89.9 \pm 5.5$\%}\unboldmath & $51.0 \pm 1.0$\% & \underline{$65.2 \pm 12.1$\%} & \boldmath\textbf{$70.9 \pm 4.4$\%}\unboldmath \\
Binary Freq. & 0.08 & $50.4 \pm 1.7$\% & \underline{$92.5 \pm 3.6$\%} & \boldmath\textbf{$94.9 \pm 0.7$\%}\unboldmath & $51.2 \pm 0.9$\% & \underline{$70.7 \pm 3.5$\%} & \boldmath\textbf{$79.1 \pm 2.1$\%}\unboldmath \\
Binary Freq. & 0.16 & $58.9 \pm 11.8$\% & \boldmath\textbf{$96.7 \pm 0.8$\%}\unboldmath & \underline{$95.0 \pm 2.2$\%} & $51.8 \pm 2.1$\% & \underline{$77.6 \pm 4.6$\%} & \boldmath\textbf{$78.5 \pm 5.9$\%}\unboldmath \\
Binary Freq. & 0.32 & $65.8 \pm 18.6$\% & \boldmath\textbf{$97.5 \pm 0.2$\%}\unboldmath & \underline{$97.0 \pm 0.8$\%} & $53.2 \pm 1.1$\% & \underline{$87.5 \pm 0.8$\%} & \boldmath\textbf{$87.6 \pm 2.1$\%}\unboldmath \\
Binary Freq. & 0.64 & \underline{$51.4 \pm 1.9$\%} & \boldmath\textbf{$97.1 \pm 0.8$\%}\unboldmath & \boldmath\textbf{$97.1 \pm 0.4$\%}\unboldmath & $51.1 \pm 1.2$\% & \underline{$89.3 \pm 1.8$\%} & \boldmath\textbf{$89.5 \pm 3.2$\%}\unboldmath \\
Binary Freq. & 1.00 & $50.8 \pm 1.5$\% & \boldmath\textbf{$97.8 \pm 0.0$\%}\unboldmath & \underline{$97.7 \pm 0.0$\%} & $50.9 \pm 1.2$\% & \boldmath\textbf{$92.3 \pm 0.9$\%}\unboldmath & \underline{$91.5 \pm 1.4$\%} \\
\midrule
ECG5000 & 0.01 & \boldmath\textbf{$55.1 \pm 22.9$\%}\unboldmath & \underline{$48.3 \pm 10.9$\%} & \boldmath\textbf{$55.2 \pm 21.2$\%}\unboldmath & \underline{$32.2 \pm 1.5$\%} & $21.5 \pm 13.7$\% & \boldmath\textbf{$42.4 \pm 9.2$\%}\unboldmath \\
ECG5000 & 0.02 & \underline{$81.1 \pm 0.8$\%} & $62.1 \pm 1.5$\% & \boldmath\textbf{$85.5 \pm 4.8$\%}\unboldmath & \boldmath\textbf{$38.7 \pm 8.2$\%}\unboldmath & \underline{$29.9 \pm 17.1$\%} & $28.1 \pm 17.8$\% \\
ECG5000 & 0.04 & \boldmath\textbf{$81.6 \pm 1.7$\%}\unboldmath & $65.1 \pm 17.7$\% & \underline{$77.5 \pm 8.4$\%} & \underline{$42.3 \pm 9.0$\%} & $31.6 \pm 16.3$\% & \boldmath\textbf{$50.4 \pm 11.6$\%}\unboldmath \\
ECG5000 & 0.08 & \boldmath\textbf{$80.8 \pm 1.2$\%}\unboldmath & $74.4 \pm 2.8$\% & \underline{$76.6 \pm 4.3$\%} & \boldmath\textbf{$41.7 \pm 8.3$\%}\unboldmath & $31.5 \pm 17.2$\% & \underline{$32.3 \pm 19.5$\%} \\
ECG5000 & 0.16 & \boldmath\textbf{$80.3 \pm 0.7$\%}\unboldmath & $64.8 \pm 3.3$\% & \underline{$69.2 \pm 5.3$\%} & \boldmath\textbf{$42.1 \pm 9.5$\%}\unboldmath & \underline{$34.0 \pm 19.5$\%} & $30.0 \pm 20.9$\% \\
ECG5000 & 0.32 & $86.8 \pm 0.2$\% & \underline{$88.9 \pm 2.1$\%} & \boldmath\textbf{$89.5 \pm 1.4$\%}\unboldmath & \boldmath\textbf{$62.1 \pm 3.6$\%}\unboldmath & \underline{$57.2 \pm 12.7$\%} & $55.7 \pm 8.4$\% \\
ECG5000 & 0.64 & $77.8 \pm 15.8$\% & \underline{$90.4 \pm 0.8$\%} & \boldmath\textbf{$90.9 \pm 0.9$\%}\unboldmath & \boldmath\textbf{$63.6 \pm 2.4$\%}\unboldmath & \underline{$63.3 \pm 7.7$\%} & $54.5 \pm 16.3$\% \\
ECG5000 & 1.00 & $78.9 \pm 16.3$\% & \boldmath\textbf{$91.5 \pm 0.3$\%}\unboldmath & \underline{$91.1 \pm 0.3$\%} & $59.4 \pm 12.3$\% & \underline{$67.9 \pm 7.1$\%} & \boldmath\textbf{$75.6 \pm 7.2$\%}\unboldmath \\
\midrule
FordA & 0.01 & \boldmath\textbf{$50.8 \pm 1.3$\%}\unboldmath & \underline{$50.5 \pm 3.1$\%} & $50.0 \pm 2.5$\% & \boldmath\textbf{$51.2 \pm 0.5$\%}\unboldmath & $50.3 \pm 1.7$\% & \underline{$50.4 \pm 1.0$\%} \\
FordA & 0.02 & \underline{$50.8 \pm 1.1$\%} & $50.0 \pm 3.5$\% & \boldmath\textbf{$52.8 \pm 2.9$\%}\unboldmath & \underline{$51.3 \pm 0.4$\%} & \boldmath\textbf{$51.4 \pm 2.7$\%}\unboldmath & $50.0 \pm 1.8$\% \\
FordA & 0.04 & $49.2 \pm 1.0$\% & \underline{$51.4 \pm 2.9$\%} & \boldmath\textbf{$52.3 \pm 2.8$\%}\unboldmath & $50.4 \pm 0.8$\% & \underline{$51.0 \pm 2.4$\%} & \boldmath\textbf{$51.1 \pm 0.6$\%}\unboldmath \\
FordA & 0.08 & $48.5 \pm 1.8$\% & \underline{$53.0 \pm 3.1$\%} & \boldmath\textbf{$59.5 \pm 1.9$\%}\unboldmath & $50.1 \pm 0.8$\% & \underline{$51.6 \pm 2.8$\%} & \boldmath\textbf{$51.8 \pm 0.8$\%}\unboldmath \\
FordA & 0.16 & $49.5 \pm 1.2$\% & \underline{$55.9 \pm 3.4$\%} & \boldmath\textbf{$69.5 \pm 2.9$\%}\unboldmath & $50.0 \pm 0.5$\% & \underline{$51.5 \pm 1.6$\%} & \boldmath\textbf{$57.6 \pm 2.4$\%}\unboldmath \\
FordA & 0.32 & $51.8 \pm 2.4$\% & \underline{$65.1 \pm 0.9$\%} & \boldmath\textbf{$66.1 \pm 5.2$\%}\unboldmath & $51.3 \pm 1.5$\% & \underline{$55.0 \pm 1.3$\%} & \boldmath\textbf{$58.3 \pm 3.3$\%}\unboldmath \\
FordA & 0.64 & $53.7 \pm 3.5$\% & \underline{$67.1 \pm 5.9$\%} & \boldmath\textbf{$71.1 \pm 4.7$\%}\unboldmath & $51.7 \pm 1.3$\% & \underline{$57.8 \pm 1.0$\%} & \boldmath\textbf{$62.3 \pm 7.5$\%}\unboldmath \\
FordA & 1.00 & $54.9 \pm 6.1$\% & \boldmath\textbf{$74.7 \pm 3.7$\%}\unboldmath & \underline{$73.9 \pm 3.0$\%} & $52.1 \pm 1.0$\% & \boldmath\textbf{$61.4 \pm 6.0$\%}\unboldmath & \underline{$59.1 \pm 4.5$\%} \\
\midrule
FordB & 0.01 & $50.2 \pm 0.5$\% & \underline{$50.9 \pm 0.9$\%} & \boldmath\textbf{$51.5 \pm 1.2$\%}\unboldmath & $49.6 \pm 0.3$\% & \boldmath\textbf{$50.7 \pm 0.2$\%}\unboldmath & \underline{$50.5 \pm 1.2$\%} \\
FordB & 0.02 & $49.6 \pm 1.6$\% & \boldmath\textbf{$55.0 \pm 2.0$\%}\unboldmath & \underline{$52.4 \pm 1.8$\%} & $50.0 \pm 1.1$\% & \boldmath\textbf{$52.3 \pm 2.0$\%}\unboldmath & \underline{$50.6 \pm 0.5$\%} \\
FordB & 0.04 & $49.5 \pm 1.8$\% & \underline{$51.3 \pm 1.8$\%} & \boldmath\textbf{$52.6 \pm 2.5$\%}\unboldmath & $50.0 \pm 1.0$\% & \underline{$51.3 \pm 0.9$\%} & \boldmath\textbf{$52.1 \pm 1.0$\%}\unboldmath \\
FordB & 0.08 & $49.6 \pm 0.7$\% & \underline{$52.2 \pm 2.5$\%} & \boldmath\textbf{$59.5 \pm 0.7$\%}\unboldmath & \underline{$49.9 \pm 0.5$\%} & $49.8 \pm 0.4$\% & \boldmath\textbf{$50.6 \pm 0.6$\%}\unboldmath \\
FordB & 0.16 & $50.0 \pm 1.3$\% & \underline{$58.0 \pm 4.4$\%} & \boldmath\textbf{$59.1 \pm 2.0$\%}\unboldmath & $50.0 \pm 0.5$\% & \underline{$51.1 \pm 0.9$\%} & \boldmath\textbf{$52.5 \pm 2.3$\%}\unboldmath \\
FordB & 0.32 & $52.4 \pm 0.2$\% & \underline{$58.7 \pm 3.7$\%} & \boldmath\textbf{$62.8 \pm 1.5$\%}\unboldmath & $50.2 \pm 0.8$\% & \boldmath\textbf{$53.2 \pm 1.4$\%}\unboldmath & \underline{$52.6 \pm 1.8$\%} \\
FordB & 0.64 & $51.1 \pm 2.4$\% & \underline{$60.5 \pm 0.5$\%} & \boldmath\textbf{$60.6 \pm 5.0$\%}\unboldmath & $50.4 \pm 0.7$\% & \boldmath\textbf{$56.4 \pm 1.1$\%}\unboldmath & \underline{$53.2 \pm 3.6$\%} \\
FordB & 1.00 & $52.7 \pm 7.0$\% & \boldmath\textbf{$62.6 \pm 1.6$\%}\unboldmath & \underline{$61.7 \pm 1.0$\%} & $49.6 \pm 0.9$\% & \boldmath\textbf{$57.1 \pm 1.4$\%}\unboldmath & \underline{$52.9 \pm 2.8$\%} \\
\midrule
MNIST & 0.01 & $10.3 \pm 0.9$\% & \underline{$42.8 \pm 2.3$\%} & \boldmath\textbf{$60.5 \pm 4.9$\%}\unboldmath & $10.5 \pm 0.9$\% & \boldmath\textbf{$22.5 \pm 2.7$\%}\unboldmath & \underline{$18.4 \pm 3.3$\%} \\
MNIST & 0.02 & $11.3 \pm 0.0$\% & \underline{$51.7 \pm 6.8$\%} & \boldmath\textbf{$72.8 \pm 4.7$\%}\unboldmath & $10.9 \pm 0.7$\% & \boldmath\textbf{$29.0 \pm 4.8$\%}\unboldmath & \underline{$23.7 \pm 7.2$\%} \\
MNIST & 0.04 & $10.8 \pm 1.0$\% & \underline{$64.2 \pm 3.8$\%} & \boldmath\textbf{$81.1 \pm 2.0$\%}\unboldmath & $10.8 \pm 1.0$\% & \boldmath\textbf{$43.3 \pm 0.9$\%}\unboldmath & \underline{$36.7 \pm 3.8$\%} \\
MNIST & 0.08 & $12.4 \pm 1.8$\% & \underline{$72.5 \pm 3.5$\%} & \boldmath\textbf{$88.4 \pm 1.2$\%}\unboldmath & $11.3 \pm 0.1$\% & \boldmath\textbf{$51.1 \pm 0.6$\%}\unboldmath & \underline{$49.9 \pm 11.0$\%} \\
MNIST & 0.16 & $11.3 \pm 0.0$\% & \underline{$74.2 \pm 7.0$\%} & \boldmath\textbf{$89.2 \pm 2.6$\%}\unboldmath & $11.1 \pm 0.2$\% & \boldmath\textbf{$60.0 \pm 5.5$\%}\unboldmath & \underline{$56.7 \pm 7.4$\%} \\
MNIST & 0.32 & $18.7 \pm 12.8$\% & \underline{$84.3 \pm 6.9$\%} & \boldmath\textbf{$90.1 \pm 5.7$\%}\unboldmath & $11.3 \pm 0.2$\% & \underline{$67.1 \pm 6.0$\%} & \boldmath\textbf{$72.0 \pm 12.1$\%}\unboldmath \\
MNIST & 0.64 & $11.3 \pm 0.0$\% & \underline{$89.2 \pm 2.1$\%} & \boldmath\textbf{$92.2 \pm 3.3$\%}\unboldmath & $11.2 \pm 0.2$\% & \boldmath\textbf{$72.2 \pm 4.3$\%}\unboldmath & \underline{$69.1 \pm 15.6$\%} \\
MNIST & 1.00 & $11.3 \pm 0.0$\% & \underline{$64.5 \pm 47.3$\%} & \boldmath\textbf{$82.1 \pm 5.2$\%}\unboldmath & $11.1 \pm 0.4$\% & \underline{$66.4 \pm 20.1$\%} & \boldmath\textbf{$67.6 \pm 5.1$\%}\unboldmath \\
\midrule
TwoPatterns & 0.01 & \boldmath\textbf{$45.6 \pm 4.8$\%}\unboldmath & $26.7 \pm 0.7$\% & \underline{$27.2 \pm 3.3$\%} & \boldmath\textbf{$28.5 \pm 2.9$\%}\unboldmath & $25.6 \pm 1.6$\% & \underline{$27.7 \pm 2.2$\%} \\
TwoPatterns & 0.02 & \boldmath\textbf{$46.3 \pm 2.9$\%}\unboldmath & $27.0 \pm 1.2$\% & \underline{$27.1 \pm 2.8$\%} & \boldmath\textbf{$29.0 \pm 4.1$\%}\unboldmath & $25.6 \pm 1.5$\% & \underline{$28.4 \pm 2.3$\%} \\
TwoPatterns & 0.04 & \boldmath\textbf{$42.0 \pm 7.9$\%}\unboldmath & \underline{$26.6 \pm 1.3$\%} & \underline{$26.6 \pm 2.7$\%} & \boldmath\textbf{$29.1 \pm 4.8$\%}\unboldmath & $25.3 \pm 1.2$\% & \underline{$26.5 \pm 3.1$\%} \\
TwoPatterns & 0.08 & \boldmath\textbf{$41.8 \pm 7.3$\%}\unboldmath & $28.0 \pm 1.4$\% & \underline{$33.2 \pm 6.8$\%} & \underline{$29.0 \pm 3.5$\%} & $26.0 \pm 1.8$\% & \boldmath\textbf{$32.6 \pm 3.8$\%}\unboldmath \\
TwoPatterns & 0.16 & \boldmath\textbf{$50.3 \pm 0.4$\%}\unboldmath & $35.3 \pm 6.2$\% & \underline{$39.8 \pm 3.6$\%} & \boldmath\textbf{$34.8 \pm 2.4$\%}\unboldmath & $28.6 \pm 4.0$\% & \underline{$32.8 \pm 3.3$\%} \\
TwoPatterns & 0.32 & \boldmath\textbf{$50.1 \pm 0.4$\%}\unboldmath & $45.0 \pm 5.4$\% & \underline{$49.6 \pm 15.9$\%} & \boldmath\textbf{$41.2 \pm 4.0$\%}\unboldmath & $32.5 \pm 3.1$\% & \underline{$35.1 \pm 4.5$\%} \\
TwoPatterns & 0.64 & $50.4 \pm 0.5$\% & \boldmath\textbf{$74.9 \pm 1.6$\%}\unboldmath & \underline{$58.0 \pm 16.4$\%} & \boldmath\textbf{$45.3 \pm 3.1$\%}\unboldmath & $40.5 \pm 2.0$\% & \underline{$40.6 \pm 6.2$\%} \\
TwoPatterns & 1.00 & \underline{$50.8 \pm 0.0$\%} & \boldmath\textbf{$82.6 \pm 3.6$\%}\unboldmath & $50.1 \pm 12.0$\% & \boldmath\textbf{$47.9 \pm 4.2$\%}\unboldmath & \underline{$46.8 \pm 3.5$\%} & $37.5 \pm 8.1$\% \\
\midrule
Wafer & 0.01 & \underline{$79.7 \pm 14.7$\%} & $78.6 \pm 9.9$\% & \boldmath\textbf{$85.2 \pm 8.3$\%}\unboldmath & $50.3 \pm 18.1$\% & \underline{$81.5 \pm 11.7$\%} & \boldmath\textbf{$83.8 \pm 7.1$\%}\unboldmath \\
Wafer & 0.02 & \underline{$79.7 \pm 14.7$\%} & $76.7 \pm 10.9$\% & \boldmath\textbf{$83.8 \pm 10.2$\%}\unboldmath & $51.1 \pm 19.1$\% & \underline{$80.3 \pm 13.0$\%} & \boldmath\textbf{$88.3 \pm 1.9$\%}\unboldmath \\
Wafer & 0.04 & \underline{$78.9 \pm 14.2$\%} & $75.9 \pm 11.8$\% & \boldmath\textbf{$86.5 \pm 2.1$\%}\unboldmath & $56.7 \pm 25.6$\% & \underline{$80.5 \pm 13.0$\%} & \boldmath\textbf{$83.1 \pm 8.9$\%}\unboldmath \\
Wafer & 0.08 & \underline{$78.8 \pm 14.1$\%} & $78.3 \pm 9.7$\% & \boldmath\textbf{$89.7 \pm 4.0$\%}\unboldmath & $56.8 \pm 26.5$\% & \underline{$81.5 \pm 12.6$\%} & \boldmath\textbf{$85.2 \pm 7.3$\%}\unboldmath \\
Wafer & 0.16 & $80.6 \pm 15.0$\% & \underline{$89.4 \pm 1.0$\%} & \boldmath\textbf{$91.7 \pm 2.8$\%}\unboldmath & $72.2 \pm 18.3$\% & \underline{$82.8 \pm 6.2$\%} & \boldmath\textbf{$89.6 \pm 0.6$\%}\unboldmath \\
Wafer & 0.32 & $89.2 \pm 0.0$\% & \underline{$91.4 \pm 1.5$\%} & \boldmath\textbf{$91.9 \pm 2.7$\%}\unboldmath & $75.7 \pm 17.3$\% & \underline{$83.7 \pm 4.8$\%} & \boldmath\textbf{$85.8 \pm 6.6$\%}\unboldmath \\
Wafer & 0.64 & $89.2 \pm 0.1$\% & \underline{$95.4 \pm 0.5$\%} & \boldmath\textbf{$96.0 \pm 1.4$\%}\unboldmath & $84.4 \pm 5.7$\% & \underline{$88.2 \pm 1.4$\%} & \boldmath\textbf{$89.9 \pm 2.6$\%}\unboldmath \\
Wafer & 1.00 & $89.2 \pm 0.0$\% & \underline{$97.8 \pm 1.0$\%} & \boldmath\textbf{$98.6 \pm 0.4$\%}\unboldmath & $87.3 \pm 3.0$\% & \underline{$90.3 \pm 2.2$\%} & \boldmath\textbf{$93.5 \pm 0.8$\%}\unboldmath \\
\bottomrule
\end{longtable}
\end{scriptsize}

}
\subsection{Complete Results for Deep SSMs}
In this section, we provide full per-dataset and per-ratio results in Table~\ref{tab:appendix-complete-deep-ssm}, including the TDI-Frozen variant where the TDI-initialized first S4 layer is kept fixed during supervised training.
\label{app:deep-ssm-full-results}
{
\captionsetup{font=normalsize}
\scriptsize
\setlength{\tabcolsep}{3.5pt}
\renewcommand{\arraystretch}{1.05}
\begin{scriptsize}
\begin{longtable}{llcccccc}
\caption{\textbf{Complete final performance for the SSM baseline, TDI, and TDI-Frozen.} The left block reports final test accuracy; the right block reports mean test accuracy over the first 5 epochs. Each cell reports mean $\pm$ standard deviation over 5 seeds. The best value within each block for each dataset-ratio row is bolded, and the second-best value is underlined.}\label{tab:appendix-complete-deep-ssm}\\
\toprule
& & \multicolumn{3}{c}{Final test accuracy} & \multicolumn{3}{c}{First-5 test accuracy} \\
\cmidrule(lr){3-5} \cmidrule(lr){6-8}
Dataset & Ratio & Baseline & TDI & TDI-Frozen & Baseline & TDI & TDI-Frozen \\
\midrule
\endfirsthead
\toprule
& & \multicolumn{3}{c}{Final test accuracy} & \multicolumn{3}{c}{First-5 test accuracy} \\
\cmidrule(lr){3-5} \cmidrule(lr){6-8}
Dataset & Ratio & Baseline & TDI & TDI-Frozen & Baseline & TDI & TDI-Frozen \\
\midrule
\endhead
CIFAR-10 & 0.005 & \boldmath\textbf{$15.0 \pm 1.3$\%}\unboldmath & \boldmath\textbf{$15.0 \pm 1.0$\%}\unboldmath & \boldmath\textbf{$15.0 \pm 0.9$\%}\unboldmath & \underline{$10.7 \pm 0.3$\%} & \boldmath\textbf{$11.3 \pm 0.4$\%}\unboldmath & \boldmath\textbf{$11.3 \pm 0.4$\%}\unboldmath \\
CIFAR-10 & 0.010 & $19.9 \pm 1.3$\% & \boldmath\textbf{$23.8 \pm 2.4$\%}\unboldmath & \underline{$23.6 \pm 2.3$\%} & \underline{$10.8 \pm 1.0$\%} & \boldmath\textbf{$11.0 \pm 0.7$\%}\unboldmath & \boldmath\textbf{$11.0 \pm 0.7$\%}\unboldmath \\
CIFAR-10 & 0.022 & $32.8 \pm 3.5$\% & \boldmath\textbf{$38.3 \pm 1.4$\%}\unboldmath & \underline{$37.7 \pm 1.7$\%} & \underline{$12.4 \pm 0.2$\%} & \boldmath\textbf{$12.8 \pm 1.3$\%}\unboldmath & \boldmath\textbf{$12.8 \pm 1.4$\%}\unboldmath \\
CIFAR-10 & 0.046 & \underline{$46.2 \pm 1.2$\%} & \boldmath\textbf{$46.4 \pm 1.1$\%}\unboldmath & $46.1 \pm 0.9$\% & $14.6 \pm 0.9$\% & \boldmath\textbf{$14.9 \pm 1.1$\%}\unboldmath & \underline{$14.8 \pm 1.1$\%} \\
CIFAR-10 & 0.100 & \boldmath\textbf{$55.4 \pm 1.1$\%}\unboldmath & \underline{$53.3 \pm 0.6$\%} & $53.2 \pm 0.7$\% & $18.9 \pm 1.9$\% & \boldmath\textbf{$22.0 \pm 3.6$\%}\unboldmath & \underline{$21.9 \pm 3.4$\%} \\
CIFAR-10 & 0.215 & \boldmath\textbf{$64.7 \pm 0.8$\%}\unboldmath & \underline{$59.5 \pm 0.9$\%} & $59.2 \pm 0.6$\% & $31.0 \pm 1.8$\% & \underline{$35.3 \pm 3.2$\%} & \boldmath\textbf{$35.6 \pm 3.1$\%}\unboldmath \\
CIFAR-10 & 0.464 & \boldmath\textbf{$74.7 \pm 0.9$\%}\unboldmath & $69.3 \pm 0.7$\% & \underline{$69.7 \pm 0.9$\%} & $46.8 \pm 2.0$\% & \boldmath\textbf{$47.7 \pm 1.7$\%}\unboldmath & \underline{$47.4 \pm 1.7$\%} \\
CIFAR-10 & 1.000 & \boldmath\textbf{$81.6 \pm 0.9$\%}\unboldmath & \underline{$79.5 \pm 0.6$\%} & $79.4 \pm 0.5$\% & \boldmath\textbf{$61.1 \pm 1.1$\%}\unboldmath & $58.0 \pm 1.3$\% & \underline{$58.2 \pm 1.3$\%} \\
\midrule
Freq. Cls. & 0.010 & $57.9 \pm 13.5$\% & \boldmath\textbf{$73.5 \pm 9.1$\%}\unboldmath & \underline{$73.4 \pm 9.3$\%} & $18.6 \pm 9.5$\% & \boldmath\textbf{$50.8 \pm 10.7$\%}\unboldmath & \underline{$50.5 \pm 10.9$\%} \\
Freq. Cls. & 0.022 & \boldmath\textbf{$97.3 \pm 0.8$\%}\unboldmath & $93.8 \pm 2.7$\% & \underline{$94.4 \pm 2.1$\%} & $54.4 \pm 12.4$\% & \underline{$74.2 \pm 9.1$\%} & \boldmath\textbf{$74.3 \pm 9.0$\%}\unboldmath \\
Freq. Cls. & 0.046 & \boldmath\textbf{$98.6 \pm 0.5$\%}\unboldmath & \underline{$98.5 \pm 0.4$\%} & $98.4 \pm 0.6$\% & $83.7 \pm 3.5$\% & \boldmath\textbf{$91.1 \pm 1.9$\%}\unboldmath & \underline{$90.4 \pm 2.0$\%} \\
Freq. Cls. & 0.100 & \underline{$98.6 \pm 0.4$\%} & $98.5 \pm 0.6$\% & \boldmath\textbf{$98.7 \pm 0.7$\%}\unboldmath & \boldmath\textbf{$96.8 \pm 0.7$\%}\unboldmath & \underline{$96.2 \pm 0.9$\%} & $95.9 \pm 0.9$\% \\
Freq. Cls. & 0.215 & \boldmath\textbf{$98.9 \pm 0.4$\%}\unboldmath & $98.7 \pm 0.6$\% & \underline{$98.8 \pm 0.5$\%} & \boldmath\textbf{$98.4 \pm 0.4$\%}\unboldmath & \underline{$97.8 \pm 0.2$\%} & \underline{$97.8 \pm 0.4$\%} \\
Freq. Cls. & 0.464 & \boldmath\textbf{$99.3 \pm 0.4$\%}\unboldmath & $98.8 \pm 0.4$\% & \underline{$99.2 \pm 0.2$\%} & \boldmath\textbf{$98.8 \pm 0.3$\%}\unboldmath & \underline{$98.6 \pm 0.2$\%} & $98.5 \pm 0.2$\% \\
Freq. Cls. & 1.000 & \underline{$99.3 \pm 0.3$\%} & \underline{$99.3 \pm 0.2$\%} & \boldmath\textbf{$99.4 \pm 0.3$\%}\unboldmath & \boldmath\textbf{$99.0 \pm 0.2$\%}\unboldmath & $98.8 \pm 0.2$\% & \underline{$98.9 \pm 0.2$\%} \\
\midrule
ListOps & 0.001 & \boldmath\textbf{$12.6 \pm 2.4$\%}\unboldmath & \underline{$12.2 \pm 3.4$\%} & $11.4 \pm 2.8$\% & $11.0 \pm 1.4$\% & \underline{$11.3 \pm 0.8$\%} & \boldmath\textbf{$11.4 \pm 0.9$\%}\unboldmath \\
ListOps & 0.002 & \boldmath\textbf{$13.3 \pm 1.2$\%}\unboldmath & $12.6 \pm 0.7$\% & \underline{$12.8 \pm 0.7$\%} & $11.0 \pm 2.5$\% & \boldmath\textbf{$12.9 \pm 2.2$\%}\unboldmath & \underline{$12.5 \pm 2.2$\%} \\
ListOps & 0.005 & \boldmath\textbf{$13.1 \pm 0.5$\%}\unboldmath & $12.5 \pm 0.6$\% & \underline{$12.7 \pm 1.2$\%} & $12.0 \pm 1.5$\% & \underline{$12.1 \pm 1.2$\%} & \boldmath\textbf{$12.6 \pm 0.7$\%}\unboldmath \\
ListOps & 0.010 & \underline{$12.4 \pm 0.5$\%} & \boldmath\textbf{$12.5 \pm 0.7$\%}\unboldmath & $11.9 \pm 0.4$\% & \underline{$11.8 \pm 1.1$\%} & \boldmath\textbf{$12.2 \pm 1.1$\%}\unboldmath & \boldmath\textbf{$12.2 \pm 1.1$\%}\unboldmath \\
ListOps & 0.022 & \boldmath\textbf{$13.0 \pm 0.9$\%}\unboldmath & \underline{$12.1 \pm 0.8$\%} & $11.9 \pm 1.2$\% & $12.5 \pm 0.7$\% & \boldmath\textbf{$13.4 \pm 0.5$\%}\unboldmath & \underline{$13.0 \pm 1.4$\%} \\
ListOps & 0.046 & \underline{$20.7 \pm 7.0$\%} & \boldmath\textbf{$22.5 \pm 4.2$\%}\unboldmath & $16.1 \pm 6.9$\% & \boldmath\textbf{$16.0 \pm 2.3$\%}\unboldmath & \underline{$15.2 \pm 0.5$\%} & $14.7 \pm 0.7$\% \\
ListOps & 0.100 & \boldmath\textbf{$31.0 \pm 1.1$\%}\unboldmath & $28.8 \pm 0.9$\% & \underline{$29.3 \pm 2.7$\%} & \boldmath\textbf{$24.8 \pm 5.9$\%}\unboldmath & $18.2 \pm 2.1$\% & \underline{$20.3 \pm 5.6$\%} \\
ListOps & 0.215 & \boldmath\textbf{$36.1 \pm 0.9$\%}\unboldmath & \underline{$36.0 \pm 1.1$\%} & $33.9 \pm 1.5$\% & \boldmath\textbf{$32.0 \pm 6.5$\%}\unboldmath & $21.7 \pm 6.6$\% & \underline{$22.8 \pm 5.3$\%} \\
ListOps & 0.464 & \boldmath\textbf{$48.5 \pm 1.7$\%}\unboldmath & \underline{$46.6 \pm 1.9$\%} & $43.7 \pm 2.5$\% & \boldmath\textbf{$36.9 \pm 3.8$\%}\unboldmath & \underline{$29.2 \pm 10.0$\%} & $28.8 \pm 9.0$\% \\
ListOps & 1.000 & \boldmath\textbf{$49.4 \pm 4.8$\%}\unboldmath & $47.0 \pm 5.4$\% & \underline{$47.9 \pm 4.1$\%} & \boldmath\textbf{$42.0 \pm 0.6$\%}\unboldmath & \underline{$32.9 \pm 9.4$\%} & $29.8 \pm 9.5$\% \\
\midrule
sMNIST & 0.005 & \underline{$10.0 \pm 0.3$\%} & \boldmath\textbf{$10.3 \pm 0.9$\%}\unboldmath & \boldmath\textbf{$10.3 \pm 0.9$\%}\unboldmath & \boldmath\textbf{$10.2 \pm 0.1$\%}\unboldmath & \underline{$10.0 \pm 0.1$\%} & \underline{$10.0 \pm 0.1$\%} \\
sMNIST & 0.010 & $10.7 \pm 0.8$\% & \boldmath\textbf{$38.6 \pm 22.0$\%}\unboldmath & \underline{$33.7 \pm 21.7$\%} & \boldmath\textbf{$10.2 \pm 0.2$\%}\unboldmath & \underline{$10.1 \pm 0.4$\%} & \underline{$10.1 \pm 0.3$\%} \\
sMNIST & 0.022 & $64.9 \pm 30.8$\% & \boldmath\textbf{$91.9 \pm 4.8$\%}\unboldmath & \underline{$80.8 \pm 26.6$\%} & $10.2 \pm 0.1$\% & \underline{$10.7 \pm 0.3$\%} & \boldmath\textbf{$11.0 \pm 0.6$\%}\unboldmath \\
sMNIST & 0.046 & \underline{$97.6 \pm 0.2$\%} & \boldmath\textbf{$97.7 \pm 0.3$\%}\unboldmath & $97.5 \pm 0.4$\% & $10.3 \pm 0.4$\% & \boldmath\textbf{$11.8 \pm 2.9$\%}\unboldmath & \underline{$11.3 \pm 1.9$\%} \\
sMNIST & 0.100 & \boldmath\textbf{$98.5 \pm 0.1$\%}\unboldmath & \underline{$98.4 \pm 0.2$\%} & \underline{$98.4 \pm 0.0$\%} & $12.0 \pm 3.7$\% & \boldmath\textbf{$41.9 \pm 17.7$\%}\unboldmath & \underline{$39.2 \pm 15.7$\%} \\
sMNIST & 0.215 & \underline{$98.8 \pm 0.2$\%} & \underline{$98.8 \pm 0.2$\%} & \boldmath\textbf{$99.0 \pm 0.1$\%}\unboldmath & $59.6 \pm 9.1$\% & \boldmath\textbf{$66.9 \pm 7.7$\%}\unboldmath & \underline{$65.6 \pm 8.1$\%} \\
sMNIST & 0.464 & \boldmath\textbf{$99.1 \pm 0.1$\%}\unboldmath & \boldmath\textbf{$99.1 \pm 0.2$\%}\unboldmath & \boldmath\textbf{$99.1 \pm 0.1$\%}\unboldmath & $81.5 \pm 5.9$\% & \underline{$91.3 \pm 5.8$\%} & \boldmath\textbf{$91.8 \pm 4.3$\%}\unboldmath \\
sMNIST & 1.000 & \boldmath\textbf{$99.2 \pm 0.1$\%}\unboldmath & \underline{$99.1 \pm 0.2$\%} & \underline{$99.1 \pm 0.1$\%} & \boldmath\textbf{$98.5 \pm 0.2$\%}\unboldmath & \boldmath\textbf{$98.5 \pm 0.1$\%}\unboldmath & \boldmath\textbf{$98.5 \pm 0.2$\%}\unboldmath \\
\midrule
Pathfinder & 0.001 & \underline{$50.0 \pm 0.4$\%} & \boldmath\textbf{$50.1 \pm 0.2$\%}\unboldmath & $49.9 \pm 0.1$\% & \boldmath\textbf{$50.0 \pm 0.2$\%}\unboldmath & \boldmath\textbf{$50.0 \pm 0.2$\%}\unboldmath & \underline{$49.9 \pm 0.2$\%} \\
Pathfinder & 0.002 & \boldmath\textbf{$50.4 \pm 0.4$\%}\unboldmath & \underline{$50.1 \pm 0.1$\%} & \underline{$50.1 \pm 0.4$\%} & \boldmath\textbf{$50.1 \pm 0.3$\%}\unboldmath & \underline{$49.9 \pm 0.2$\%} & \underline{$49.9 \pm 0.2$\%} \\
Pathfinder & 0.005 & \underline{$50.2 \pm 0.4$\%} & \boldmath\textbf{$50.4 \pm 0.5$\%}\unboldmath & $50.1 \pm 0.2$\% & \boldmath\textbf{$50.1 \pm 0.1$\%}\unboldmath & \underline{$50.0 \pm 0.1$\%} & $49.9 \pm 0.2$\% \\
Pathfinder & 0.010 & \underline{$50.2 \pm 0.4$\%} & \boldmath\textbf{$50.3 \pm 0.4$\%}\unboldmath & \underline{$50.2 \pm 0.6$\%} & \boldmath\textbf{$50.0 \pm 0.1$\%}\unboldmath & \underline{$49.9 \pm 0.2$\%} & \underline{$49.9 \pm 0.1$\%} \\
Pathfinder & 0.022 & \boldmath\textbf{$50.7 \pm 0.6$\%}\unboldmath & \boldmath\textbf{$50.7 \pm 0.4$\%}\unboldmath & \underline{$50.4 \pm 0.3$\%} & \boldmath\textbf{$50.0 \pm 0.2$\%}\unboldmath & \boldmath\textbf{$50.0 \pm 0.1$\%}\unboldmath & \boldmath\textbf{$50.0 \pm 0.1$\%}\unboldmath \\
Pathfinder & 0.046 & \underline{$51.6 \pm 0.5$\%} & \boldmath\textbf{$52.5 \pm 1.1$\%}\unboldmath & $51.5 \pm 0.9$\% & \boldmath\textbf{$50.1 \pm 0.1$\%}\unboldmath & \underline{$50.0 \pm 0.2$\%} & \underline{$50.0 \pm 0.1$\%} \\
Pathfinder & 0.100 & \underline{$56.8 \pm 1.2$\%} & \boldmath\textbf{$57.4 \pm 1.4$\%}\unboldmath & $54.8 \pm 0.8$\% & \boldmath\textbf{$50.2 \pm 0.2$\%}\unboldmath & \underline{$50.1 \pm 0.1$\%} & \underline{$50.1 \pm 0.2$\%} \\
Pathfinder & 0.215 & \underline{$62.8 \pm 2.5$\%} & \boldmath\textbf{$64.5 \pm 0.8$\%}\unboldmath & $59.6 \pm 2.6$\% & $50.0 \pm 0.1$\% & \underline{$50.1 \pm 0.2$\%} & \boldmath\textbf{$50.4 \pm 0.3$\%}\unboldmath \\
Pathfinder & 0.464 & \underline{$69.4 \pm 8.8$\%} & \boldmath\textbf{$70.9 \pm 1.4$\%}\unboldmath & $66.3 \pm 1.1$\% & \underline{$50.1 \pm 0.3$\%} & \underline{$50.1 \pm 0.2$\%} & \boldmath\textbf{$50.2 \pm 0.1$\%}\unboldmath \\
Pathfinder & 1.000 & $67.8 \pm 16.4$\% & \underline{$69.7 \pm 14.1$\%} & \boldmath\textbf{$80.8 \pm 2.4$\%}\unboldmath & \underline{$51.9 \pm 3.1$\%} & $50.0 \pm 0.1$\% & \boldmath\textbf{$52.8 \pm 4.1$\%}\unboldmath \\
\midrule
pMNIST & 0.005 & \underline{$10.0 \pm 0.3$\%} & \boldmath\textbf{$10.3 \pm 0.6$\%}\unboldmath & \boldmath\textbf{$10.3 \pm 0.6$\%}\unboldmath & \boldmath\textbf{$10.2 \pm 0.1$\%}\unboldmath & \underline{$10.1 \pm 0.1$\%} & \underline{$10.1 \pm 0.1$\%} \\
pMNIST & 0.010 & $10.7 \pm 0.8$\% & \boldmath\textbf{$11.6 \pm 1.2$\%}\unboldmath & \underline{$11.3 \pm 0.8$\%} & \boldmath\textbf{$10.3 \pm 0.2$\%}\unboldmath & \underline{$10.2 \pm 0.3$\%} & \underline{$10.2 \pm 0.3$\%} \\
pMNIST & 0.022 & \boldmath\textbf{$54.9 \pm 20.6$\%}\unboldmath & \underline{$54.2 \pm 26.0$\%} & $51.1 \pm 26.5$\% & \underline{$10.2 \pm 0.2$\%} & \boldmath\textbf{$10.3 \pm 0.3$\%}\unboldmath & \boldmath\textbf{$10.3 \pm 0.3$\%}\unboldmath \\
pMNIST & 0.046 & \underline{$91.6 \pm 0.7$\%} & \boldmath\textbf{$92.1 \pm 0.6$\%}\unboldmath & $91.5 \pm 1.0$\% & \boldmath\textbf{$10.8 \pm 0.8$\%}\unboldmath & \underline{$10.1 \pm 0.3$\%} & \underline{$10.1 \pm 0.3$\%} \\
pMNIST & 0.100 & $94.0 \pm 1.2$\% & \underline{$94.7 \pm 0.5$\%} & \boldmath\textbf{$94.8 \pm 0.5$\%}\unboldmath & $12.9 \pm 2.9$\% & \underline{$20.8 \pm 9.6$\%} & \boldmath\textbf{$21.4 \pm 10.0$\%}\unboldmath \\
pMNIST & 0.215 & \underline{$96.1 \pm 0.4$\%} & $95.4 \pm 0.7$\% & \boldmath\textbf{$96.3 \pm 0.4$\%}\unboldmath & $42.8 \pm 16.8$\% & \boldmath\textbf{$50.9 \pm 7.0$\%}\unboldmath & \underline{$49.0 \pm 7.2$\%} \\
pMNIST & 0.464 & \boldmath\textbf{$97.0 \pm 0.3$\%}\unboldmath & \boldmath\textbf{$97.0 \pm 0.2$\%}\unboldmath & \underline{$96.9 \pm 0.1$\%} & \boldmath\textbf{$79.9 \pm 1.9$\%}\unboldmath & \underline{$78.7 \pm 1.5$\%} & $73.6 \pm 9.5$\% \\
pMNIST & 1.000 & \boldmath\textbf{$97.7 \pm 0.2$\%}\unboldmath & \underline{$97.6 \pm 0.1$\%} & \underline{$97.6 \pm 0.2$\%} & \underline{$94.2 \pm 0.9$\%} & $93.6 \pm 1.8$\% & \boldmath\textbf{$94.5 \pm 1.3$\%}\unboldmath \\
\midrule
SC & 0.001 & \underline{$7.6 \pm 2.5$\%} & $7.5 \pm 3.8$\% & \boldmath\textbf{$10.8 \pm 3.0$\%}\unboldmath & $2.8 \pm 1.0$\% & \underline{$3.5 \pm 1.0$\%} & \boldmath\textbf{$3.6 \pm 0.7$\%}\unboldmath \\
SC & 0.002 & $12.2 \pm 4.1$\% & \boldmath\textbf{$16.5 \pm 7.5$\%}\unboldmath & \underline{$15.6 \pm 5.7$\%} & $3.0 \pm 0.6$\% & \boldmath\textbf{$4.4 \pm 0.6$\%}\unboldmath & \underline{$4.3 \pm 0.6$\%} \\
SC & 0.005 & $35.2 \pm 6.7$\% & \boldmath\textbf{$44.7 \pm 4.4$\%}\unboldmath & \underline{$40.8 \pm 12.2$\%} & \underline{$3.6 \pm 0.5$\%} & \boldmath\textbf{$6.8 \pm 0.8$\%}\unboldmath & \boldmath\textbf{$6.8 \pm 1.4$\%}\unboldmath \\
SC & 0.010 & $62.3 \pm 3.1$\% & \boldmath\textbf{$65.5 \pm 5.7$\%}\unboldmath & \underline{$62.8 \pm 4.1$\%} & $11.4 \pm 4.6$\% & \boldmath\textbf{$16.9 \pm 5.3$\%}\unboldmath & \underline{$16.7 \pm 5.3$\%} \\
SC & 0.022 & \underline{$75.7 \pm 0.7$\%} & \boldmath\textbf{$77.1 \pm 2.1$\%}\unboldmath & $74.6 \pm 2.3$\% & $25.0 \pm 3.1$\% & \boldmath\textbf{$41.5 \pm 1.8$\%}\unboldmath & \underline{$39.2 \pm 3.3$\%} \\
SC & 0.046 & \boldmath\textbf{$82.9 \pm 1.0$\%}\unboldmath & \boldmath\textbf{$82.9 \pm 1.2$\%}\unboldmath & \underline{$82.1 \pm 0.5$\%} & $50.0 \pm 6.8$\% & \boldmath\textbf{$62.1 \pm 3.6$\%}\unboldmath & \underline{$61.1 \pm 3.7$\%} \\
SC & 0.100 & \boldmath\textbf{$86.9 \pm 0.8$\%}\unboldmath & $86.0 \pm 1.9$\% & \underline{$86.6 \pm 1.0$\%} & $70.1 \pm 3.7$\% & \underline{$74.0 \pm 2.2$\%} & \boldmath\textbf{$75.0 \pm 1.6$\%}\unboldmath \\
SC & 0.215 & $86.4 \pm 1.8$\% & \boldmath\textbf{$88.1 \pm 1.4$\%}\unboldmath & \underline{$87.6 \pm 1.3$\%} & $78.4 \pm 4.5$\% & \boldmath\textbf{$83.1 \pm 0.6$\%}\unboldmath & \underline{$82.6 \pm 1.2$\%} \\
SC & 0.464 & $87.4 \pm 1.8$\% & \boldmath\textbf{$88.1 \pm 1.4$\%}\unboldmath & \underline{$87.8 \pm 0.7$\%} & $83.9 \pm 0.5$\% & \boldmath\textbf{$84.9 \pm 0.8$\%}\unboldmath & \underline{$84.4 \pm 0.7$\%} \\
SC & 1.000 & \underline{$88.0 \pm 0.8$\%} & $87.5 \pm 3.6$\% & \boldmath\textbf{$88.7 \pm 0.6$\%}\unboldmath & $85.5 \pm 1.0$\% & \underline{$85.6 \pm 0.6$\%} & \boldmath\textbf{$86.2 \pm 0.8$\%}\unboldmath \\
\midrule
SC10 & 0.001 & \boldmath\textbf{$10.9 \pm 1.1$\%}\unboldmath & $10.6 \pm 0.7$\% & \underline{$10.7 \pm 1.0$\%} & \underline{$10.1 \pm 0.2$\%} & \boldmath\textbf{$10.2 \pm 0.3$\%}\unboldmath & \boldmath\textbf{$10.2 \pm 0.2$\%}\unboldmath \\
SC10 & 0.002 & $17.5 \pm 5.3$\% & \underline{$17.8 \pm 6.3$\%} & \boldmath\textbf{$18.4 \pm 4.7$\%}\unboldmath & \underline{$10.3 \pm 0.4$\%} & \boldmath\textbf{$10.5 \pm 0.7$\%}\unboldmath & \boldmath\textbf{$10.5 \pm 0.7$\%}\unboldmath \\
SC10 & 0.005 & $23.7 \pm 8.3$\% & \boldmath\textbf{$27.9 \pm 9.6$\%}\unboldmath & \underline{$25.3 \pm 6.9$\%} & \underline{$10.3 \pm 0.3$\%} & \boldmath\textbf{$12.9 \pm 1.2$\%}\unboldmath & \boldmath\textbf{$12.9 \pm 1.9$\%}\unboldmath \\
SC10 & 0.010 & $43.0 \pm 10.9$\% & \boldmath\textbf{$59.5 \pm 3.1$\%}\unboldmath & \underline{$57.5 \pm 10.0$\%} & $11.2 \pm 0.9$\% & \underline{$14.5 \pm 1.7$\%} & \boldmath\textbf{$14.8 \pm 1.5$\%}\unboldmath \\
SC10 & 0.022 & $76.7 \pm 2.7$\% & \boldmath\textbf{$79.0 \pm 2.7$\%}\unboldmath & \underline{$77.8 \pm 5.4$\%} & $14.6 \pm 2.0$\% & \boldmath\textbf{$24.3 \pm 5.3$\%}\unboldmath & \underline{$22.5 \pm 4.0$\%} \\
SC10 & 0.046 & $84.5 \pm 4.4$\% & \boldmath\textbf{$85.8 \pm 1.1$\%}\unboldmath & \underline{$84.6 \pm 2.3$\%} & $36.2 \pm 8.1$\% & \boldmath\textbf{$46.5 \pm 6.7$\%}\unboldmath & \underline{$40.9 \pm 3.4$\%} \\
SC10 & 0.100 & \underline{$87.8 \pm 2.8$\%} & \boldmath\textbf{$89.4 \pm 1.0$\%}\unboldmath & \underline{$87.8 \pm 2.4$\%} & $59.5 \pm 8.6$\% & \underline{$66.3 \pm 5.9$\%} & \boldmath\textbf{$67.1 \pm 3.8$\%}\unboldmath \\
SC10 & 0.215 & \underline{$91.4 \pm 0.8$\%} & \boldmath\textbf{$92.1 \pm 0.5$\%}\unboldmath & $91.1 \pm 1.0$\% & $77.2 \pm 6.3$\% & \boldmath\textbf{$82.8 \pm 1.7$\%}\unboldmath & \underline{$82.2 \pm 2.5$\%} \\
SC10 & 0.464 & \underline{$92.5 \pm 0.5$\%} & \boldmath\textbf{$92.7 \pm 0.8$\%}\unboldmath & $91.6 \pm 1.7$\% & $84.4 \pm 2.5$\% & \underline{$88.1 \pm 1.1$\%} & \boldmath\textbf{$88.3 \pm 1.3$\%}\unboldmath \\
SC10 & 1.000 & \boldmath\textbf{$93.4 \pm 0.9$\%}\unboldmath & \underline{$93.3 \pm 0.3$\%} & $93.2 \pm 1.1$\% & $89.7 \pm 1.3$\% & \underline{$90.5 \pm 0.3$\%} & \boldmath\textbf{$90.6 \pm 0.7$\%}\unboldmath \\
\bottomrule
\end{longtable}
\end{scriptsize}
}
\subsection{Early Learning versus Extended Training}
\label{app:extended-training}
The main experiments report performance under fixed training budgets. To examine whether the observed gains persist with longer training, we extend the training budget from 30 to 100 epochs on sMNIST and from 10 to 50 epochs on Frequency Classification. We compare the baseline and TDI using five matched seeds at the training ratios listed in Table~\ref{tab:extended-training}.

\paragraph{Evaluation metrics.}
We report three complementary measures. \emph{Early-10} is the mean test accuracy over the first ten
epochs, computed separately for each run and then averaged over five seeds. \emph{Time-to-80\%} is the first completed epoch at which a run reaches 80\% test accuracy, reported as mean $\pm$ standard deviation across seeds. All runs reach this threshold. \emph{Final $\Delta$Acc.} is the TDI-minus-baseline difference in test accuracy at the final epoch of the extended training budget, computed for each matched seed pair and reported as mean $\pm$ standard deviation in percentage points.

\begin{table*}[t]
\centering
\setlength{\tabcolsep}{4pt}
\caption{
\textbf{Early learning and accuracy after extended training.}
Time-to-80\% reports the number of epochs required to reach
80\% test accuracy (mean $\pm$ SD over five seeds).
Early-10 reports mean test accuracy over the first ten epochs,
averaged over seeds.
Final $\Delta$Acc.\ reports the paired TDI-minus-baseline
final-epoch accuracy difference (mean $\pm$ SD), in percentage
points (pp).
Bold indicates the lower mean Time-to-80\% and the higher
Early-10 accuracy within each row; ties are both bolded.
}
\label{tab:extended-training}
\small
\begin{tabular}{lccrrrrr}
\toprule
& & & \multicolumn{2}{c}{Time-to-80\% (epochs)}
& \multicolumn{2}{c}{Early-10 (\%)}
& Final $\Delta$Acc. \\
\cmidrule(lr){4-5}
\cmidrule(lr){6-7}
Task & Train ratio & Epochs
& Baseline & TDI
& Baseline & TDI
& (pp) \\
\midrule
sMNIST
& 2.15\% & 100
& 34.2 \textpm{} 4.5 & \textbf{20.0 \textpm{} 4.6}
& 10.19 & \textbf{14.15}
& +0.12 \textpm{} 0.47 \\
sMNIST
& 21.5\% & 100
& 4.6 \textpm{} 0.9 & \textbf{3.0 \textpm{} 0.0}
& 69.71 & \textbf{83.69}
& \textminus{}0.08 \textpm{} 0.07 \\
\midrule
Freq. Cls.
& 2.15\% & 50
& 11.6 \textpm{} 1.7 & \textbf{9.6 \textpm{} 4.0}
& 32.72 & \textbf{63.25}
& \textminus{}0.69 \textpm{} 0.73 \\
Freq. Cls.
& 10.0\% & 50
& 2.0 \textpm{} 0.0 & \textbf{1.2 \textpm{} 0.4}
& 92.04 & \textbf{96.03}
& +0.29 \textpm{} 0.29 \\
Freq. Cls.
& 21.5\% & 50
& \textbf{1.0 \textpm{} 0.0} & \textbf{1.0 \textpm{} 0.0}
& \textbf{98.04} & 97.93
& +0.13 \textpm{} 0.15 \\
\bottomrule
\end{tabular}
\end{table*}

\paragraph{Results and interpretation.}
On sMNIST, TDI reaches 80\% test accuracy earlier and achieves higher Early-10 accuracy at both evaluated training ratios, while the baseline largely catches up by 100 epochs. For example, at 2.15\% training data, the mean time-to-80\% decreases from 34.2 to 20.0 epochs, whereas the final accuracy difference is only $+0.12 \pm 0.47$ percentage points. Frequency Classification shows a similar early-learning advantage at 2.15\% and 10.0\% training data, but little early-learning difference at 21.5\%. After 50 epochs, final accuracy differences are small and can favor either initialization.

These experiments indicate that, in the evaluated settings, TDI primarily improves the learning trajectory without systematically increasing accuracy after extended training. They qualify the interpretation of the main experiments: a higher final-epoch accuracy under a fixed training budget does not by itself establish a persistent advantage with longer training. These finite-horizon results do not establish asymptotic performance or imply that the same behavior holds on every benchmark.

\subsection{First-Layer versus All-Layer TDI}
\label{app:tdi-layer-scope}

We examine whether applying TDI beyond the first SSM layer improves performance. In the all-layer variant, we apply the same input-derived Fisher log-power spectrum to every SSM layer, whereas the main method applies TDI only to the first layer. The two variants use the same architecture and training protocol and differ only in which layers receive TDI. We use five matched random seeds and train both variants for 30 epochs on sMNIST, 30 epochs on CIFAR-10, and 20 epochs on Frequency Classification.

\begin{table}[htbp]
\centering
\caption{
\textbf{First-layer versus all-layer TDI.}
Test accuracy (\%) is reported as mean $\pm$ standard deviation
over five matched seeds.
$\Delta$ denotes the paired per-seed accuracy difference
(all-layer minus first-layer), reported as mean $\pm$
standard deviation in percentage points.
Bold indicates the higher mean accuracy in each row.
}
\label{tab:tdi-layer-scope}
\small
\begin{tabular}{@{}llccc@{}}
\toprule
Dataset & Train ratio & First-layer TDI & All-layer TDI
& $\Delta$ (pp) \\
\midrule
sMNIST & 0.464\%
& 10.47 \textpm{} 0.85 & \textbf{30.39 \textpm{} 14.76}
& +19.92 \textpm{} 14.81 \\
sMNIST & 1.000\%
& 34.97 \textpm{} 23.93 & \textbf{80.41 \textpm{} 5.61}
& +45.45 \textpm{} 23.50 \\
sMNIST & 2.154\%
& 91.64 \textpm{} 3.74 & \textbf{95.22 \textpm{} 1.28}
& +3.58 \textpm{} 4.15 \\
\midrule
CIFAR-10 & 0.464\%
& 15.85 \textpm{} 1.36 & \textbf{17.79 \textpm{} 1.65}
& +1.94 \textpm{} 2.35 \\
CIFAR-10 & 1.000\%
& 22.63 \textpm{} 2.37 & \textbf{27.58 \textpm{} 3.32}
& +4.95 \textpm{} 3.43 \\
CIFAR-10 & 2.154\%
& \textbf{37.03 \textpm{} 1.92} & 33.93 \textpm{} 0.97
& \textminus{}3.09 \textpm{} 2.20 \\
\midrule
Freq. Cls. & 0.464\%
& \textbf{78.34 \textpm{} 3.56} & 72.60 \textpm{} 5.02
& \textminus{}5.74 \textpm{} 5.61 \\
Freq. Cls. & 1.000\%
& 84.98 \textpm{} 6.26 & \textbf{87.50 \textpm{} 5.80}
& +2.52 \textpm{} 1.86 \\
Freq. Cls. & 2.154\%
& 96.28 \textpm{} 1.33 & \textbf{96.54 \textpm{} 0.93}
& +0.26 \textpm{} 1.90 \\
\bottomrule
\end{tabular}
\end{table}

Table~\ref{tab:tdi-layer-scope} shows that all-layer TDI achieves higher mean accuracy at all three sMNIST ratios and the two smallest CIFAR-10 ratios, but underperforms first-layer TDI at the largest CIFAR-10 ratio and the smallest Frequency Classification ratio. The difference at the largest Frequency Classification ratio is small. Thus, all-layer TDI can help under the evaluated training budgets, but is not uniformly preferable to first-layer TDI.

This ablation reuses the input-level spectrum at every depth, although deeper layers receive representations transformed by preceding blocks. It therefore does not constitute a layer-dependent spectral initialization method. We retain first-layer TDI as the main method because it most directly follows our input-level analysis, while acknowledging that it is not optimal in every evaluated setting. Estimating task-relevant spectra from intermediate representations remains future work.

\subsection{Label Permutation during Task-Spectrum Estimation}
\label{app:label-permutation}

To examine whether the benefits of TDI depend on label-relevant spectral structure, we compare task spectra estimated using the true training labels (TDI-True) with those estimated after randomly permuting the training labels (TDI-Permuted). The permutation is applied only during task-spectrum estimation; downstream supervised training uses the original, correct labels in both conditions. We also include the default initialization as a baseline. Table~\ref{tab:label-permutation} reports the number of epochs required to reach 80\% test accuracy, defined as the first completed epoch at which this threshold is reached. Results are reported as mean $\pm$ standard deviation over five matched seeds.

\begin{table}[htbp]
\centering
\setlength{\tabcolsep}{4pt}
\caption{
\textbf{Effect of label permutation during task-spectrum estimation.}
Entries report epochs to reach 80\% test accuracy
(mean $\pm$ SD over five matched seeds; lower is better).
Bold indicates the lowest mean in each row.
}
\label{tab:label-permutation}
\small
\begin{tabular}{@{}lrccc@{}}
\toprule
Dataset & Train ratio & TDI-True & TDI-Permuted & Baseline \\
\midrule
sMNIST
& 2.15\%
& \textbf{20.0 \textpm{} 4.6}
& 28.6 \textpm{} 8.7
& 34.2 \textpm{} 4.5 \\

sMNIST
& 4.64\%
& \textbf{11.4 \textpm{} 2.6}
& 13.4 \textpm{} 2.9
& 15.4 \textpm{} 6.3 \\

sMNIST
& 10\%
& \textbf{5.2 \textpm{} 1.3}
& 7.0 \textpm{} 2.1
& 10.6 \textpm{} 1.8 \\
\midrule
Freq. Cls.
& 2.15\%
& \textbf{9.6 \textpm{} 4.0}
& 21.6 \textpm{} 13.8
& 11.6 \textpm{} 1.7 \\

Freq. Cls.
& 4.64\%
& \textbf{2.8 \textpm{} 1.6}
& 7.0 \textpm{} 3.7
& 4.2 \textpm{} 0.8 \\

Freq. Cls.
& 10\%
& \textbf{1.2 \textpm{} 0.4}
& 3.8 \textpm{} 1.9
& 2.0 \textpm{} 0.0 \\
\bottomrule
\end{tabular}
\end{table}

TDI-True has the lowest mean time-to-80\% in all six settings. Permuting labels increases the mean time by 1.8--12.0 epochs relative to TDI-True, and TDI-Permuted is not consistently faster than the baseline. These results support the role of label-dependent spectral structure in the observed early-learning benefits of TDI.

\section{Existing Assets}
We use standard public benchmarks and existing open-source software only for research evaluation. Table~\ref{tab:existing-assets} lists the existing assets used in our experiments, their original sources, and license or terms-of-use information when available. We do not redistribute these datasets. For CIFAR-10, we download data from \href{https://www.cs.toronto.edu/~kriz/cifar.html}{official CIFAR-10 website}. For UCR Time Series Archive, we download data from the \href{https://www.cs.ucr.edu/%7Eeamonn/time_series_data_2018/}{official archive website} and cite the archive paper.

\begin{table}[ht]
\centering
\caption{\textbf{Existing assets used in the experiments.}}
\label{tab:existing-assets}
\small
\setlength{\tabcolsep}{3.5pt}
\begin{tabular}{lll}
\toprule
Asset & Use in this paper & License / Terms \\
\midrule
MNIST~\citep{MNIST} & One-layer SSMs and Deep SSMs & MIT License \\
CIFAR-10~\citep{krizhevsky2009learning} & Deep SSMs & Official source \\
UCR Time Series Archive~\citep{UCR} & One-layer SSMs & Official source \\
PathFinder~\citep{tay2021long} & Deep SSMs & Apache-2.0 License \\
ListOps~\citep{nangia2018listops,tay2021long} & Deep SSMs & Apache-2.0 License \\
Speech Commands / SC10~\citep{warden2018speech} & Deep SSMs & CC-BY 4.0 License \\
Kernel Regression implementation~\citep{canatar2021spectral} & Kernel Regression & MIT License\\
S4/S4D implementation~\citep{gu2022parameterizationinitializationdiagonalstate} & Deep SSMs & Apache-2.0 License \\
\bottomrule
\end{tabular}
\end{table}


\newpage
\section*{NeurIPS Paper Checklist}
\begin{enumerate}

\item {\bf Claims}
    \item[] Question: Do the main claims made in the abstract and introduction accurately reflect the paper's contributions and scope?
    \item[] Answer: \answerYes{} 
    \item[] Justification: The main claims are stated in the abstract and Introduction, including the kernel-theoretic characterization of LTI SSM inductive bias and the proposed Task-Dependent Initialization method. The empirical scope of the claims is validated in Section~4, where we emphasize that the gains are strongest in low-data regimes and under spectral mismatch.

\item {\bf Limitations}
    \item[] Question: Does the paper discuss the limitations of the work performed by the authors?
    \item[] Answer: \answerYes{} 
    \item[] Justification:  We discuss limitations in Section 5, including the reliance of our theory on the LTI SSM case, its current focus on one-dimensional scalar sequences, and the restriction of TDI to first-layer initialization. We also clarify that TDI is not intended as a universal improvement and is expected to be most effective when task-relevant structure is spectral and mismatched with the default SSM bias.

\item {\bf Theory assumptions and proofs}
    \item[] Question: For each theoretical result, does the paper provide the full set of assumptions and a complete (and correct) proof?
    \item[] Answer: \answerYes{} 
    \item[] Justification: The theoretical assumptions and statements are given in Sections~2--3, including the LTI SSM formulation, whitened-coordinate analysis, and absolute-summability condition for the frequency-response result. Proofs of the main theorems and supporting lemmas are provided in Appendix~B.

    \item {\bf Experimental result reproducibility}
    \item[] Question: Does the paper fully disclose all the information needed to reproduce the main experimental results of the paper to the extent that it affects the main claims and/or conclusions of the paper (regardless of whether the code and data are provided or not)?
    \item[] Answer: \answerYes{} 
    \item[] Justification: We provide the information needed to reproduce the experiments supporting the main claims in Sections~3--4 and Appendices~C--G, including the TDI procedure, dataset choices, training-ratio protocol, model configurations, hyperparameters, random seeds, and evaluation metrics. Appendices~F--G further report full per-dataset and per-ratio results for the one-layer and deep SSM experiments.

\item {\bf Open access to data and code}
    \item[] Question: Does the paper provide open access to the data and code, with sufficient instructions to faithfully reproduce the main experimental results, as described in supplemental material?
    \item[] Answer: \answerYes{} 
    \item[] Justification: We provide a public code repository with environment files, README documentation, and scripts required to reproduce the main experiments. The repository documents the software environment, dataset preparation procedures, and commands for running the kernel regression, one-layer SSM, and deep SSM experiments. It also specifies how datasets are obtained or generated, including public datasets used in the experiments and synthetic datasets created by our code.

\item {\bf Experimental setting/details}
    \item[] Question: Does the paper specify all the training and test details (e.g., data splits, hyperparameters, how they were chosen, type of optimizer) necessary to understand the results?
    \item[] Answer: \answerYes{}
    \item[] Justification: We describe the experimental settings in Section~4 and provide full details in Appendix~E, including datasets, data splits, training-ratio protocols, model configurations, optimizers, hyperparameters, evaluation metrics, and random seed counts. Full per-dataset and per-ratio results are reported in Appendices~F--G to make the reported trends interpretable.

\item {\bf Experiment statistical significance}
    \item[] Question: Does the paper report error bars suitably and correctly defined or other appropriate information about the statistical significance of the experiments?
    \item[] Answer: \answerYes{}
    \item[] Justification: We report variability across random seeds and ratio-level averages in Tables~1 and~3, where each cell gives the mean and standard deviation as defined in the captions. We further report paired TDI--baseline improvements with cluster-bootstrap 95\% confidence intervals in Tables~2 and~4, matching runs by dataset, training ratio, and seed.

\item {\bf Experiments compute resources}
    \item[] Question: For each experiment, does the paper provide sufficient information on the computer resources (type of compute workers, memory, time of execution) needed to reproduce the experiments?
    \item[] Answer: \answerYes{} 
    \item[] Justification: We report compute resources in Appendix~E, including GPU types, the one-GPU-per-run setup, experiment scale in terms of datasets, ratios, seeds, and model variants, and the approximate total budget of 400--500 GPU-hours. We also state that TDI only adds an initialization step and does not change inference cost.
    
\item {\bf Code of ethics}
    \item[] Question: Does the research conducted in the paper conform, in every respect, with the NeurIPS Code of Ethics \url{https://neurips.cc/public/EthicsGuidelines}?
    \item[] Answer: \answerYes{} 
    \item[] Justification: We have reviewed the NeurIPS Code of Ethics and believe that the research conducted in this paper conforms to it. 

\item {\bf Broader impacts}
    \item[] Question: Does the paper discuss both potential positive societal impacts and negative societal impacts of the work performed?
    \item[] Answer: \answerNA{} 
    \item[] Justification:  The paper is foundational and methodological, focusing on the theoretical characterization of spectral inductive bias in LTI SSMs and a task-dependent initialization procedure. It does not introduce a deployed system, application-specific model, new dataset, or capability with a direct path to identifiable societal impacts beyond general improvements in sequence-model training.
    
\item {\bf Safeguards}
    \item[] Question: Does the paper describe safeguards that have been put in place for responsible release of data or models that have a high risk for misuse (e.g., pre-trained language models, image generators, or scraped datasets)?
    \item[] Answer: \answerNA{}{} 
    \item[] Justification: The paper does not release data or models with a high risk of misuse, such as pretrained language models, image generators, or scraped datasets. The experiments use public benchmark datasets and synthetic data, and the proposed method is an initialization procedure rather than a deployable high-risk model.

\item {\bf Licenses for existing assets}
    \item[] Question: Are the creators or original owners of assets (e.g., code, data, models), used in the paper, properly credited and are the license and terms of use explicitly mentioned and properly respected?
    \item[] Answer: \answerYes{} 
    \item[] Justification:  We cite the original sources for the public datasets and existing code assets used in the experiments, and Appendix~H lists the available license or terms-of-use information for each asset. For assets whose official sources do not specify a standard license, such as CIFAR-10 and the UCR Time Series Archive, we use the official sources, provide the requested citations, and do not redistribute the datasets.

\item {\bf New assets}
    \item[] Question: Are new assets introduced in the paper well documented and is the documentation provided alongside the assets?
    \item[] Answer: \answerYes{} 
    \item[] Justification: The new assets released with the submission consist of an anonymized code package and accompanying documentation, including environment specifications, reproduction commands, and license information for each component. The release does not include newly collected human-subject data, and the experiments rely on synthetic or standard public datasets that are documented in the repository.


\item {\bf Crowdsourcing and research with human subjects}
    \item[] Question: For crowdsourcing experiments and research with human subjects, does the paper include the full text of instructions given to participants and screenshots, if applicable, as well as details about compensation (if any)? 
    \item[] Answer: \answerNA{} 
    \item[] Justification:  The paper does not involve crowdsourcing experiments, human-subject research, user studies, or participant data collection. All experiments use synthetic data or standard public machine-learning benchmarks.

\item {\bf Institutional review board (IRB) approvals or equivalent for research with human subjects}
    \item[] Question: Does the paper describe potential risks incurred by study participants, whether such risks were disclosed to the subjects, and whether Institutional Review Board (IRB) approvals (or an equivalent approval/review based on the requirements of your country or institution) were obtained?
    \item[] Answer: \answerNA{} 
    \item[] Justification:  The paper does not involve crowdsourcing, human-subject research, user studies, or participant data collection. Therefore, IRB approval or equivalent human-subjects review is not applicable.

\item {\bf Declaration of LLM usage}
    \item[] Question: Does the paper describe the usage of LLMs if it is an important, original, or non-standard component of the core methods in this research? Note that if the LLM is used only for writing, editing, or formatting purposes and does \emph{not} impact the core methodology, scientific rigor, or originality of the research, declaration is not required.
    \item[] Answer: \answerNA{} 
    \item[] Justification: The core method development, theoretical analysis, and experimental evaluation in this paper do not involve LLMs as an important, original, or non-standard component. Any LLM assistance, if used, was limited to writing, editing, or formatting and did not affect the methodology, scientific claims, or experimental results.

\end{enumerate}
\end{document}